\documentclass[final]{article}

\usepackage{arxiv}

\usepackage[utf8]{inputenc} % allow utf-8 input
\usepackage[T1]{fontenc}    % use 8-bit T1 fonts
\usepackage{hyperref}       % hyperlinks
\usepackage{booktabs}       % professional-quality tables
\usepackage{amsfonts}       % blackboard math symbols
\usepackage{nicefrac}       % compact symbols for 1/2, etc.
\usepackage{microtype}      % microtypography
\usepackage{lipsum}
\usepackage{graphicx}
\usepackage{url}

\usepackage{xcolor}
\usepackage{tikz}
\usepackage{amsmath}
\usepackage{amssymb}

\usepackage{lineno}
\usepackage{algorithm2e}
\usepackage{textcomp} 
\usepackage{subcaption}
\usepackage{adjustbox}
\usepackage{url} 

\modulolinenumbers[5]
\usetikzlibrary{calc}
\usetikzlibrary{intersections}
\usetikzlibrary{through}

\usepackage{nomencl}
\usepackage{etoolbox}

\makenomenclature

\title{Review of Kernel Learning for Intra-Hour Solar Forecasting with Infrared Sky Images and Cloud Dynamic Feature Extraction}

\author{
 Guillermo Terr\'en-Serrano \\
  Department of Electrical and Computer Engineering \\
  The University of New Mexico \\
  Albuquerque, NM 87131, United States\\
  \texttt{guillermoterren@unm.edu} \\
 \And
  Manel Mart\'inez-Ram\'on \\
  Department of Electrical and Computer Engineering \\
  The University of New Mexico \\
  Albuquerque, NM 87131, United States\\
  \texttt{manel@unm.edu} \\
}

\begin{document}

\maketitle

\begin{abstract}
    The uncertainty of the energy generated by photovoltaic systems incurs an additional cost for a guaranteed, reliable supply of energy (i.e., energy storage). This investigation aims to decrease the additional cost by introducing probabilistic multi-task intra-hour solar forecasting (feasible in real time applications) to increase the penetration of photovoltaic systems in power grids. The direction of moving clouds is estimated in consecutive sequences of sky images by extracting features of cloud dynamics with the objective of forecasting the global solar irradiance that reaches photovoltaic systems. The sky images are acquired using a low-cost infrared sky imager mounted on a solar tracker. The solar forecasting algorithm is based on kernel learning methods, and uses the clear sky index as predictor and features extracted from clouds as feature vectors. The proposed solar forecasting algorithm achieved 16.45\% forecasting skill 8 minutes ahead with a resolution of 15 seconds. In contrast, previous work reached 15.4\% forecasting skill with the resolution of 1 minute. Therefore, this solar forecasting algorithm increases the performances with respect to the state-of-the-art, providing grid operators with the capability of managing the inherent uncertainties of power grids with a high penetration of photovoltaic systems.
\end{abstract}

\keywords{Flow Visualization \and Girasol Dataset \and Kernel Learning \and Machine Learning \and Solar Forecasting \and Sky Imaging}

\printnomenclature[2cm]
% Abbreviations
\nomenclature[A]{\textbf{PV}}{Photovoltaic}
\nomenclature[A]{\textbf{TSI}}{Total Sky Imager}
\nomenclature[A]{\textbf{GSI}}{Global Solar Irradiance}
\nomenclature[A]{\textbf{CSI}}{Clear Sky Index}
\nomenclature[A]{\textbf{IR}}{Infrared}
\nomenclature[A]{\textbf{ANN}}{Artificial Neural Networks}
\nomenclature[A]{\textbf{ML}}{Machine Learning}
\nomenclature[A]{\textbf{FOV}}{Field of View}
\nomenclature[A]{\textbf{SVM}}{Support Vector Machine for Regression}
\nomenclature[A]{\textbf{GPR}}{Gaussian Processes for Regression}
\nomenclature[A]{\textbf{KRR}}{Kernel Ridge Regression}
\nomenclature[A]{\textbf{RKHS}}{Reproducing Kernel Hilbert Space}
\nomenclature[A]{\textbf{L}}{Linear kernel}
\nomenclature[A]{\textbf{P}$^n$}{Polynomial kernel of order $n$}
\nomenclature[A]{\textbf{RBF}}{Radial Basis Function kernel}
\nomenclature[A]{\textbf{RQ}}{Rational Quadratic kernel}
\nomenclature[A]{\textbf{M}$_{\nu}$}{Mat\'ern kernel with parameter $\nu$}
\nomenclature[A]{\textbf{MSE}}{Mean Squared Error}
\nomenclature[A]{\textbf{MT-KRR}}{Multi-Task Kernel Ridge Regression}
\nomenclature[A]{\textbf{MLL}}{Marginal Log-Likelihood}
\nomenclature[A]{\textbf{MT-GPR}}{Multi-Task Gaussian Process for Regression}
\nomenclature[A]{\textbf{RVM}}{Relevance Vector Machine for Regression}
\nomenclature[A]{\textbf{KKT}}{Karush-Kuhn-Tucker}
\nomenclature[A]{\textbf{QP}}{Quadratic Programming}
\nomenclature[A]{\textbf{MT-SVM}}{Multi-Task Support Vector Machine for Regression}
\nomenclature[A]{\textbf{MT-RVM}}{Relevance Vector Machine for Regression}

\nomenclature[A]{\textbf{MAPE}}{Mean Absolute Percentage Error}

\nomenclature[A]{\textbf{MALR}}{Moist Adiabatic Lapse Rate}
\nomenclature[A]{\textbf{FS}}{Forecasting Skill}

\nomenclature[U]{$W$}{Watts}
\nomenclature[U]{$K$}{Kelvin}
\nomenclature[U]{$cK$}{Centikelvin}
\nomenclature[U]{$s$}{Seconds}
\nomenclature[U]{$ms$}{Milliseconds}
\nomenclature[U]{$km$}{Kilometers}
\nomenclature[U]{$m$}{Meters}
\nomenclature[U]{$min$}{Minutes}
\nomenclature[U]{$\%$}{Percentage (i.e., amount per hundred)}
\nomenclature[U]{$mm$}{Millimeters}
\nomenclature[U]{$\mu m$}{Micrometers}
\nomenclature[U]{${}^\circ$}{Degrees}

The current energy system is transitioning towards renewable and sustainable resources driven by technological innovations \cite{Green2021} and climate change mitigation policies \cite{SINGH2019}. Policies in the European Union aimed towards carbon neutrality are in place with others on the horizon \cite{BERTOLDI2020}. In addition, the United States recently introduced climate policy packages to encourage the usage of renewable energy \cite{STEFFEN2020}, and China announced important measures to reduce carbon emissions \cite{CARLSON2021}. The decrease in manufacturing cost and increase in efficiency of Photovoltaic (PV) panels are making solar energy a viable alternative to conventional thermal power plants \cite{ELIA2020} as well as the least expensive energy source in the market in some regions \cite{BOGDANOV2021}. New technological advances have resulted in 47.1\% efficiency in solar cells \cite{Geisz2020}. However, this efficiency is still far from the feasible theoretical maximum of a solar cell \cite{Vos1981}.

The energy generated from PV systems is sensitive to the fluctuations in irradiance caused by moving clouds \cite{JARVELA2020}. These events produce instabilities on the grid that need to be addressed \cite{LAPPALAINEN2020}. Moreover, the intermittency of solar radiation follows nonlinear global patterns that will decrease the reliability of PV systems as climate disruption intensifies \cite{Yin2020}. The proper configuration of  PV arrays can reduce the impact of the fluctuations, but these still exceed the ramp rates allowed by grid operators \cite{LAPPALAINEN2017}. To stabilize the operation and increase the penetration of PV systems in the power grid, grid operators will require intra-day solar forecasting (5 - 15 minutes in advance) for energy providers that have solar resources in their generation mix \cite{WEST2014}. Intra-hour solar forecasting would be used for scheduling transmission services, so that grid operators have the capability of controlling possible voltage fluctuations in the smart grid caused by moving clouds \cite{URQUHART2013}. In intra-hour forecasting, the most effective models include cloud information extracted from ground-based sky imagers \cite{BARBIERI2017}.

The Total Sky Imager (TSI) was initially developed for cloud coverage estimation \cite{Long2001}, but was subsequently found useful in solar forecasting applications \cite{CHOW2011}. The TSI consists of a concave reflective mirror that concentrates the light beams in a visible light camera. The "all-sky" or "whole sky" imager is a low-cost alternative to the TSI that enlarges the Field of View (FOV) of visible light cameras (i.e., skycams) by attaching a fish-eye lens \cite{FU2013}. The main disadvantage of these sky imagers is that the pixels in the circumsolar area are saturated \cite{GOHARI2014}. To overcome this problem, the proposed sky imagers may include a Sun blocking mechanism, but this obstructs part of the sky, reducing the visibility of the imager \cite{Dev2014}. This drawback is important for intra-hour solar forecasting \cite{Shields2013}. To reduce the saturation, a viable alternative is a reflective sky imager (similar to the TSI, but using an Infrared (IR) camera instead) \cite{Redman2018}. Another proposed alternative is enlarging the FOV of an IR sky imager using multiple low-cost far IR cameras \cite{MAMMOLI2019}. In this research, we propose a sky imager that has a low-cost far IR camera mounted on a double axis solar tracker, which maintains the Sun in the center of the images throughout the day and reduces the saturation of the pixels in the circumsolar area \cite{TERREN2020d}.

In intra-hour forecasting applications, Artificial Neural Networks (ANNs) are the most commonly used Machine Learning (ML) algorithms \cite{ANTONANZAS2016, PAZIKADIN2020}. ANNs are very efficient at handling big data problems, since the training is performed via stochastic gradient descent \cite{lecun1998}. The main disadvantage of using a global learning approach (i.e., ANNs) is that these models might overlook local characteristics of the data space for the sake of global accuracy \cite{Hand2003}. In these circumstances, local learning methods may provide a better understanding of the features contributions and the data space \cite{Kopitar2019}, leading to better performances \cite{bottou1992}. The drawback of local learning models is that they require training multiple models \cite{kivinen2004}, which is computationally expensive (involving matrix inversion). The reason behind the approach proposed in this investigation of training local models using kernel learning, is due to the relatively low number of samples. Kernel learning methods (i.e., shallow learning) have hyperparameters for fine-tuning the regularization of the model parameters \cite{Varma2009}. In contrast, ANNs require large amounts of data to avoid overfitting problems \cite{Xie2021}.

Kernel learning has been mainly implemented in day-ahead forecasting \cite{Mellit2020}. The methods most commonly found in the literature are the Support Vector Machine for Regression (SVM) and more recently the Gaussian Process for Regression (GPR) \cite{AKHTER2019}. SVM are used in monthly \cite{OLATOMIWA2015}, intra-week \cite{DEO2016}, and intra-day solar forecasting application \cite{ZENDEHBOUDI2018} using feature vectors composed of weather features acquired in one or multiple weather stations \cite{JIANG2016}. When SVM are used for intra-hour PV power forecasting (from 15 to 300 minutes ahead) satellite images are required to estimate the motion of clouds \cite{Jang2016}. The most effective intra-day PV power forecasting methods without sky images either apply processing techniques to the feature vectors \cite{PAN2020} or  are used in combination with measurements acquired from sensors installed in the PV systems \cite{Preda2018}. Least-squares SVM have also been previously implemented in GSI \cite{Guermoui2020} and PV power forecasting \cite{ZENG2013}. Kernel Ridge Regression (KRR) has been put forward in PV power forecasting. In a particular case, the formulation of the KRR was improved for adding robustness to outliers \cite{Dash2020}. In the most recent investigations that used kernel learning, GPRs are used in intra-week solar \cite{ROHANI2018} and PV power forecasting \cite{Wang2021}. 

In intra-hour solar forecasting, the direction of clouds is an important feature to determine if a cloud will occlude the Sun \cite{zhen2019, dissawa2021}. To do this, it is necessary to analyze the dynamics of a cloud over a sequence of consecutive images \cite{mondragon2020}. This investigation introduces a method to compute the probability of a cloud intersecting the Sun, and uses it in the quantification of the cloud dynamics features extracted from a series of IR sky images. The statistics of the extracted features are computed to form the feature vectors that are utilized in the proposed intra-hour solar forecasting.

The performance of PV panels and power electronics depends on the manufacturing process and degrades following different patterns \cite{hayashi2018}. When the GSI is used as a predictor, the information obtained from solar forecasting may be shared between nearby PV systems if the degradation pattern of each PV system is known \cite{elsinga2017, rosato2019}. A time series model improves forecasting performances when the deterministic component is removed \cite{khashei2010, makridakis2018}. In this investigation, the GSI measurements are detrended to obtained the Clear Sky Index (CSI) \cite{ENGERER2014}, and the performances of multi-task kernel learning models (i.e., SVM, Relevance Vector Machine (RVM), KRR and GPR) are compared when using multiple kernel functions in combination with different feature vectors. The features included in the vectors were extracted from the analyses of cloud dynamics.

\section{Kernel Methods}\label{sec:kernel_learning}

A kernel method can be seen as an extension of a linear algorithm that acquires nonlinear properties through the so called Kernel trick. a kernel is a definite positive function $\mathcal{K
(\cdot,\cdot)}$ that maps a couple of vectors in a space $\mathbb{R}^{D}$ into $\mathcal{R}$. Since this function is positive definite, the Mercer's theorem \cite{Mercer1909,aronszajn1954invariant, aizerman1964theoretical} states that it is a dot product in a Hilbert space $\mathcal{H}$ spanned by functions $\varphi({\bf x})$ such that $\mathcal{K}({\bf x},{\bf x}')=\langle\varphi({\bf x}),\varphi({\bf x}') \rangle $. The Generalized Representer Theorem \cite{Scholkopf2001} states that the parameters $\bf w$ of an estimation function $f({\bf x})={\bf w}^\top\varphi(\bf x)$ are expressable as a linear combination  ${\bf w}=\sum_i \alpha_i \varphi({\bf x}_i)$ of the training data. Therefore, for any (nonlinear) kernel, a linear algorithm has a nonlinear counterpart that can be written as
\begin{equation}
    f({\bf x}) = \sum_i \alpha_i \langle \varphi({\bf x}_i), \varphi({\bf x}) \rangle =  \sum_i  \alpha_i  \mathcal{K}({\bf x}_i, {\bf x})
\end{equation}
thus being possible to generalize any linear algorithm to have nonlinear properties. In particular, SVM or GPR classification or regression algorithms, KRR or others can be endowed with nonlinear properties in a straightforward way as it will be summarized below. 

Kernel regression models can be extended to multi-task kernel regression models. Multi-task models produce multiple predictors $\hat{\mathbf{y}}_i$ from an unique covariate vector $\mathbf{x}_i$ (i.e., feature vector). In this investigation, the kernel methods are divided into dense and sparse methods. At the same time, a deterministic and probabilistic model are proposed within the dense and sparse methods. 

\paragraph{Kernel Functions}\label{sec:kernels}
The proposed kernel functions in this analysis are linear ($L$), polynomial of order $n$ ($P^n$), Radial Basis Function ($RBF$), Rational Quadratic ($RQ$), and Mat\'ern ($M$) \cite{SHAWE2004}. Their respective functions are,
\begin{equation}
    \begin{split}
        \mathcal{K}_{L} \left( \mathbf{x}_i, \mathbf{x}_j \right) &= \gamma \mathbf{x}_i^\top \mathbf{x}_j, \\ 
        \mathcal{K}_{P^n} \left( \mathbf{x}_i, \mathbf{x}_j \right) &= \left( \gamma \mathbf{x}_i^\top \mathbf{x}_j + \beta \right)^n, \\
        \mathcal{K}_{RBF} \left( \mathbf{x}_i, \mathbf{x}_j \right) &= \exp \left( - \gamma || \mathbf{x}_i - \mathbf{x}_j ||^2 \right), \\
        \mathcal{K}_{RQ} \left( \mathbf{x}_i, \mathbf{x}_j \right) &= \left( 1 + \frac{1}{2\alpha} \gamma || \mathbf{x}_i - \mathbf{x}_j ||^2 \right)^{-\alpha}, \\
        \mathcal{K}^{\nu}_{M} \left( \mathbf{x}_i, \mathbf{x}_j \right) &= \frac{2^{1 - \nu}}{\Gamma (\nu)} \left( \sqrt{2\nu} \cdot \gamma || \mathbf{x}_i - \mathbf{x}_j ||^2  \right)^\nu K_\nu \left( \sqrt{2\nu} \cdot \gamma || \mathbf{x}_i - \mathbf{x}_j ||^2 \right),
    \end{split}
\end{equation}
where $\gamma, \ \beta, \ \alpha, \ \nu \in \mathbb{R}^+$, and $n \in \mathbb{N}$ are the kernel hyperparameters \cite{SCHOLKOPF2000}. $\Gamma (\cdot)$ is the Gamma function, and $K_\nu$ is the modified Bessel function of second kind. 

The hyperparameter $\gamma, \ \beta, \ \alpha$ are cross-validated. The only polynomial kernel validated is when $n = 2$, since exploratory results showed that polynomial kernels of higher order leads to poor performances. The hyperparameters of the Mat\'en Kernel are independently validated for $\nu \in \{1/2, 3/2, 5/3 \}$, which are the standard values used for the smoothing parameter. 

\subsection{Dense Kernel Methods}

A dense kernel method utilizes the entire training set to produce a prediction. In contrast, a sparse kernel method utilizes a subset of the training data yielding (in general) to faster models in the implementation. In this investigation, sparsity is not an essential requirement because a subset of samples is selected from the database, instead of training a model using the entire dataset.

\subsubsection{Kernel Ridge Regression}

Assume a linear regression model of the form:
\begin{equation}
   y_i = \hat{y}_i  + \varepsilon_i = \mathbf{w}^\top \varphi\left( \mathbf{x}_i \right) + \varepsilon_i.
\end{equation}
The KRR is a simplification of the Tikhonov regularization \cite{TIKHONOV1977}, in which the minimized loss functions is the Mean Squared Error (MSE) with a quadratic norm regularization applied to the parameters $\mathbf{w}$,
\begin{equation}
    \label{eq:ridge_regression}
    \min_{\mathbf{w}} \ \mathbb{E} \left[ y_i - \mathbf{w}^\top \varphi \left( \mathbf{x}_i \right) \right] + \gamma \| \mathbf{w} \|_2.
\end{equation}
where $\gamma$ is the regularization hyperparameter. 

Applying the representer theorem \cite{Scholkopf2001}, which states that the dual formulation of the model parameters is $\mathbf{w} = \boldsymbol{\varPhi} \boldsymbol{\alpha}$, and adding the Vapnik-Chervonenkis generalization bound \cite{vapnik2013} to the formulation, we have that,
\begin{equation}
    \label{eq:kernel_ridge_regression}
    \min_{\boldsymbol{\alpha}} \ \mathbb{E} \left[ y_i - \boldsymbol{\alpha}^\top \boldsymbol{\varPhi}^\top \varphi \left( \mathbf{x}_i \right) \right] + \frac{\gamma}{N} \boldsymbol{\alpha}^\top \boldsymbol{\varPhi}^\top \boldsymbol{\varPhi} \boldsymbol{\alpha}
\end{equation}
where ${\boldsymbol \varPhi} = \left [\varphi( \mathbf{x}_1) ~\cdots ~\varphi ( \mathbf{x}_N) \right]$ is a matrix containing all training samples mapped into the RKHS.

Finding the optimal regularization hyperparameter requires cross-validation. Nevertheless, the model parameters that minimized the MSE are analytically found nulling the gradient,
\begin{equation}
    \begin{split}
        0 &= \frac{\partial}{\partial \boldsymbol{\alpha}} \left[ \left( \mathbf{y} - \boldsymbol{\alpha}^\top \mathbf{K} \right)\left( \mathbf{y} - \boldsymbol{\alpha}^\top \mathbf{K} \right)^\top + \frac{\gamma}{N} \boldsymbol{\alpha}^\top \mathbf{K} \boldsymbol{\alpha} \right], \\
        0 &= \frac{\partial}{\partial \boldsymbol{\alpha}} \left[ \mathbf{y} \mathbf{y}^\top + \boldsymbol{\alpha}^\top \mathbf{K} \mathbf{K}^\top \boldsymbol{\alpha} - 2 \boldsymbol{\alpha}^\top \mathbf{K} \mathbf{y}^\top + \frac{\gamma}{N} \boldsymbol{\alpha}^\top \mathbf{K} \boldsymbol{\alpha} \right], \\
        0 &= 2 \mathbf{K} \mathbf{K}^\top \boldsymbol{\alpha} - 2 \mathbf{K} \mathbf{y}^\top + \frac{2\gamma}{N} \mathbf{K} \boldsymbol{\alpha}, \\
        \boldsymbol{\alpha} &= \left( \mathbf{K} + \frac{\gamma}{N} \mathbf{I} \right)^{-1} \mathbf{y},
    \end{split}
\end{equation}
where $\mathbf{K} = \boldsymbol{\varPhi}^\top \boldsymbol{\varPhi}$.
A prediction for a new observation is obtained with this formula,
\begin{equation}
    \hat{y}_* =\sum_{i = 1}^N  \alpha_i \varphi \left(\mathbf{x}_i \right)^\top \varphi \left( \mathbf{x}_* \right) = \sum_{i = 1}^N \alpha_i \mathcal{K} \left( \mathbf{x}_i, \mathbf{x}_* \right).
\end{equation}
The optimal kernel hyperparamters are found implementing a cross-validation method.

\paragraph{Multi-Task Kernel Ridge Regression (MT-KRR)}
 
The dual formulation of the MT-KRR as a MSE minimization problem is,
\begin{equation}
    \label{eq:multiouput_kernel_ridge_regression}
    \min_{\tilde{\boldsymbol{\alpha}}} \ \mathbb{E} \left[ \tilde{\mathbf{y}} - \tilde{\boldsymbol{\alpha}}^\top \tilde{\mathbf{K}} \right] + \frac{\gamma}{CN} \| \tilde{\boldsymbol{\alpha}}^\top \tilde{\mathbf{K}} \tilde{\boldsymbol{\alpha}} \|_2.
\end{equation}
where $\tilde{\mathbf{K}} = \boldsymbol{\Gamma} \otimes \mathbf{K}$, the extended vector of predictors is $\tilde{\mathbf{y}} = \left[ \mathbf{y}_1^\top ~\cdots ~\mathbf{y}_N^\top \right] \in \mathbb{R}^{1 \times CN}$, and $C$ is the number of forecasting outputs in the model.

The dual parameters $\tilde{\boldsymbol{\alpha}}$ have an analytical solution analogous to the KRR which is $\tilde{\boldsymbol{\alpha}} = ( \tilde{\mathbf{K}} + \frac{\gamma}{CN} \mathbf{I}_{N \times N} \otimes \mathbf{I}_{C \times C} )^{-1} \tilde{\mathbf{y}}$. The ridge regularization parameter is $\gamma$, and $\boldsymbol{\Gamma}$ is the matrix that contains the correlation coefficients between the multiple outputs. A prediction for a new observation is obtained from
\begin{equation}
    \label{eq:multiouput_kernel_ridge_regression_prediction}
    \hat{\mathbf{y}}_* = \sum_{i = 1}^{CN} \tilde{\alpha}_i \left[ \boldsymbol{\Gamma} \otimes \mathcal{K} \left( \mathbf{x}_i, \mathbf{x}_* \right) \right],
\end{equation}
where a prediction $\hat{\mathbf{y}}_*$ is the vector of the multiple forecasting horizons $\hat{\mathbf{y}}_* = [\hat{y}_1 ~\cdots ~\hat{y}_C]^\top$.

\subsubsection{Gaussian Process for Regression}

Consider the standard model in Bayesian regression whose feature vectors are projected into a feature space \cite{williams2006},
\begin{equation}
   y_i = f \left( \mathbf{x}_i \right) + \varepsilon_i = \mathbf{w}^\top \varphi\left( \mathbf{x}_i \right) + \varepsilon_i.
\end{equation}
In this model, the prediction of the latent function $f \left( \mathbf{x}_i \right)$ is assumed to have a Gaussian independent and identically distributed error $\varepsilon_i \sim \mathcal{N} (0, \sigma_n^2)$, and the likelihood function of the observations is $y_i \sim \mathcal{N} ( \mathbf{w}^\top \varphi(\mathbf{x}_i), \sigma_n^2 \mathbf{I})$. The parameters $\mathbf{w}$ are defined as a latent random variable whose prior is $ \mathcal{N} ( \mathbf{0}, \Sigma_p )$, and when applying the Representer Theorem, it is obtained that the dual representation of $\mathbf{w}$ is 
\begin{equation}
    f \left( \mathbf{x}_i \right) = \sum_{i = 1}^N \alpha_i \varphi \left( \mathbf{x}_i \right)^\top \varphi \left( \mathbf{x}_j \right) = \boldsymbol{\alpha} \mathbf{K}
\end{equation}
where $ \mathcal{K} ( \mathbf{x}_i, \mathbf{x}_j ) \triangleq \varphi ( \mathbf{x}_i )^\top \Sigma_p \varphi ( \mathbf{x}_j )$ and $\bf K$ is the matrix of dot products between training data. Therefore, the prior of the dual representation of the parameters is $p ( \boldsymbol{\alpha} | \mathbf{X} ) \sim  \mathcal{N} ( \mathbf{0}, \mathbf{K}^{-1} )$

The dual of parameters ${\boldsymbol \alpha}$ that maximize the posterior distribution given the training predictors $y_i$ (i.e., Maximum a Posteriori estimation) are proportional to the prior times the likelihood,
\begin{equation}
    p\left( {\boldsymbol \alpha}|{\bf X},{\bf y}\right) \propto p\left({\bf y}|{\bf X},{\boldsymbol \alpha}\right) p \left( \boldsymbol{\alpha} \middle| \mathbf{X} \right).
\end{equation}
The mean and covariance matrix of the posterior distribution $p ( \boldsymbol{\alpha} | \mathbf{X}, \mathbf{y} ) \sim  \mathcal{N} ( \bar{\boldsymbol{\alpha}}, \mathbf{B} )$ are $\bar{\boldsymbol \alpha} = ( \mathbf{K} + \sigma^2_n \mathbf{I} )^{-1} \mathbf{y}$ and $\mathbf{B} = \mathbf{K}^{-1} - ( \mathbf{K} +\sigma_n^2 \mathbf{I} )^{-1}$ respectively.

As the posterior probability of a prediction $p ( f ( \mathbf{x}_* ) | {\mathbf{X}}, \mathbf{y} )$ is a Gaussian distribution, it is sufficient to compute the mean and variance with respect to the posterior of $\boldsymbol\alpha$ \cite{martinez2021}. Hence, the mean and the variance of a predictions are respectively
\begin{equation}\label{eq:gaussian_process_prediction}
    \begin{split}
        \bar{f}({\bf x}_*) &= \mathbb{E}_{\boldsymbol \alpha|{\bf X},{\bf y}}\left[{\boldsymbol \alpha}^{\top} {\bf k}\left({\bf x}_*\right)\right] = \bar{\boldsymbol{\alpha}}^\top {\bf k}({\bf x}_*),\\
        \sigma^2_*  &= \mathbb{E}_{\boldsymbol \alpha|{\bf X},{\bf y}}\left[ {\bf k}\left({\bf x}_*\right)^{\top}{\boldsymbol \alpha}{\boldsymbol \alpha}^{\top} {\bf k}\left({\bf x}_*\right)\right] = \mathbf{k} \left( \mathbf{x}_* \right) \mathbf{B} \mathbf{k} \left( \mathbf{x}_* \right) \\
        &= \mathcal{K} \left( \mathbf{x}_* , \mathbf{x}_* \right) - \mathbf{k} \left( {\bf x}_* \right)^\top \left( \mathbf{K} + \sigma_n^2 \mathbf{I} \right)^{-1} \mathbf{k} \left( \mathbf{x}_* \right).
    \end{split}
\end{equation}
where $\mathbf{k}( \mathbf{x}_* ) \triangleq \mathcal{K} ( \mathbf{x}_i, \mathbf{x}_*)$ is a column vector containing all dot products. 

The Marginal Log-Likelihood (MLL) (i.e., evidence) is used to optimize the kernel hyperparameters via gradient descent,
\begin{equation}\label{eq:gaussian_process_mll}
    \log p \left( \mathbf{y} \middle| \mathbf{X} \right) = - \frac{1}{2} \mathbf{y}^\top \bar{\boldsymbol{\alpha}} - \frac{1}{2} \left| \mathbf{K} + \sigma_n^2 \mathbf{I} \right| - \frac{N}{2} \log 2 \pi.
\end{equation}

\paragraph{Multi-Task Gaussian Process for Regression (MT-GPR)}

Similarly to a GPR, the dual representation of parameters $\tilde{\boldsymbol{\alpha}}$ in a MT-GPR have analytical solution when the likelihood $p ( \mathbf{Y} | \mathbf{X}, \tilde{\boldsymbol{\alpha}} ) \sim \mathcal{N} ( \tilde{\boldsymbol{\alpha}} [ \boldsymbol{\Gamma} \otimes \mathbf{K}], \boldsymbol{\Sigma}_n \otimes \mathbf{I}) $ and the prior $p ( \tilde{\boldsymbol{\alpha}} | \mathbf{X} ) \sim \mathcal{N} ( \mathbf{0}, [\boldsymbol{\Gamma}\otimes \mathbf{K}]^{-1} )$ are defined as multivariate normal distribution. The predicted mean vector and covariance matrix for a new sample are 
\begin{equation}\label{eq:multitask_gaussian_process_prediction}
    \begin{split}
        \bar{f} \left(\mathbf{x}_* \right) &= \tilde{\boldsymbol{\alpha}}^\top \left[ \boldsymbol{\Gamma} \otimes \mathcal{K} \left( \mathbf{x}_i, \mathbf{x}_* \right) \right], \\
        \hat{\mathbf{\Sigma}}_* &= \boldsymbol{\Gamma} \otimes \mathcal{K} \left( \mathbf{x}_*, \mathbf{x}_* \right) - \left[ \boldsymbol{\Gamma} \otimes \mathcal{K} \left( \mathbf{x}_i, \mathbf{x}_* \right) \right]^\top \left( \tilde{\mathbf{K}} + \boldsymbol{\Sigma}_n \mathbf{I} \right)^{-1}  \boldsymbol{\Gamma} \otimes \mathcal{K} \left( \mathbf{x}_i, \mathbf{x}_* \right),
    \end{split}
\end{equation}
where $\tilde{\boldsymbol{\alpha}} = ( \tilde{\mathbf{K}} + \boldsymbol{\Sigma}_n \otimes \mathbf{I} )^{-1} \tilde{\mathbf{y}}$, and $\tilde{\mathbf{K}} = \mathbf{\Gamma} \otimes \mathbf{K}$. The matrix that models the correlation between outputs is $\mathbf{\Gamma}$, and the matrix of noises is defined as $\boldsymbol{\Sigma}_n = \mathrm{diag} (\sigma^2_{t,n})$, where $\sigma^2_{t,n}$ is the noise of output $t$.

The the optimal kernel hyperparatermes in a MT-GPR, are obtained minimizing the negative MLL via gradient descent,
\begin{equation}\label{eq:multitask_gaussian_process_mll}
    \log p \left( \mathbf{Y} \middle| \mathbf{X} \right) = - \frac{1}{2} \tilde{\mathbf{y}}^\top 
    \tilde{\boldsymbol{\alpha}} - \frac{1}{2} \left| \tilde{\mathbf{K}} + \boldsymbol{\Sigma}_n \otimes \mathbf{I} \right| - \frac{CN}{2} \log 2 \pi,
\end{equation}
where $C$ is the number of forecasting horizons. The correlation between outputs matrix $\boldsymbol{\Gamma}$ and the noise matrix $\boldsymbol{\Sigma}_n$ have analytical solutions when the Expectation-Maximization algorithm is implemented in the hyperparameters optimization \cite{Bonilla2008}.

\subsection{Sparse Kernel Methods}

The main idea behind sparse kernel methods, is to assume that exist noisy vectors in the nonlinear transformation to the feature space $\varphi(\mathbf{x}_i)$. Therefore, the transformation can be approximated by a subset of the data. The results are learning models that are computationally faster during the test.

The sparse kernel methods proposed in this research are $\varepsilon$-SVM and RVM. The disadvantage of $\varepsilon$-SVM is that does not predict a distribution. For that reason, we implement a RVM that is Bayesian sparse kernel method. We compare their performances in the application of solar irradiance forecasting.

\subsubsection{Support Vector Machine for Regression}\label{sec:support_vector_machine}

The regression problem in a $\varepsilon$-SVM applies an $\varepsilon$-insensitive loss function to the formulation for sparseness purposes \cite{SCHOLKOPF2000}, 
\begin{align}\label{eq:svm_regression}
    \left| y_i - f \left( \mathbf{x}_i \right)\right|_{\varepsilon} = \max \left[ 0, \left| y_i - f \left( \mathbf{x}_i \right)\right|-\varepsilon \right], \quad \forall i = 1, \dots, N, \quad y_i,\varepsilon \in \mathbb{R},
\end{align}
where $f ( \mathbf{x}_i ) = \mathbf{w}^\top \varphi ( \mathbf{x}_i ) + b$, $b \in \mathbb{R}$. The $\varepsilon$-insensitive loss function does not penalize errors that are below $ |\varepsilon| > 0$ \cite{DRUCKER1997}. 

The $\varepsilon$-SVM aims to estimate the $f ( \cdot )$ that minimizes following constrained problem, 
\begin{equation}
    \label{eq:primal_constraints}  
    \begin{split}
        \min_{\mathbf{w}, b, \xi, \xi^*} & \quad \frac{1}{2} \| \mathbf{w} \|^2 + \mathcal{C} \sum_{i = 1}^N \left( \xi_i + \xi_i^* \right) \\ 
        \mathrm{s.t.} & 
        \begin{cases}
            y_i - \mathbf{w}^\top \varphi \left( \mathbf{x}_i \right) - b & \leq \varepsilon + \xi_i \\
             \mathbf{w}^\top \varphi \left( \mathbf{x}_i \right) + b - y_i & \leq \varepsilon + \xi_i^* \\
            \xi_i, \xi_i^* & \geq 0
        \end{cases}  \quad i = 1, \ldots, N,  
    \end{split}
\end{equation}
in which the L2-norm is applied to the parameters $\mathbf{w}$ to control the model complexity, the $\varepsilon$-loss function controls the training error through the hyperparameter $\mathcal{C}$, and $\xi_i$ are the slack variables. The slack variables are introduced to relax the constraints of the optimization problem, so it is feasible to deal with non-convex problems \cite{CORTES1995}. 

To minimize the constrained problem in Eq.~\eqref{eq:primal_constraints}, a Lagrangian functional is defined through a set of dual variables using the functional and the set of constraints \cite{SMOLA2004}. The derivatives of the functional with respect to the primal variables $\mathbf{w},\varepsilon, \xi_i, \xi_i^*$ leads to a set of equations that are a case of Karush-Kuhn-Tucker (KKT) conditions. Substituting these equations on the Lagrangian functional, together with the complimentary KKT condition (which forces the product of dual parameters $\alpha_i,\alpha^*_i$ with the constraints to be zero) yield to the following dual functional that has the form of a solvable Quadratic Programming (QP) problem,
\begin{equation}\label{eq:support_vector_machine_dual}
    \begin{split}
        \min_{\boldsymbol{\alpha}, \boldsymbol{\alpha}^*} &\quad \frac{1}{2} \left( \boldsymbol{\alpha} - \boldsymbol{\alpha}^* \right)^\top \mathbf{K} \left( \boldsymbol{\alpha} - \boldsymbol{\alpha}^* \right) + \left( {\boldsymbol \alpha} - {\boldsymbol \alpha}^* \right)^{\top}{\bf y} + \varepsilon  \mathbf{1}^\top \left( \boldsymbol{\alpha} + \boldsymbol{\alpha}^* \right) \\
        &\mathrm{s.t.}
        \begin{cases} 
            \mathbf{1}^\top \left( \boldsymbol{\alpha} - \boldsymbol{\alpha}^* \right) = 0 \\ 
            0 \leq \alpha_i, \alpha_i^* \leq \mathcal{C}
        \end{cases} \ \forall i = 1, \dots, N.
    \end{split}
\end{equation}
where $\bf{K}$ is the Gram matrix that contains the dot products ${\bf K}_{i,j} = \mathcal{K}({\bf x}_i,{\bf x}_j)$ and $\mathbf{1}_{1 \times N} = [1 \cdots 1 ]^\top$ is a vector of ones. 

The approximated function in Eq.~\eqref{eq:svm_regression} evaluated for a new sample $\mathbf{x}_*$ is,
\begin{equation}
    f \left( \mathbf{x}_* \right) = \sum_{i = 1}^N  \left(\alpha_i  - \alpha^*_{i}\right) \mathcal{K} \left( \mathbf{x}_i, \mathbf{x}_* \right) + b,
\end{equation}
where $b$ is obtained from the complimentary KKT conditions. 

\paragraph{Multi-Task Support Vector Machine for Regression ($\varepsilon$-MT-SVM)}

The primal $\varepsilon$-MT-SVM problem can be formulated as 
\begin{equation}
    {\bf y}_i = \tilde{\bf w}^\top [\mathbf{\Gamma} \otimes \varphi({\bf x}_i)] + {\bf b},
\end{equation}
where the column vectors of primal parameter $\tilde{\bf w}$ and model bias $\mathbf{b} = [b_1 ~\cdots ~b_C]^\top$ approximates each one of the predictors ${\bf y}_i \in \mathbb{R}^C$. The matrix $\mathbf{\Gamma}$ contains the correlation parameters between outputs. Primal parameters are a function of the dual parameters $\boldsymbol{\alpha}_i, \boldsymbol{\alpha}_i^{*}$ as well.

The Gram matrix $\tilde{\mathbf{K}}_{CN \times CN}$ in the $\varepsilon$-MT-SVM formulation for correlated outputs is $\tilde{\mathbf{K}} = \boldsymbol{\Gamma} \otimes \mathbf{K}$. Therefore, the dual formulation of the QP problem for the $\varepsilon$-MT-SVM is,
\begin{equation}\label{eq:multitask_support_vector_machine_dual}
    \begin{split}
        \min_{\tilde{\boldsymbol{\alpha}}, \tilde{\boldsymbol{\alpha}}^*} & \quad \frac{1}{2} \left( \tilde{\boldsymbol{\alpha}} - \tilde{\boldsymbol{\alpha}}^* \right)^\top \tilde{\mathbf{K}} \left( \tilde{\boldsymbol{\alpha}} - \tilde{\boldsymbol{\alpha}}^* \right) + \tilde{\mathbf{y}}^\top \left( \tilde{\boldsymbol{\alpha}} - \tilde{\boldsymbol{\alpha}}^* \right) + \varepsilon \mathbf{1}^\top \left( \tilde{\boldsymbol{\alpha}} + \tilde{\boldsymbol{\alpha}}^* \right) \\
        &\mathrm{s.t.}
        \begin{cases} 
            \mathbf{1}^\top \left( \tilde{\boldsymbol{\alpha}} - \tilde{\boldsymbol{\alpha}}^* \right) = 0 \\ 
            \mathbf{0} \leq \tilde{\alpha}_i, \tilde{\alpha}_i^* \leq \mathcal{C} \\
        \end{cases}  \ \forall i = 1, \dots, 2 CN.
    \end{split}
\end{equation}
A multi-task prediction is performed applying this formula,
\begin{equation}\label{eq:multitask_support_vector_machine_prediction}
    \hat{\mathbf{y}}_* =  \left[ \boldsymbol{\Gamma} \otimes \mathcal{K} \left( \mathbf{x}_i, \mathbf{x}_* \right) \right] \left( \tilde{\boldsymbol{\alpha}}  - \tilde{\boldsymbol{\alpha}}^* \right)^\top + \mathbf{b}.
\end{equation}

\subsubsection{Relevance Vector Machine for Regression}\label{sec:relevance_vector_machine}

This model presents as an alternative solution of the functional in a $\varepsilon$-SVM with the advantage of producing probabilistic predictions \cite{Tipping1999}. The Bayesian formulation of the likelihood and prior in a RVM is equivalent to a Gaussian process, but this model is endowed with an automatic relevance determination mechanisms that causes a subset of the parameters $\mathbf{w}$ to drive to zero \cite{Tipping2001}.

The likelihood function of the observations is defined as multivariate normal distribution $p ( y_i | \mathbf{x}_i, \mathbf{w}, \sigma_n^{-2}) \sim \mathcal{N} ( \mathbf{w}^\top \varphi ( \mathbf{x}_i), \sigma_n^{-2} \mathbf{I})$, and a zero-mean Gaussian prior is set on the weights $p ( w_i | \lambda^{-1} ) \sim \mathcal{N} ( w_i | 0, \lambda^{-1}_i )$. Applying the Bayes theorem, we obtained that the posterior distribution of the weights is also Gaussian,
\begin{equation}
    p \left( \mathbf{w} \middle| \mathbf{y}, \mathbf{X}, \boldsymbol{\alpha}, \sigma_n^{-2} \right) = \mathcal{N} \left( \mathbf{w} \middle| \boldsymbol{\mu}, \boldsymbol{\Sigma} \right).
\end{equation}
The mean and covariance matrix of the posterior distribution are,
\begin{equation}
    \begin{split}
        \boldsymbol{\Sigma}^{(t + 1)} &= \left( \boldsymbol{\Lambda}^{(t)} + {\sigma_n^{-2}}^{(t)} \mathbf{K}^\top \mathbf{K} \right)^{-1} \\
        \boldsymbol{\mu}^{(t + 1)} &= {\sigma_n^{-2}}^{(t)} \boldsymbol{\Sigma}^{(t + 1)} \mathbf{K}^\top \mathbf{y}
    \end{split}
\end{equation}
where $\sigma^2$ is the variance of the noise, t represents the current optimization iteration, $\boldsymbol{\Lambda} = \mathrm{diag} (\boldsymbol{\lambda})$ are the precision of the parameters, and  $\mathbf{K}$ is the Gram matrix.

The marginal likelihood is maximized to find the optimal hyperparameters,
\begin{equation}
    p \left( \mathbf{y} \middle| \mathbf{X}, \boldsymbol{\lambda}, \sigma_n^{-2} \right) = \int p \left(\mathbf{y} \middle| \mathbf{X}, \mathbf{w}, \sigma_n^{-2} \right) p \left( \mathbf{w} \middle| \boldsymbol{\lambda}\right) d \mathbf{w}.
\end{equation}
The hyperparameters that maximizes the marginal log-likelihood are found analytically equaling the derivatives to zero,
\begin{equation}
    \begin{split}
        \gamma^{(t + 1)}_i &= 1 - \lambda^{(t)}_i \Sigma_{i,i}^{(t + 1)} \\
        \lambda^{(t + 1)}_i &= \frac{\gamma^{(t + 1)}_i}{\mu^{(t + 1)}_i} \\
        {\sigma_n^2}^{(t + 1)} &= \frac{\| \mathbf{y} - \mathbf{K} \boldsymbol{\mu}^{(t + 1)} \|^2}{N - \sum^N_{i = N} \gamma^{(t + 1)}_i}
    \end{split}
\end{equation}
where $\gamma_i$ is the relevance measure of vector $i$. The optimization algorithm updates the parameters iteratively until reaches the convergence criterion.

The RVM characteristic sparseness arises when there is poor alignment between the direction of $\varphi (\mathbf{x}_i)$ and $y_i$, then $\lambda_i \rightarrow \infty$ and consequently the parameter $w_i$ posterior distribution mean and variance will tend to zero \cite{BISHOP2006}. This causes the vector $\varphi (\mathbf{x}_i)$ to be removed from the model. The remaining of the feature vectors $\varphi (\mathbf{x}_i)$ with non-zero posterior mean and variance are the so-called relevance vectors \cite{Faul2001}.

A prediction for a new sample $\mathbf{x}_*$ is obtained as,
\begin{equation}
    \begin{split}
        \hat{y}_* &= \sum_{i = 1}^N \mu_i \mathcal{K} \left( \mathbf{x}_i, \mathbf{x}_* \right), \\
        \hat{\sigma}^2_* &= \sigma^2_n + \mathcal{K} \left( \mathbf{x}_i, \mathbf{x}_* \right)^\top \boldsymbol{\Sigma} \mathcal{K} \left( \mathbf{x}_i, \mathbf{x}_* \right).
    \end{split}
\end{equation}
The mechanism of sparsity can be exploited to derive a faster optimization of hyperparameters \cite{Tipping2003}. However, that method was not implemented in this investigation due to our low number of samples.

\paragraph{Multi-Task Relevance Vector Machine for Regression (MT-RVM)}\label{sec:multitask_relevance_vector_machine}

This model is implemented converting the predictors matrix into a vector form $\tilde{\mathbf{y}}$, and extending the model parameters using the Kronecker product, in an analogous manner that previous multi-task models, 
\begin{equation}
    \begin{split}
        \tilde{\boldsymbol{\Sigma}}^{(t + 1)} &= \left( \boldsymbol{\Lambda}^{(t)} \otimes \mathbf{I} + {\sigma_n^{-2}}^{(t)} \tilde{\mathbf{K}}^\top \tilde{\mathbf{K}} \right)^{-1}, \\
        \tilde{\boldsymbol{\mu}}^{(t + 1)} &= {\sigma_n^{-2}}^{(t)} \tilde{\boldsymbol{\Sigma}}^{(t + 1)} \tilde{\mathbf{K}}^\top \tilde{\mathbf{y}},
    \end{split}
\end{equation}
where $\tilde{\mathbf{K}} = \boldsymbol{\Gamma} \otimes \mathbf{K}$. The extension of the parameters that maximized the log-likelihood to a multi-task model are,
\begin{equation}
    \begin{split}
        \gamma^{(t + 1)}_i &= 1 - \lambda^{(t)}_i \tilde{\Sigma}^{(t + 1)}_{i,i}, \\
        \lambda^{(t + 1)}_i &= \frac{\gamma^{(t + 1)}_i}{\tilde{\mu}^{(t + 1)}_i}, \\
        {\sigma_n^2}^{(t + 1)} &= \frac{\| \tilde{\mathbf{y}} - \tilde{\mathbf{K}} \tilde{\boldsymbol{\mu}}^{(t + 1)} \|^2}{CN - \sum^{CN}_{i = 1} \gamma^{(t + 1)}_i},
    \end{split}
\end{equation}
where $C$ is the number of outputs in the forecasting model.

A prediction of a new multi-task observation in a MT-RVM is,
\begin{equation}\label{eq:multitask_relevance_vector_machine_prediction}
    \begin{split}
        \hat{\mathbf{y}}_* &= \left[ \boldsymbol{\Gamma} \otimes \mathcal{K} \left( \mathbf{x}_i, \mathbf{x}_* \right) \right] \tilde{\boldsymbol{\mu}}^\top, \\
        \hat{\boldsymbol{\sigma}}^2_* &= \sigma_n^2 + \left[ \boldsymbol{\Gamma} \otimes \mathcal{K} \left( \mathbf{x}_i, \mathbf{x}_* \right) \right]^\top \tilde{\boldsymbol{\Sigma}} \left[ \boldsymbol{\Gamma} \otimes \mathcal{K} \left( \mathbf{x}_i, \mathbf{x}_* \right) \right].      
    \end{split}
\end{equation}
The optimization yields to a sparse solution as well, so similarly the extended posterior mean and variance of the non-relevant vectors tends to zero.

\subsection{Kernel Regression Chain}

Another type of multi-task model proposed in this investigation are the regression chains \cite{garcia2021}. The models in a chain are arranged in a chronological sequence \cite{Melki2017}, so that the prediction of a model is added to the feature vectors of following models,
\begin{equation}
    \hat{\mathbf{y}}_* =     
    \begin{bmatrix}
        \hat{y}_{1*} \\
        \hat{y}_{2*} \\
        \hat{y}_{3*} \\
        \vdots \\
        \hat{y}_{C*} \\
    \end{bmatrix} =
    \begin{bmatrix}
        \mathbf{w}_1^\top \varphi \left( \mathbf{x}_* \right) \\ \mathbf{w}_2^\top
        \varphi \left( \mathbf{x}_*, \hat{y}_{1*} \right) \\
        \mathbf{w}_3^\top \varphi \left( \mathbf{x}_*, \hat{y}_{1*}, \hat{y}_{2*}\right)  \\
        \vdots \\
        \mathbf{w}_C^\top \varphi \left( \mathbf{x}_*, \hat{y}_{1*}, \hat{y}_{2*}, \dots, \hat{y}_{C* - 1} \right) 
    \end{bmatrix},
\end{equation}
where $C$ is the number of sequential forecasting horizons in the regression chain. 

In this investigation, the inference of the model parameters and the kernel hyperparameters is performed independently for each model in the chain. A regression chain is implemented for each one of the explained kernel methods.

\subsection{Multi-Task Kernel Simplification}\label{sec:multi-task Kernel Simplification}

We propose to model the process $\dots, y_{k - 1}, y_k, y_{k + 1}, \dots$ as an autoregressive process. To do that, the matrix $\boldsymbol{\Gamma} \in \mathbb{R}^{C \times C}$ is defined to contain the correlation coefficients between each forecasting output $C$, and the correlation coefficients $\gamma_{i,j}$ require cross-validation. Unfortunately, the number of correlation coefficients $\gamma_{i,j}$ is large $C^2$, so the cross-validation procedure is generally intractable in kernel learning applications. For simplification, we propose to determine the correlation coefficients modelling them as a decaying autoregressive model, 
\begin{equation}
    \gamma_{i,j} = \exp \left( \frac{C - |i - j|}{C\ell} \right), \quad \forall i,j = 1, \dots, C,
\end{equation}
where $\ell$ is the length-scale and it is the only hyperparameter in $\boldsymbol{\Gamma}$ that requires cross-correlation. An advantage of this simplification is that the matrix $\boldsymbol{\Gamma}$ is symmetric (i.e., $\forall i,j, \gamma_{i,j} = \gamma_{j,i}$), so the computation cost is also reduced.

\section{Feature Extraction}\label{sec:feature_extraction}

The IR camera used in this research, is an uncooled microbolometer that measures the changes of temperature produced by the radiation emitted by a black body. The Wein's displacement law states that the black body radiation wavelength is inversely proportional to the temperature of the object. Consequently, we know that temperature of the clouds moving in the troposphere is within long-wave IR radiation. This is because the black body radiation spectrum for different temperatures have their maxima at different wavelengths. In this way, the radiometric measurements of the IR camera allow us to derive physical features related to the clouds thermal dynamics and motion using consecutive IR images.

\subsection{Cloud Infrared Images}\label{sec:cloud_infrared_images}

A pixel in an IR image is defined by a pair of Euclidean coordinates $i,j$, and has measured temperature $\mathbf{T} = \{ T_{i,j} \in \mathbb{R}^+ \mid \forall i = 1, \ldots, M, \ \forall j = 1, \ldots, N \}$ in centi-Kelvin. The IR images are sensitive to the irradiance emitted by the Sun and particles in atmosphere, in addition to the radiation emitted by dust and water stains in the outdoor camera window. These effects are removed from the IR images so the IR images only contains the radiation received by the clouds \cite{TERREN2020a}. 

The irradiance scattering produced by particles floating in the atmosphere and the solar direct irradiance are deterministic effects. Thus the effects of the irradiance are removed from the IR images using parametric models. The parameters of the Sun direct irradiance model are constant, but the parameters of the atmospheric scattered irradiance model are a function of the air temperature, the dew point, the day of the year, and the elevation and azimuth angles \cite{TERREN2021a}. The temperature of the pixels after removing the atmospheric scattered and the Sun direct irradiance are $\mathbf{T}^\prime = \{ T^\prime_{i,j} \in \mathbb{R}^+ \mid \ \forall i = 1, \ldots, M, \ \forall j = 1, \ldots, N \}$.

The radiation emitted by dust and water stains in the outdoor window of IR camera is also removed from the IR images. This radiation is approximated as the median image of the set that contains the last $L = 250$ clear sky IR images \cite{TERREN2021a}. The set of clear sky images is updated every time that a sky condition model detects  a IR image with clear sky. This model is capable of detecting the sky condition in IR image using the temperature of the pixels, the magnitude of the velocity vectors, the atmospheric pressure and the clear sky index. The sky conditions model accurately distinguished when an IR image displays clear sky or a cloud (such as cumulus, stratus or nimbus). The temperatures after removing the atmospheric scattered and the Sun direct irradiance, and the radiation emitted by particles in the camera outdoor window are defined as $\mathbf{T}^{\prime \prime} = \{ T_{i,j}^{\prime \prime} \in \mathbb{R} \mid \forall i = 1, \ldots, M, \ \forall j = 1, \ldots, N \}$.

The heights of the air parcels contained in the pixels are defined as $\mathbf{H} = \{ H_{i,j} \in \mathbb{R}^+ \mid \forall i = 1, \ldots, M, \ \forall j = 1, \ldots, N \}$. The heights of particles in the Troposphere can be approximated as $H_{i,j} = (T_{i,j}^{\prime \prime} - T^{air}) / \Gamma_{MARL}$, $\Gamma_{MARL}$ is the Moist Adiabatic Lapse Rate (MARL) \cite{Hess1959, Stone1979},  $T^{air}$ being the air temperature. $\Gamma_{MARL}$ is given by a function $\phi : (T^{air}, T^{dew}, P^{atm}) \mapsto \Gamma_{MARL}$ that depends on the air temperature $T^{air}$, the dew point $T^{dew}$ and the atmospheric pressure $P^{atm}$ measured on ground-level \cite{TERREN2021a}.

\subsection{Cloud Dynamics}\label{sec:cloud_dynamics}

The displacement of the pixels in consecutive IR images is analyzed to compute the cloud dynamics. The cloud velocity vectors are first computed for further extracting second order dynamics from them (i.e., vorticity and divergence). The following features extracted from the cloud dynamics, are included in the feature vectors used in the forecasting.

\begin{itemize}

\item Velocity Vectors

We define the cloud's velocity vector in the classic manner \cite{BOUCHER2013}, which is the rate of change of an object along time,
\begin{equation}
 \left( \frac{\partial x}{\partial t}, \frac{\partial y }{\partial t}, \frac{\partial z }{\partial t}\right) = \left( u, v, 0 \right),
\end{equation}
where $x$ and $y$ are the coordinate system of the camera plane, and $t$ is the time variable. Variable $z$ is the height of the cloud layer, that we assume that does not have vertical movement. The velocity vectors of the clouds are ${\bf U} = \{ \text{u}_{i,j} \in \mathbb{R} \mid \forall i = 1, \ldots, M, \ \forall j = 1, \ldots, N \}$ in the $x$ component, and ${\bf V} = \{ \text{v}_{i,j} \in \mathbb{R} \mid \forall i = 1, \ldots, M, \ \forall j = 1, \ldots, N \}$ in the $y$ component. The cloud velocity vectors are computed using a weighted implementation of the Lucas-Kanade algorithm \cite{TERREN2021c}. We propose to use the velocity vectors to extract second order features of the cloud dynamics, and their magnitude $m_{i,j} = ( u_{i,j}^2 + v_{i,j}^2)^{1/2}$ to estimate the time instant a cloud will intersect the Sun.

\item Divergence

The divergence ${\bf D} = \{ \text{d}_{i,j} \in \mathbb{R} \mid \forall i = 1, \ldots, M, \ \forall j = 1, \ldots, N \}$ of the clouds in the images, is a scalar field that describes how this is expanding or compressing at a given point. In the case of a 2-dimensions velocity field the divergence is calculated as the dot product between  operator $\nabla$ and $\mathbf{V}$,
\begin{equation}
    {\bf D} = \nabla \cdot \mathbf{V} = 
        \left(
        \begin{matrix}
            \frac{\partial}{\partial x} & \frac{\partial}{\partial y} & \frac{\partial}{\partial z} \\
        \end{matrix}
        \right)
    \cdot 
        \left(
        \begin{array}{c}
            u \\
            v \\
            0 \\
        \end{array}
        \right) 
    = \left( \frac{\partial u}{\partial x} + \frac{\partial v}{\partial y} \right).
\end{equation}

\item Vorticity

The vorticity or curl ${\bf C} = \{ \text{c}_{i,j} \in \mathbb{R} \mid \forall i = 1, \ldots, M, \ \forall j = 1, \ldots, N \}$, is a pseudo-vector field that describes the rotation of the velocity field $\mathbf{V}$. The curl in an image is obtained taking the cross-product between the operator $\nabla$ and the velocity field $\mathbf{V}$. In a 2-dimensions space can be calculated as
\begin{equation}
    {\bf C} = \nabla \times \mathbf{V} =
    \begin{vmatrix}
        \mathbf{\hat{i}} & \mathbf{\hat{j}} & \mathbf{\hat{k}} \\
        \frac{\partial}{\partial x} & \frac{\partial}{\partial y} & \frac{\partial}{\partial z} \\
        u & v & 0 \\
    \end{vmatrix}
     = \left( \frac{\partial u}{\partial x} - \frac{\partial v}{\partial y} \right)
    \mathbf{\hat{k}}.
\end{equation}

\end{itemize}

\subsection{Feature Selection}\label{sec:features_selection}

The wind velocity field in the atmosphere cross-section plane of a cloud layer $\hat{\mathbf{V}} = \{ ( \hat{u}_{i,j} , \hat{v}_{i,j} ) \in \mathbb{R}^2 \ | \ \forall i = 1, \dots, M, \ \forall j = 1, \dots, M  \}$ is approximated using a wind flow visualization algorithm \cite{TERREN2020b}. Assuming that the air parcel in the IR image is sufficient small to consider the wind velocity field incompressible and irrotational, the streamlines are equivalent to pathlines. The streamlines and potential lines are computed using the approximated wind velocity field $\hat{\mathbf{V}}$, and are  $\boldsymbol{\Phi} = \{ \phi_{i,j} \in \mathbb{R} \ | \ \forall i = 1, \dots, M, \ \forall j = 1, \dots, M \}$ and $\boldsymbol{\Psi} = \{ \psi_{i,j} \in \mathbb{R} \ | \ \forall i = 1, \dots, M, \ \forall j = 1, \dots, M \}$ respectively.

\subsubsection{Sun Intersecting Streamline}

The Sun intersecting streamline in the potential direction can be found following an iterative connected component scheme. The proposed scheme begins at the position of the Sun $x_0 = \{i_0, j_0\}$, so that $i^{(1)} = i_0$ and $j^{(1)} = j_0$, and follows this iterative position update
\begin{equation}
    \begin{split}
        i^{(k + 1)}, j^{(k + 1)} &= \underset{i,j}{\operatorname{argmin}} \ \left( \boldsymbol{\Phi}_{i,j \in \mathcal{C}} - \boldsymbol{\Phi}_{i^{(k)}, j^{(k)}} \right)^2, \\
        \mathrm{s.t.} &
        \begin{cases}
            \boldsymbol{\Psi}_{i^{(k + 1)}, j^{(k + 1)}} > \boldsymbol{\Psi}_{i_0, j_0} \\
            i^{(k + 1)} \neq i^{(k)} \land j^{(k + 1)} \neq j^{(k)}, i^{(k + 1)},j^{(k + 1)} \in \mathcal{C},
        \end{cases}
    \end{split}
\end{equation}
where $\mathcal{C}$ is the set of pixels in the neighborhood of $i^{(k)}, j^{(k)}$, which is defined as $\mathcal{C} = \{ ( i^{(k)} + m, j^{(k)} + n ) \notin \ m = 0 \land n = 0 \mid \forall m, n = -1, 0, 1 \}$. The optimization scheme continues while $0 < i^{(k + 1)} < M \land 0 < j^{(k + 1)} < N$, when the streamlines connects with a pixel in the edge of the image, the algorithm stops. Therefore, the intersecting streamline $\mathcal{S} = \{ (i^{(\ell)}, j^{(\ell)} ) \ | \ \forall \ell = 1, \ldots, L \}$ is the set of connected pixels from the Sun to the edge of the image.

The above constrains have to be applied to the problem in the light of finding a feasible intersecting streamline. First, the flow potential direction has to be greater than the potential at the position of the Sun $\boldsymbol{\Psi}_{i_0, j_0} $. Second, non-pixel coordinates can be repeated in a streamline.

\subsubsection{Probability of a Pixel Intersecting the Sun}

The space coordinates in an image are distorted by the perspective of the camera, which depends on the altitude where the cloud layer is flowing. 

The Euclidean coordinates system of the IR sensor plane is reprojected to the atmosphere cross-section plane of the cloud layer that is distorted by the perspetive in the image \cite{TERREN2021b}. The geospatial perspective reprojection is function $\psi : (i,j;\varepsilon, \alpha, h) \mapsto \mathbf{x}_{i,j}, \Delta \mathbf{x}_{i,j}$ that maps the Euclidean coordinates to the atmosphere cross-section plane. The function  depends on the Sun elevation $\varepsilon$ and azimuth $\alpha$ angles, and the height of the cloud layer $h$. The function maps the coordinate system $i,j$ to the atmosphere cross-section plane $\mathbf{X} = \{ ( x_{i,j} , y_{i,j} ) \in \mathbb{R}^2 \ | \ \forall i = 1, \dots, M, \ \forall j = 1, \dots, M \}$, and to the dimensions of the atmosphere cross-section plane contained in a pixel $\Delta \mathbf{X} = \{ ( \Delta x_{i,j} , \Delta y_{i,j} ) \in \mathbb{R}^2 \ | \ \forall i = 1, \dots, M, \ \forall j = 1, \dots, M \}$. The units of velocity vectors are in meters per second, and the space units are in meters.

Therefore, the pixels in the streamline $\mathcal{S}$ have a pair of space coordinates $\mathbf{X}^\prime = \{ ( {x^\prime}^{(1)}, {y^\prime}^{(1)} ), \ldots, ( {x^\prime}^{(L)}, {y^\prime}^{(L)} ) \in \mathbb{R}^2 \mid \ \forall \ell = 1, \ldots, L \}$ in the atmosphere cross-section plane. The dimension of the pixels in the atmosphere cross-section plane are $\Delta \mathbf{X}^\prime = \{ ( \Delta {x^\prime}^{(1)} , \Delta {y^\prime}^{(1)} ), \ldots, ( \Delta {x^\prime}^{(L)},  \Delta {y^\prime}^{(L)} ) \in \mathbb{R}^2 \mid \ \forall \ell = 1, \ldots, L \}$, and their respective wind velocity field are $\hat{\mathbf{V}} = \{ ( \hat{u}^{(1)} , \hat{v}^{(1)} ), \ldots, ( \hat{u}^{(L)} , \hat{v}^{(L)} ) \in \mathbb{R}^2 \ | \ \forall \ell = 1, \ldots, L \}$. 

Once we know the sequential order of the pixels in the intercepting streamline, their dimensions, and their velocity field components, we want to estimate the time $t^{(\ell)}$ when a pixel will intersect the Sun. However, the estimation of $t^{(\ell)}$ is not certain because the approximation of a velocity vector in the wind velocity field has an error. The root mean squared error represent the uncertainty of a wind velocity vector in the $u$ and $v$ velocity component is $\mathbf{e} = \{ e_u, e_v \}$. We have that the intersecting time $t^{(\ell)}$ of a pixel in the streamline is given by the classical equation with plus an uncertainty,
\begin{equation}
    t^{(\ell)} + e_t = \left[ \frac{ a_x^{(\ell)} \left( \Delta x^{\prime(\ell)} \right)^2 + a_y^{(\ell)} \left( \Delta y^{\prime(\ell)} \right)^2 }{\left( \hat{u}^{(\ell)} + e_u \right)^2 + \left( \hat{v}^{(\ell)} + e_v \right)^2} \right]^{1/2}, \ \forall \ell = 1, \dots, L,
\end{equation}
where $a^{(\ell)}_x$ and $a^{(\ell)}_y$ are coefficients such as $a^{(\ell)}_x, a^{(\ell)}_y \in \{0, 1\}$. They value is 1, if the transition to a neighboring pixel in connected component algorithm involves a translation in the x-axis ($a^{(\ell)}_x$) or y-axis ($a^{(\ell)}_y$). Otherwise, the coefficient value is 0.

We draw $N$ independent samples from $e_{u,i} \sim \mathcal{N} (0, e_u)$ and $e_{v,i} \sim \mathcal{N} (0, e_v)$ to infer the distribution of $e_{t}$. If the intercepting the times $t^{(\ell)}$ along the streamlines are considered gamma random variables $t^{(\ell)} \sim \mathcal{G} (\alpha^{(\ell)}, \beta^{(\ell)})$, the maximum likelihood estimation of gamma distribution parameters can be numerically approximated as \cite{minka2002},
\begin{equation}
    \alpha^{(\ell)} \approx \frac{0.5}{ \left( \log \frac{1}{N} \sum_{i = 1}^N t^{(\ell)}_i \right) \left( \frac{1}{N} \sum_{i = 1}^N \log t_i^{(\ell)} \right)}, \quad
    \beta^{(\ell)} = \frac{\frac{1}{N}\sum_{i = 1}^N t^{(\ell)}_i}{\alpha^{(\ell)}}.
\end{equation}
Consequently, the distributions when an air parcel will intersect the Sun is given by 
\begin{equation}\label{eq:cum_gamma}
    \hat{t}^{(\ell)} \sim \mathcal{G} \left( \sum^\ell_{\ell^\prime = 1} \alpha^{(\ell^\prime)}, \beta^{(\ell)} \right), \ \ell = 1, \dots, L,
\end{equation}
which is the sequential cumulative sum of gamma random variables with same rate $\beta$ parameter. The parameters $\beta^{(\ell)}$ are not the same in the approximated gamma distributions but they are very close to each other, so we approximated them as the same for all gamma distributions. 

Therefore, the probability $ p_{int}(t_c | \alpha^{(\ell)}, \beta^{(\ell)}) $ of a pixel in the streamline intersecting the Sun at a time $t_c$ is
\begin{equation}
    p_{int} \left(t_c \middle| \alpha^{(\ell)}, \beta^{(\ell)} \right) \triangleq w^{(\ell)}_{c} = \frac{{\beta^{(\ell)}}^{\alpha^{(\ell)}}}{\Gamma (\alpha)} t_c^{\alpha^{(\ell)} - 1} e^{- \beta^{(\ell)} t_c},
\end{equation}
where $t_c$ is the time ahead of a forecasting horizon $t_c = \{t_1, \dots, t_C\}$. Once the uncertainty along the intersecting position in the streamline is known for each forecasting horizon $t_c$, the uncertainty in the intersecting position are computed in the 2-dimensional coordinate system of the atmosphere cross-section plane,
\begin{equation}
    \begin{split}
        \mathbb{E}_{p_{int}(t_c)} \left[ \mathbf{x}^\prime \right] &= \bar{\mathbf{x}}_c \approx \sum_{\ell = 1}^L \mathbf{x}^{\prime(\ell)} w_c^{(\ell)} \Delta \bar{t}^{(\ell)}, \\
        \mathbb{C}_{p_{int}(t_c)} \left[ \mathbf{x}^\prime \right] &= \mathbf{S}_c \approx \sum_{\ell = 1}^L \mathbf{x}^{\prime(\ell)} {\mathbf{x}^{\prime(\ell)}}^\top w_c^{(\ell)} \Delta \bar{t}^{(\ell)} - \bar{\mathbf{x}}_c {\bar{\mathbf{x}}_c}^\top,
    \end{split}
\end{equation}
where $\Delta \bar{t}^{(\ell)} = \left( \bar{t}^{(\ell + 1)} - \bar{t}^{(\ell)} \right)$, and $\bar{t}^{(\ell)} = \alpha^{(\ell)}\beta^{(\ell)}$ is the expected time to traverse pixel $\ell$.

In addition to the uncertainty in the intersecting position in the streamline, it exists an uncertainty $e_{\Delta \mathbf{x}^\prime}$ in the displacement due to the uncertainty in the estimation of the wind velocity field. The uncertainty of the displacement in the 2-dimensional space of the atmosphere cross-section plane for each forecasting horizon $c$ is assumed a multivariate normally distributed random variable $e_{\Delta \mathbf{x}^\prime, c, i} \sim \mathcal{N} (\mathbf{0}, e_{\Delta \mathbf{x}^\prime, c} \mathbf{I})$. We proposed to estimated the uncertainty $e_{\Delta \mathbf{x}^\prime, c}$ numerically
\begin{equation}
    e_{\Delta \mathbf{x}^\prime, c} \approx  \frac{1}{N} \sum^N_{i = 1} t^2_c \left( e^2_{v,i} + e^2_{v,i} \right),
\end{equation}
drawing $N$ independent samples from $e_{u,i} \sim \mathcal{N} (0, e_u)$ and $e_{v,i} \sim \mathcal{N} (0, e_v)$.

As the uncertainty in the intersecting position and the uncertainty on the displacement are both assumed i.i.d. Normal random variables, the probability of a pixel intersecting the Sun is computed given by their sum. The sum of two i.i.d. Normal random variables is the Normal distribution obtained from the convolution of the two Normal distributions, which is
\begin{equation}
     p \left( \mathbf{x}_{i,j} \middle| \bar{\mathbf{x}}_c, \bar{\mathbf{S}}_c \right) \sim \mathcal{N} \left( \bar{\mathbf{x}}_c, \mathbf{S}_c + e_{\Delta \mathbf{x}^\prime, c} \mathbf{I} \right),
\end{equation}
and it is computed for each time $t_c$. Therefore, the probability for any pixel in an image $\mathbf{x}_{i,j}$ intersecting the Sun at a time $t_c$ is
\begin{equation}%\label{eq:intercepting_prob}
    p \left( \mathbf{x}_{i,j} \middle| \bar{\mathbf{x}}_c, \bar{\mathbf{S}}_c \right) = \frac{1}{( 2\pi )^{1/2} | \bar{\mathbf{S}}_c |^{1/2}}
    \exp \left[ - \frac{1}{2} \left( \mathbf{x}_{i,j} - \bar{\mathbf{x}}_c \right)^\top \bar{\mathbf{S}}_c^{-1} \left( \mathbf{x}_{i,j} - \bar{\mathbf{x}}_c \right) \right].
\end{equation}
This is because the gamma distribution parameters $\alpha^{(\ell)}$ in Eq.~\eqref{eq:cum_gamma} are sufficiently large, so the gamma distribution can be approximated by a normal distribution.

The probability of a pixel being in the intersecting streamline $p (\mathbf{x}_{i,j}| \boldsymbol{\Phi}, \boldsymbol{\Psi})$ is computed using the equations in \ref{sec:wavefunction_probability}. Finally, the probability of a pixel intersecting the Sun and being in the intersecting streamline is computed as
\begin{equation}\label{eq:intercepting_prob}
    z_{i,j,c}\triangleq p \left( \mathbf{x}_{i,j} \middle| \bar{\mathbf{x}}_c, \bar{\mathbf{S}}_c \right) \cdot p \left( \mathbf{x}_{i,j} \middle| \boldsymbol{\Phi}, \boldsymbol{\Psi}\right),
\end{equation}
this are the probabilities used to weight the features in each pixel. 

\subsubsection{Statistical Features and Data Structure}

The features of clouds are extracted in each IR image and their first and second statistical  sample moments are computed to quantify them. The sample moments of third (i.e., skewness) and fourth (i.e., kurtosis) order were explored but the experiments did not show any improvement in the prediction. The feature vector obtained after the statistical quantification, is added to the database. 

The features of the cloud are quantified applying more importance to the pixels that are more likely to intersect the Sun. The importance weights are the previously computed probabilities $z_{i,j,c}$. The weighted sample first moment is,
\begin{equation}
    m^X_{c} = \frac{\sum^M_{i = 1} \sum^N_{j = 1} z_{i,j,c} x_{i,j}}{\sum^M_{i = 1} \sum^N_{j = 1} z_{i,j,c}}.
\end{equation}
where $x_{i,j,c}$ represents the value of any feature $X$ in pixel $i,j$. In an analogous manner, the weighted second moment $s_{X,c}$ of any given feature $X$ for horizon $c$ is,
\begin{equation}
    s^X_{c} = \left[ \frac{\sum^M_{i = 1} \sum^N_{j = 1} z_{i,j,c} (x_{i,j} - m^X_{c} )^2}{\sum^M_{i = 1} \sum^N_{j = 1} z_{i,j,c}} \right]^{1/2},
\end{equation}
where $m_{X,c}$ is the weighted sample first moment of feature $X$ for horizon $c$.

The exogenous features extracted from the cloudy pixels in image $k$ are: temperature $T_{i,j}$, height $H_{i,j}$, cloud velocity vectors magnitude $M_{i,j}$, divergence $D_{i,j}$, and vorticity $C_{i,j}$. Their quantified statistics are the means and standard deviations,
\begin{equation}
        \mathbf{t}_k =        
        \begin{bmatrix}
            m_{T,1} &  s_{T,1} \\
            \vdots & \vdots \\
            m_{T,C} & s_{T,C} 
        \end{bmatrix}
        \mathbf{h}_k =     
        \begin{bmatrix}
            m_{H,1} &  s_{H,1} \\
            \vdots & \vdots \\
            m_{H,C} & s_{H,C} 
        \end{bmatrix}
        \mathbf{m}_k = 
        \begin{bmatrix}
            m_{M,1} &  s_{M,1} \\
            \vdots & \vdots \\
            m_{M,C} & s_{M,C} 
        \end{bmatrix};
        \textbf{d}_k =         
        \begin{bmatrix}
            m_{D,1} &  s_{D,1} \\
            \vdots & \vdots \\
            m_{D,C} & s_{D,C} 
        \end{bmatrix}
        \mathbf{c}_k =  
        \begin{bmatrix}
            m_{C,1} &  s_{C,1} \\
            \vdots & \vdots \\
            m_{C,C} & s_{C,C} 
        \end{bmatrix}       
\end{equation}

In addition to the statistics, the feature vector in the database include endogenous irradiance measurements,
\begin{equation}
    \mathbf{y}_k = \left[ y_{k - 1} ~\dots ~y_{k - p} \right]^\top \in \mathbb{R}^p, 
\end{equation}
$p$ is defined as the lag in the time series, and the Sun elevation and azimuth angles at the time when the IR image $k$ was recorded,
\begin{equation}
    \mathbf{a}_k = \left[ \varepsilon ~\alpha \right]^\top \in \mathbb{R}^{2}.
\end{equation}

The different features vectors are concatenated together, so that a feature vectors $k$ in the database is defined as,
\begin{equation}\label{eq:feature_vectors}
    \begin{split}
        \mathbf{x}_{c,k} &= [ ~\mathbf{y}_k  ~\mathbf{a}_k  ~\mathbf{t}_{1,k} \cdots ~\mathbf{t}_{C,k} ~\mathbf{h}_{1,k} \cdots ~\mathbf{h}_{C,k} \cdots \\
        &\quad \quad \quad \cdots ~\mathbf{m}_{1,k} \cdots ~\mathbf{m}_{C,k}
        ~\textbf{d}_{1,k} \cdots ~\textbf{d}_{C,k} ~\mathbf{c}_{1,k} \cdots ~\mathbf{c}_{C,k} ]^\top \in \mathbb{R}^d,   
    \end{split}
\end{equation}
where $d$ represents the number of dimensions of any vector $\mathbf{x}_k$.

The corresponding multi-task independent variable vector to be predicted in instant $k$ (composed of future CSI measurement) is,
\begin{equation}
    \mathbf{y}_{k + 1} = \left[ y_{1} ~\dots ~y_{C} \right]^\top\in \mathbb{R}^{C},
\end{equation}
this independent variable vector is also stored in the database.

Henceforth, $\mathbf{y}_k$ is considered a stationary stochastic process defined as $\mathbf{y}_k = \{ \mathbf{y}_k : k \in [1, \infty ) \} \in \mathbb{R}^C$, that we aim to model using the feature vector $\mathbf{x}_k$ formed by exogenous and endogenous variables. 

\section{Experiments}

\subsection{Study Area and Data Acquisition System}

The acquisition system is located on the roof area of the University of New Mexico (UNM) Electrical and Computer engineering Department in Albuquerque, New Mexico. Albuquerque is located at $1,620$m of altitude. The climate is arid semi-continental with precipitation more likely during the summer. In summer, the sky is clear or partly cloudy 80\% of the time. In the yearlong, approximately $170$ days are sunny ($<30$\% cloud coverage), and $110$ days are partly sunny ($\geq30$\% to $<80$\% cloud coverage). The average of historic minima temperatures is $268.71$K in winter and the average of historic maxima temperatures is $306.48$K in summer. The average rainfall and snowfall accounts for approximately $279.4$mm per year.

\begin{figure}[!htb]
    \begin{subfigure}{\linewidth}
        \centering
        \includegraphics[scale = 0.2, trim = {1cm, 1.5cm, 2cm, 1.5cm}, clip]{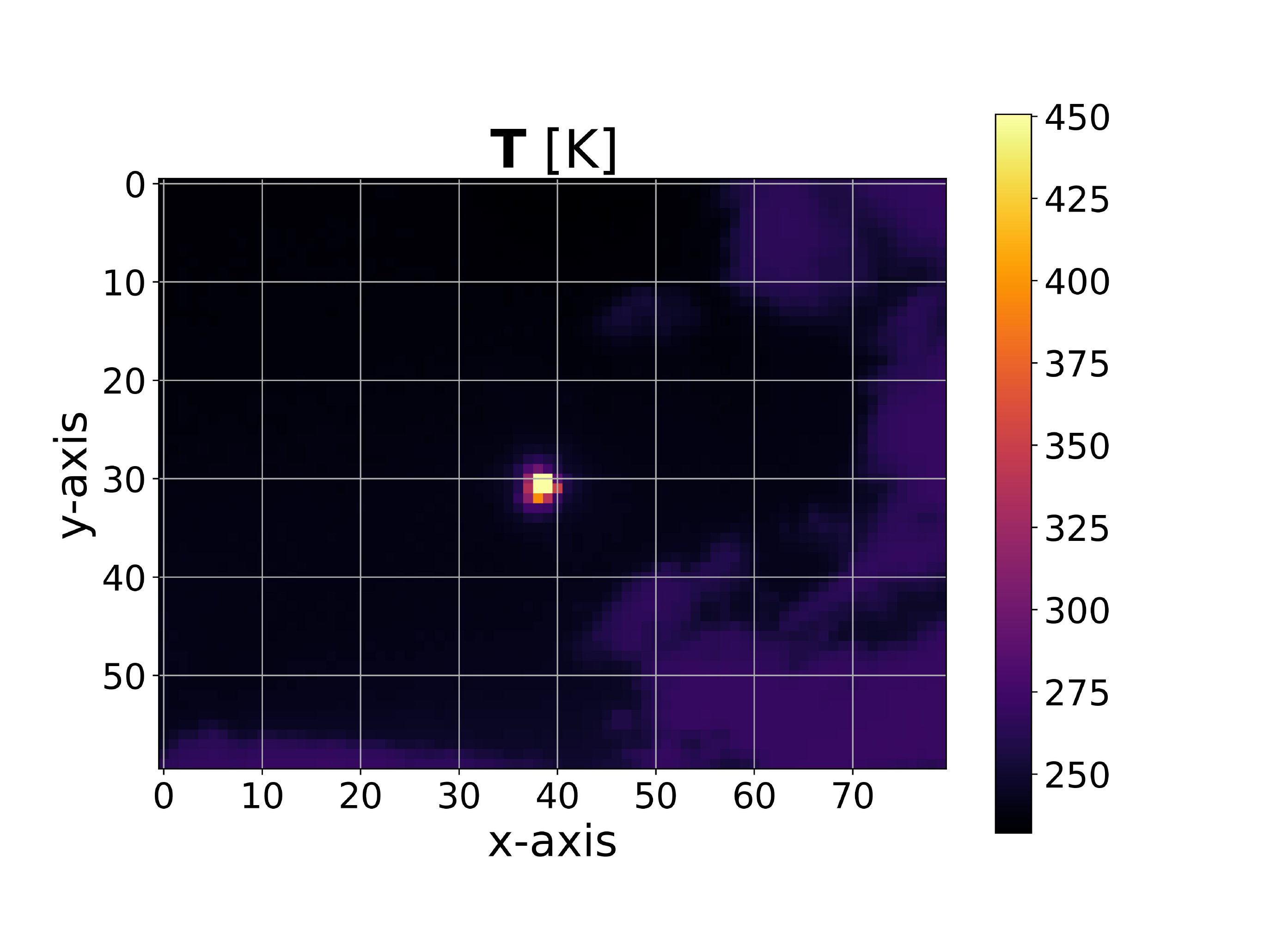}
        \includegraphics[scale = 0.2, trim = {1cm, 1.5cm, 2cm, 1.5cm}, clip]{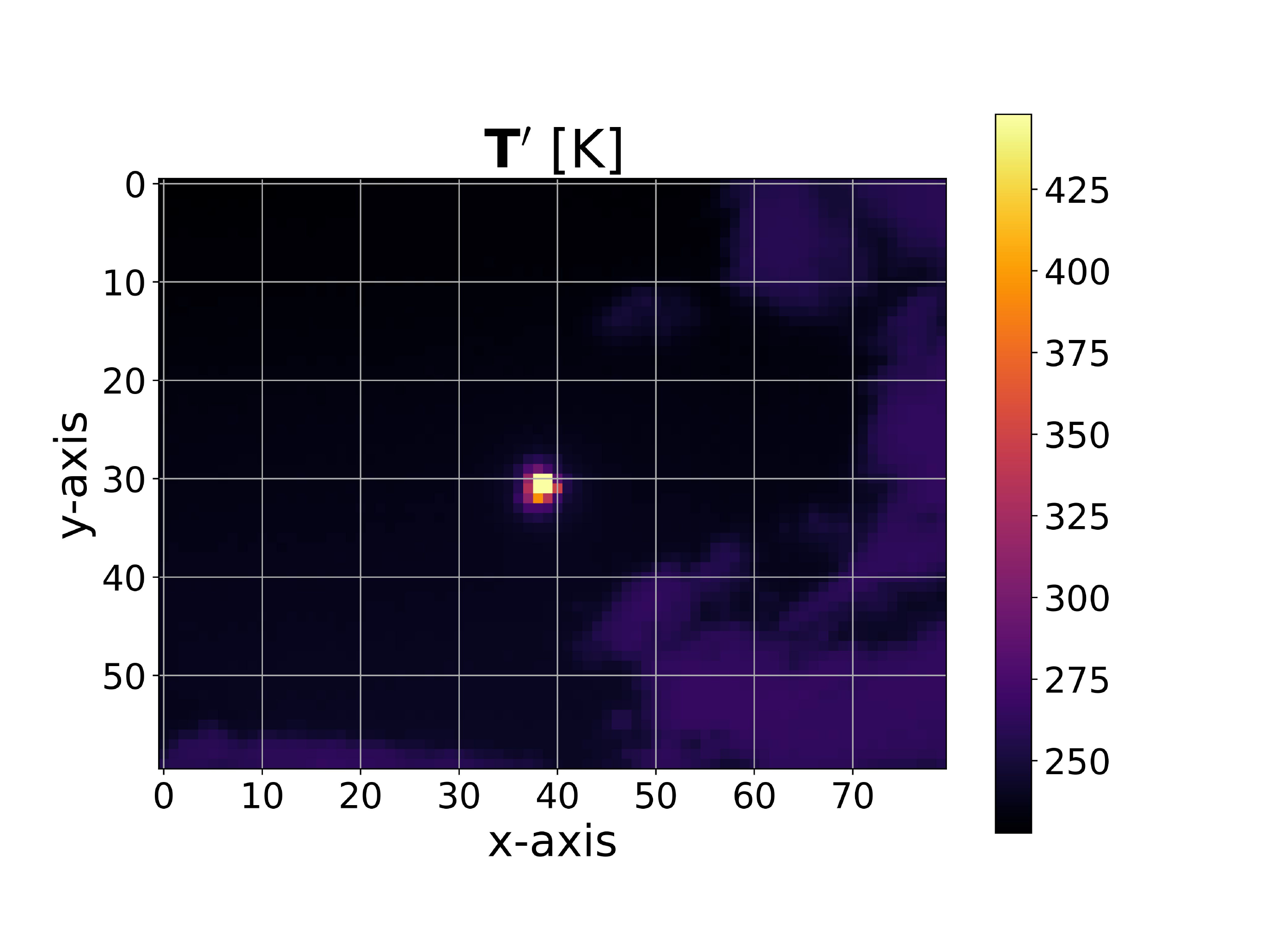}
        \includegraphics[scale = 0.2, trim = {1cm, 1.5cm, 2cm, 1.5cm}, clip]{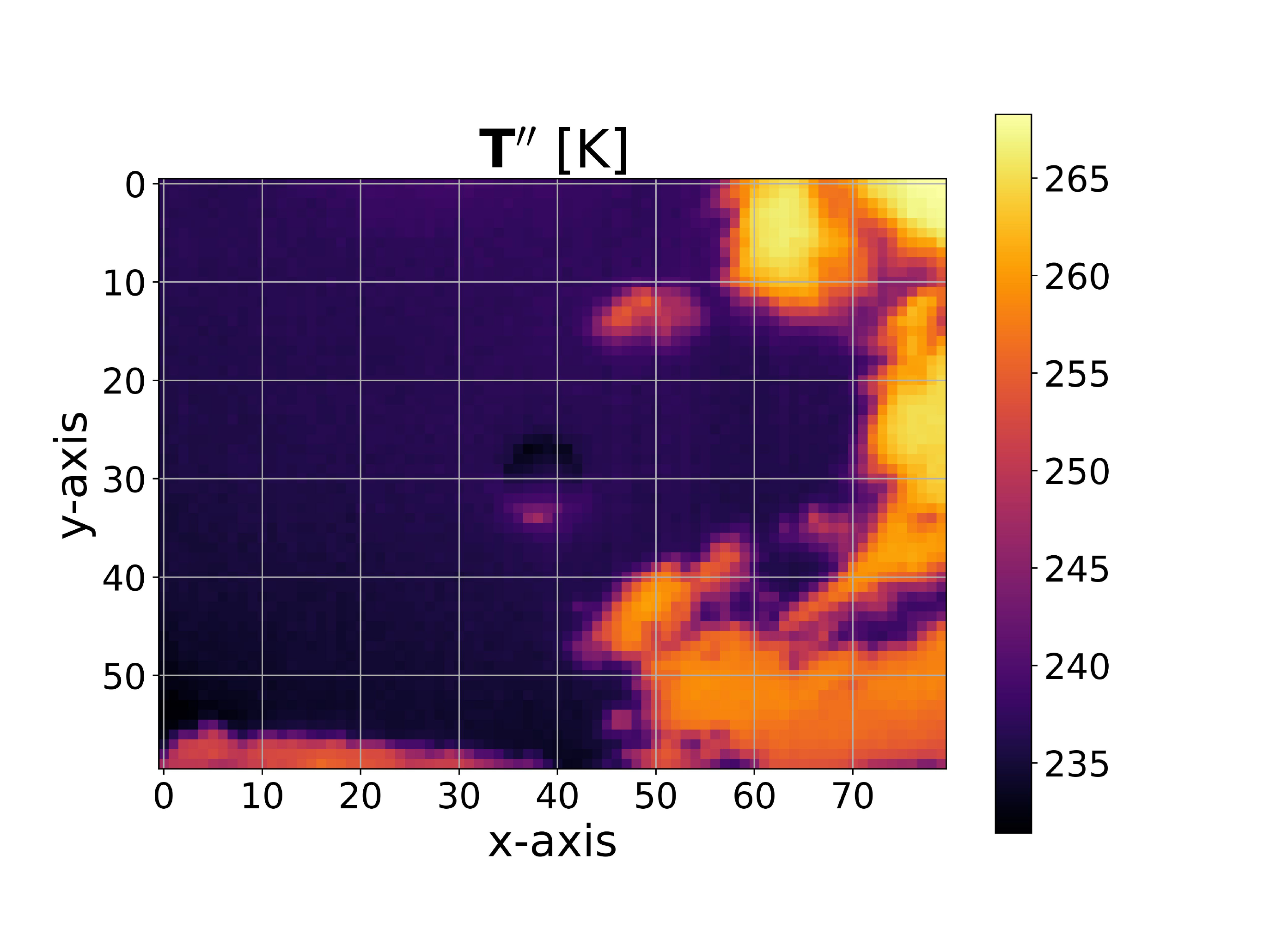}
    \end{subfigure}
    \begin{subfigure}{\linewidth}
        \centering
        \includegraphics[scale = 0.2, trim = {2.5cm, 0cm, 2.5cm, 0cm}, clip]{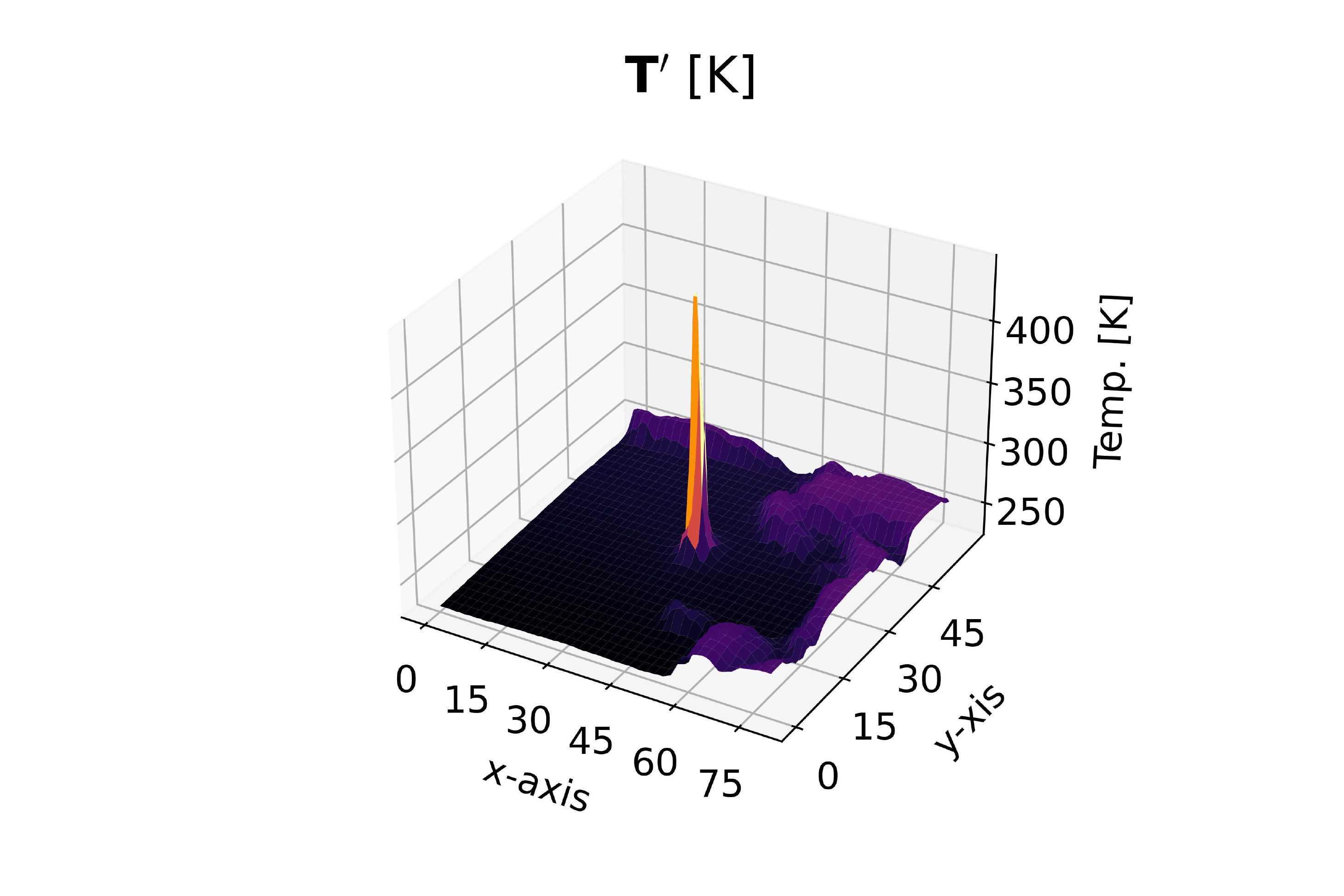}
        \includegraphics[scale = 0.2, trim = {2.5cm, 0cm, 2.5cm, 0cm}, clip]{images/516_dir_temp_3D.pdf}
        \includegraphics[scale = 0.2, trim = {2.5cm, 0cm, 2.5cm, 0cm}, clip]{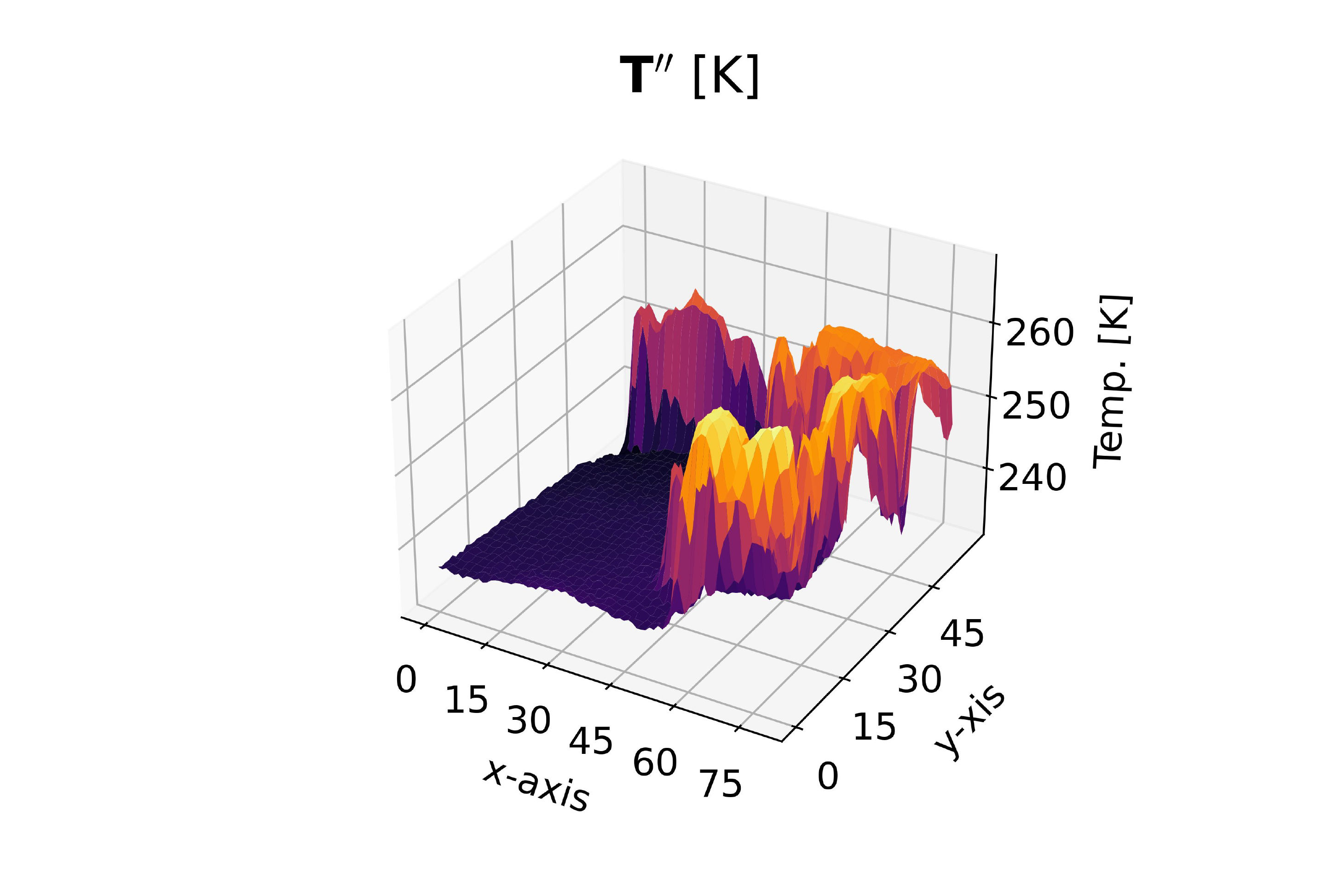}
    \end{subfigure}
    \caption{Illustration of the image processing workflow applied to the IR images. The top row shows the IR images in 2 dimensions. The bottom row shows these IR images in 3 dimensions. The IR images in the first column are not processed. The radiation effect produced by the germanium outdoor camera window is removed from the IR images in the middle column. The IR images in the left column show the result obtained after removing the effect of the Sun and atmospheric radiation. The radiation models applied to this IR image are displayed in Figure~\ref{fig:radiantion_effect_models}.}
    \label{fig:image_processing}
\end{figure}

The proposed acquisition system is composed of an IR sky imager, and a pyranometer situated in an horizontal surface. The IR sky imager is mounted on a solar tracker. The solar tracker is programmed to update its pan and tilt every second to maintain the Sun in the center of the images throughout the day. The IR camera is Lepton 2.5 witch shutter. This camera is a low-cost long-wave infrared radiometric camera manufactured by Lepton\footnote{\url{https://www.flir.com}}. The detection wavelength ranges from $8$ to $14$ $\mu$m. The resolution of the images is $80 \times 60$, and the diagonal FOV is $63.5^\circ$. The intensity of the pixels in an image are temperature measurements in centikelvin. The pyranometer is a LI-200R\footnote{\url{https://www.licor.com}}. The IR sky imager acquires an IR image every $15$ seconds, while the pyranometer makes $6$ measurements per second. The pyranometer measurements were averaged to match the sampling interval of the IR sky imager.

The weather features used in the image processing methods applied to remove cyclostationary effects and estimate the height of a cloud in the IR sky images are: atmospheric pressure, air temperature, dew point and humidity. The weather features were acquired by a weather station located at the helipad of the UNM Hospital (approx. $500$m apart from the sky imager). The sample interval of the weather station is $10$. The weather features were interpolated to match the sampling interval of the IR sky imager. 

\begin{figure}[!htb]
    \centering
    \includegraphics[scale = 0.3, trim = {2.5cm, 0.5cm, 2.5cm, 0.5cm}, clip]{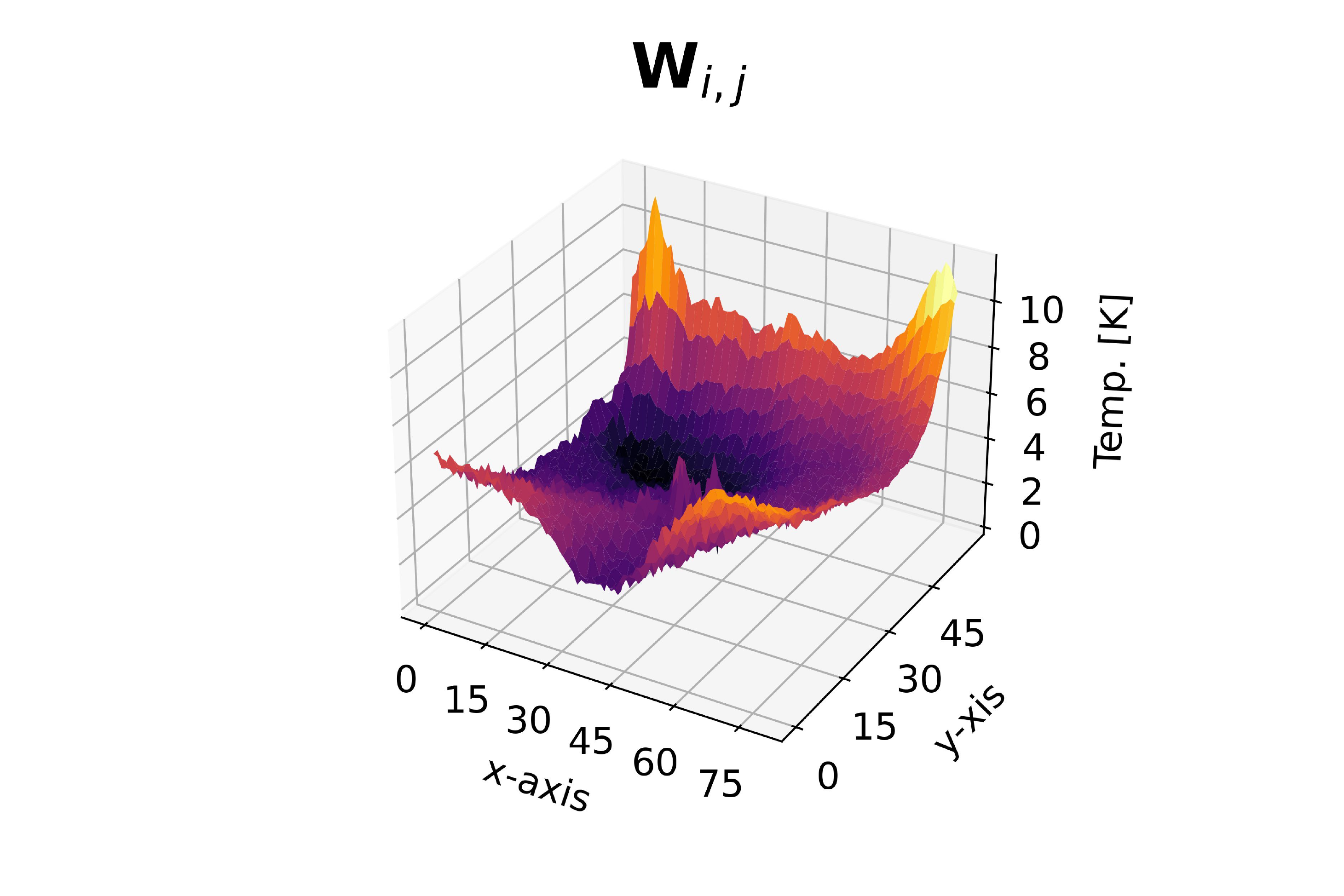}
    \includegraphics[scale = 0.3, trim = {2.5cm, 0.5cm, 2.5cm, 0.5cm}, clip]{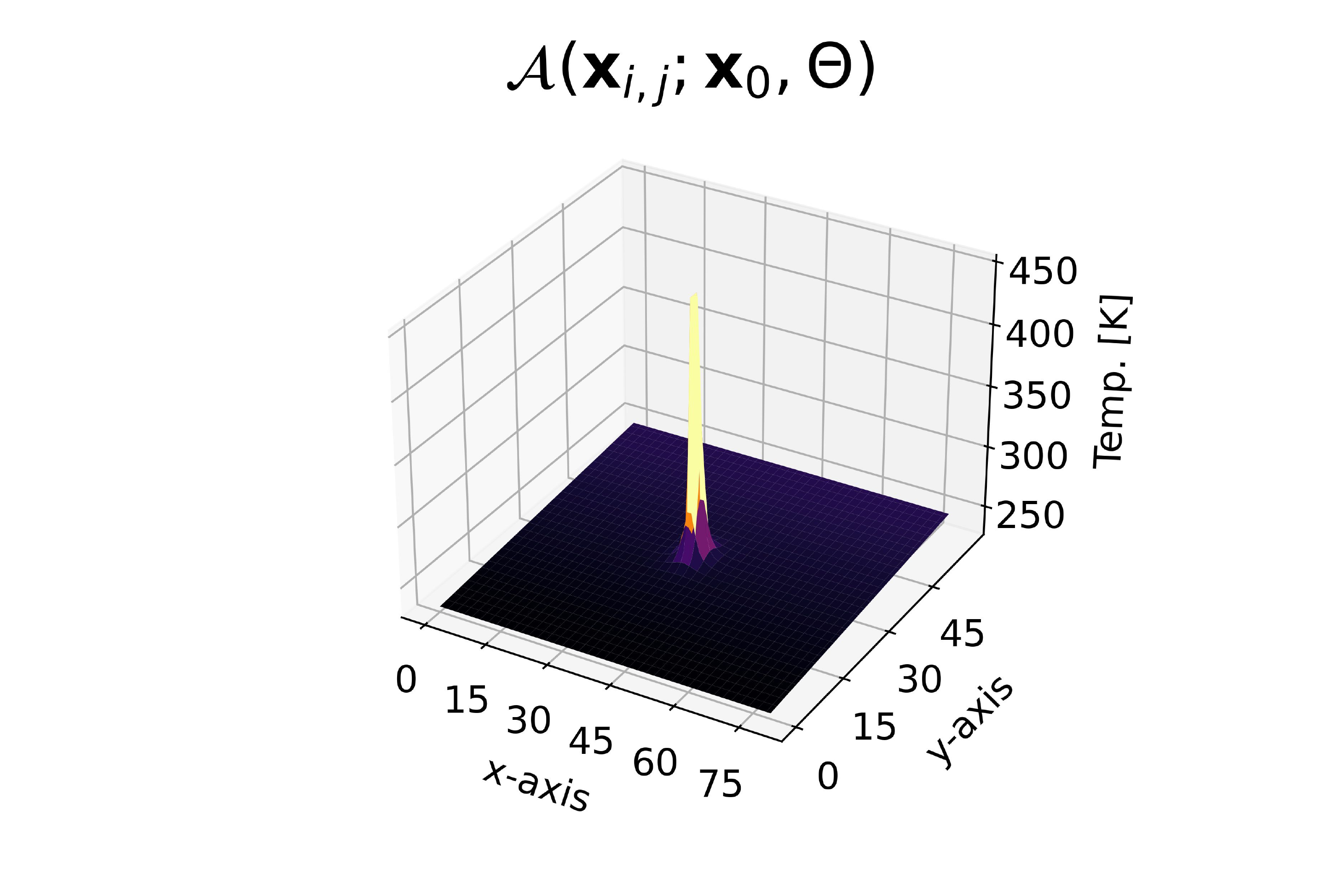}
    \caption{The graph on the left shows the persistent model of the radiation effect produced by the germanium outdoor camera window. The graph on the right shows the atmospheric model that combines the effect of the Sun's direct radiation and the radiation emitted by particles in the atmosphere. The results obtained after applying the radiation to IR image are displayed in Figure~\ref{fig:image_processing}.}
    \label{fig:radiantion_effect_models}
\end{figure}

\subsection{Image Processing, Feature Extraction and Selection}

\begin{figure}[!htb]
    \begin{subfigure}{\linewidth}
        \centering
        \includegraphics[scale = 0.2, trim = {1cm, 1.5cm, 2cm, 1.5cm}, clip]{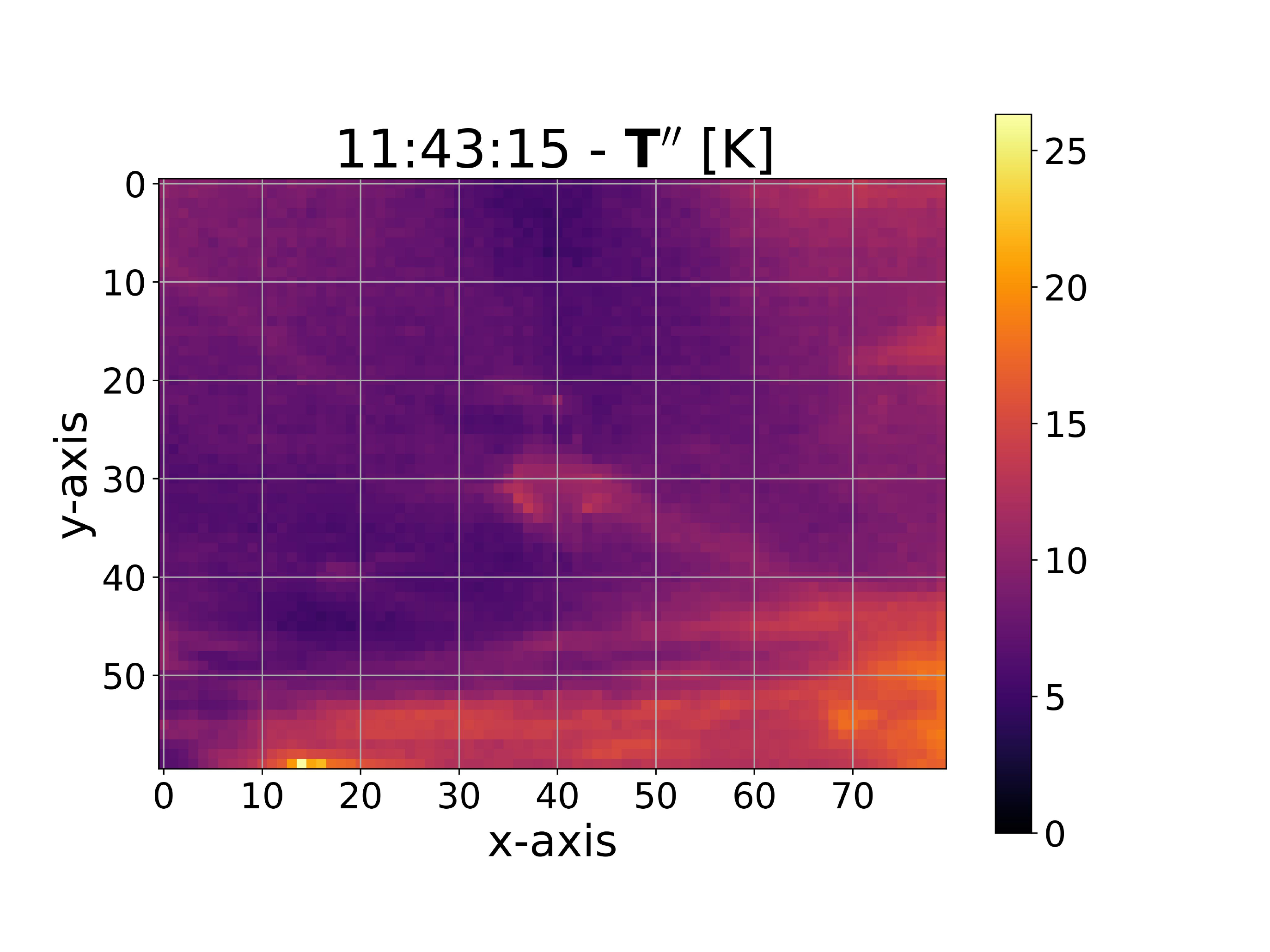}
        \includegraphics[scale = 0.2, trim = {1cm, 1.5cm, 2cm, 1.5cm}, clip]{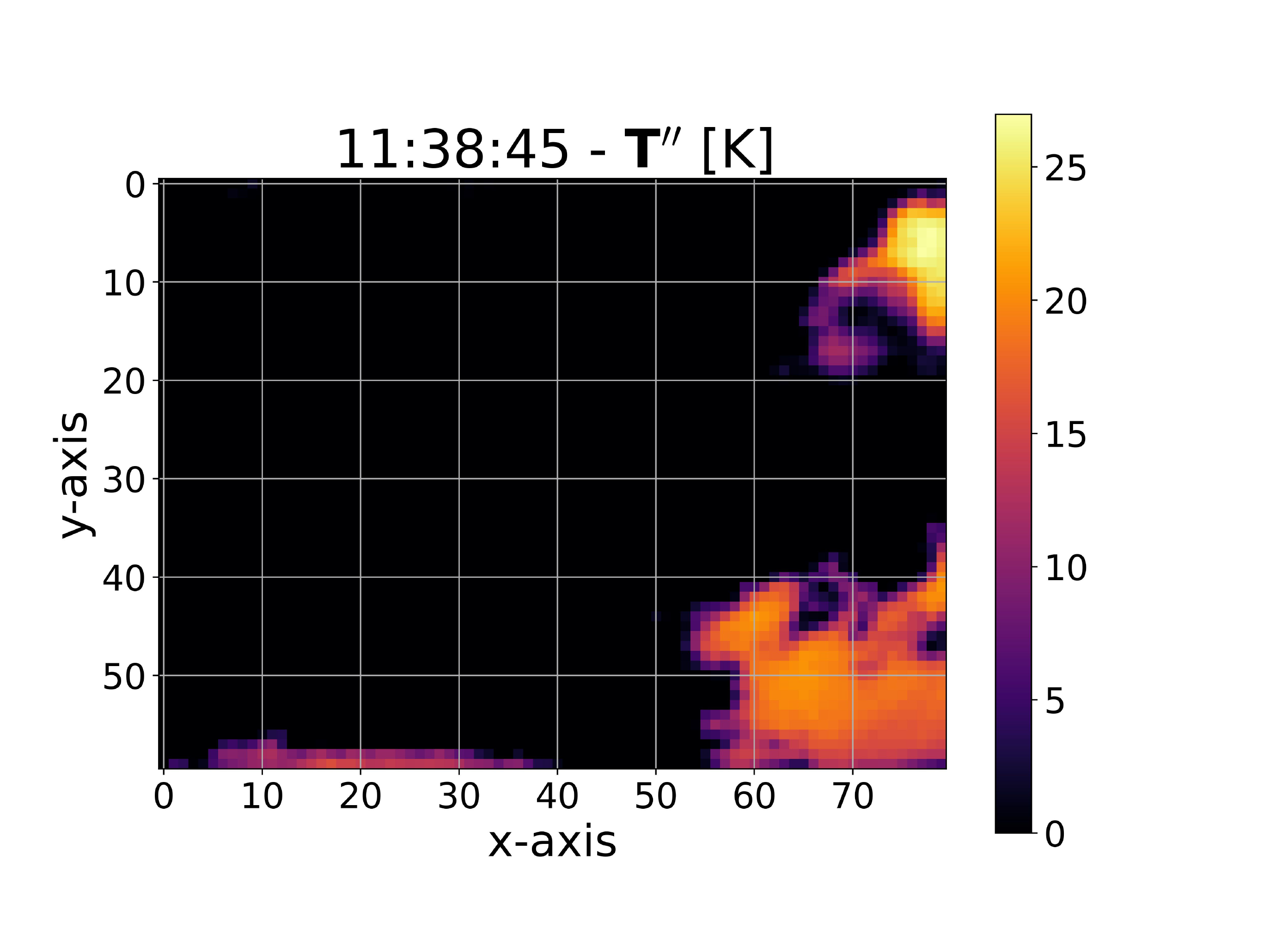}
    \end{subfigure}
    \begin{subfigure}{\linewidth}
        \centering
        \includegraphics[scale = 0.2, trim = {1cm, 1.5cm, 2cm, 1.5cm}, clip]{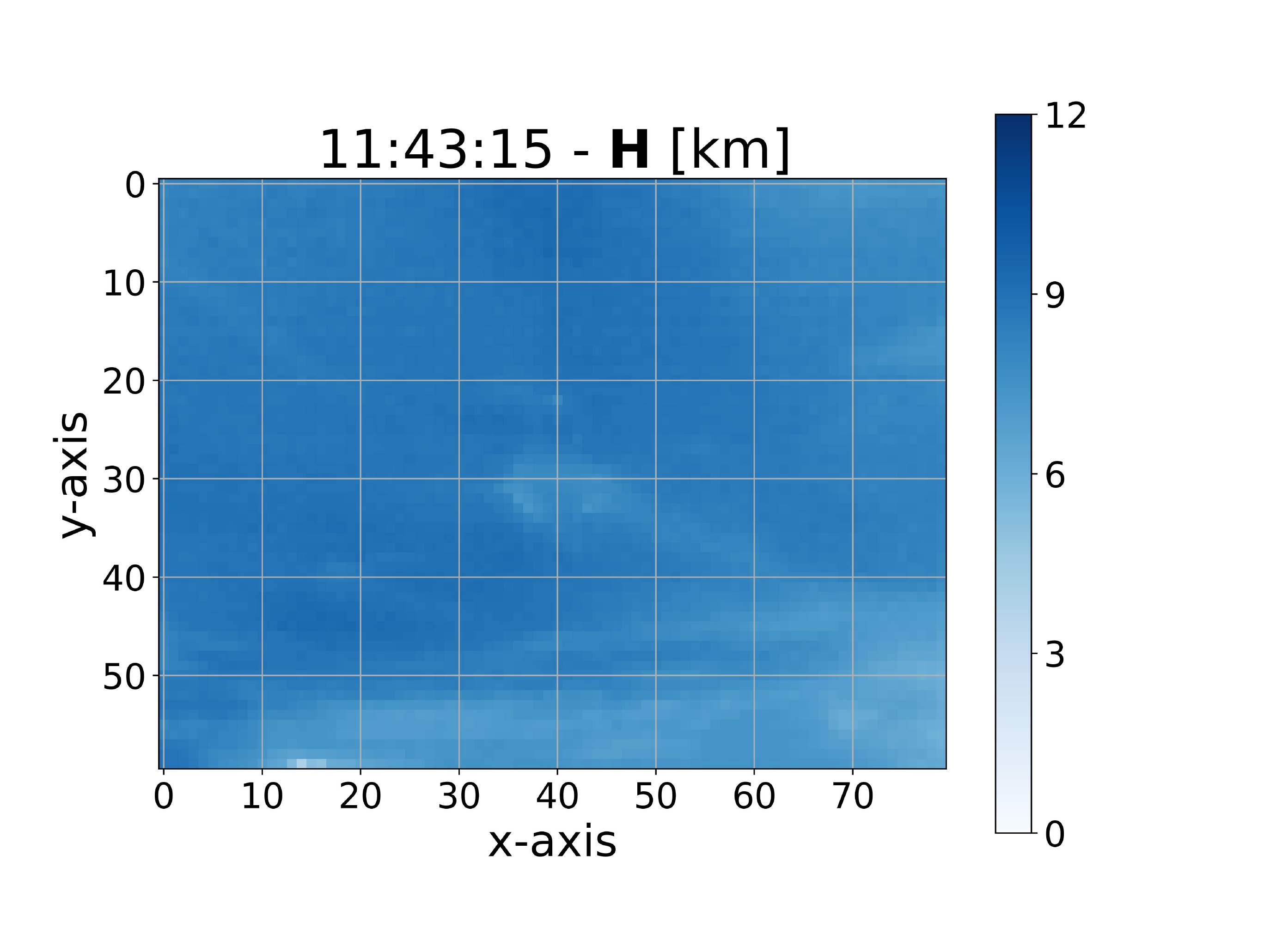}
        \includegraphics[scale = 0.2, trim = {1cm, 1.5cm, 2cm, 1.5cm}, clip]{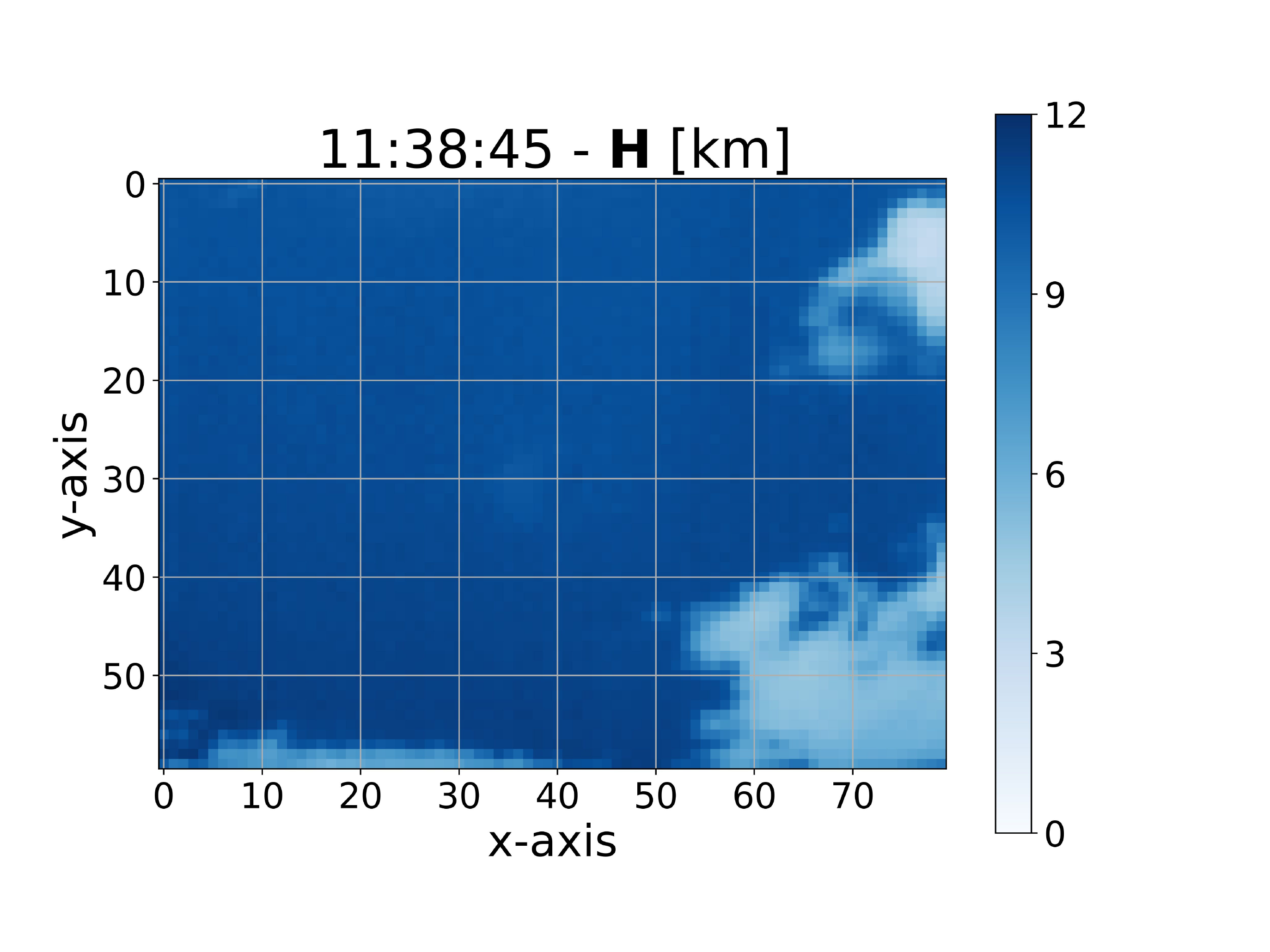}
    \end{subfigure}
    \begin{subfigure}{\linewidth}
        \centering
        \includegraphics[scale = 0.2, trim = {1cm, 1.5cm, 2cm, 1.5cm}, clip]{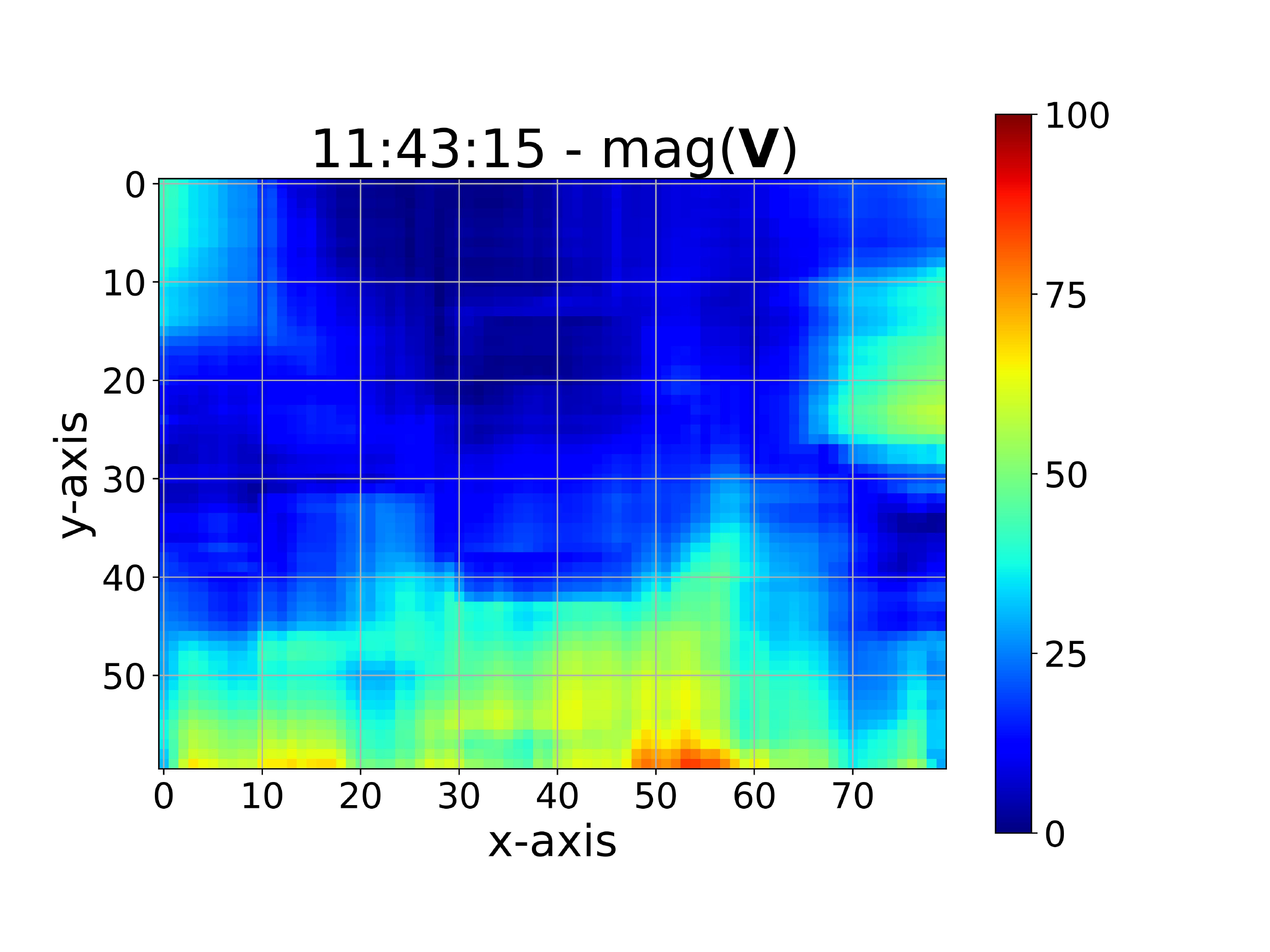}
        \includegraphics[scale = 0.2, trim = {1cm, 1.5cm, 2cm, 1.5cm}, clip]{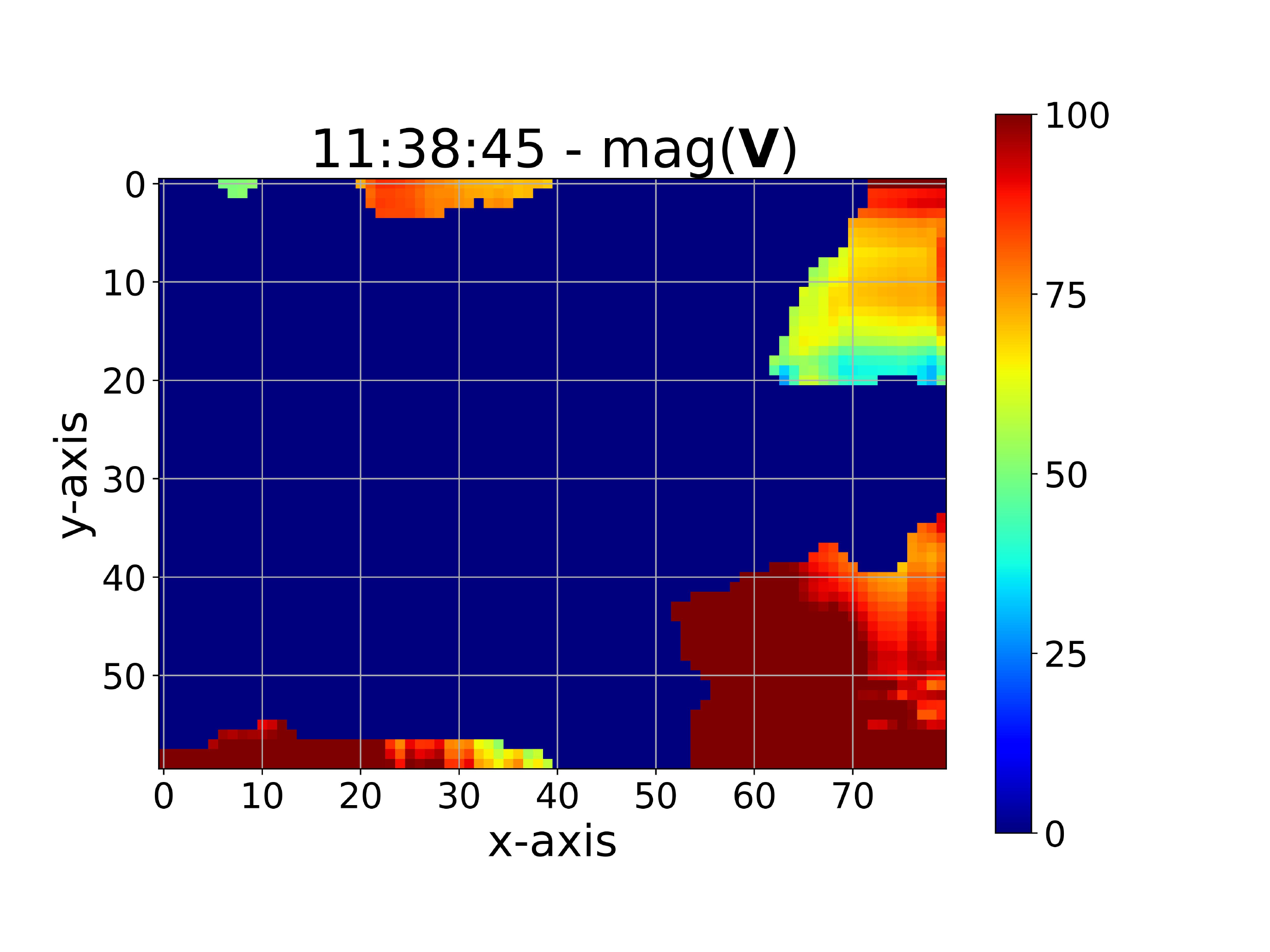}
    \end{subfigure}
    \begin{subfigure}{\linewidth}
        \centering
        \includegraphics[scale = 0.2, trim = {1cm, 1.5cm, 2cm, 1.5cm}, clip]{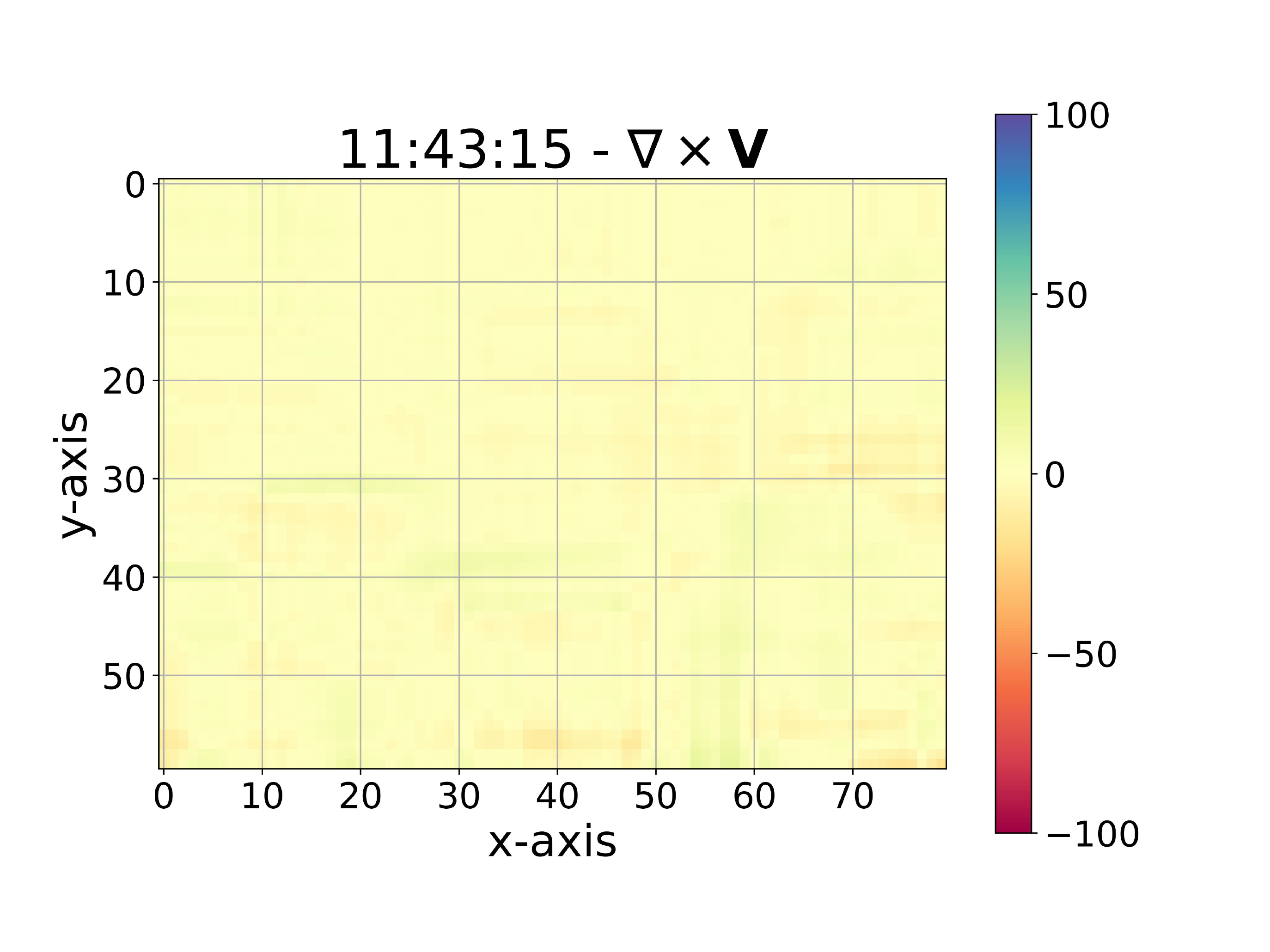}
        \includegraphics[scale = 0.2, trim = {1cm, 1.5cm, 2cm, 1.5cm}, clip]{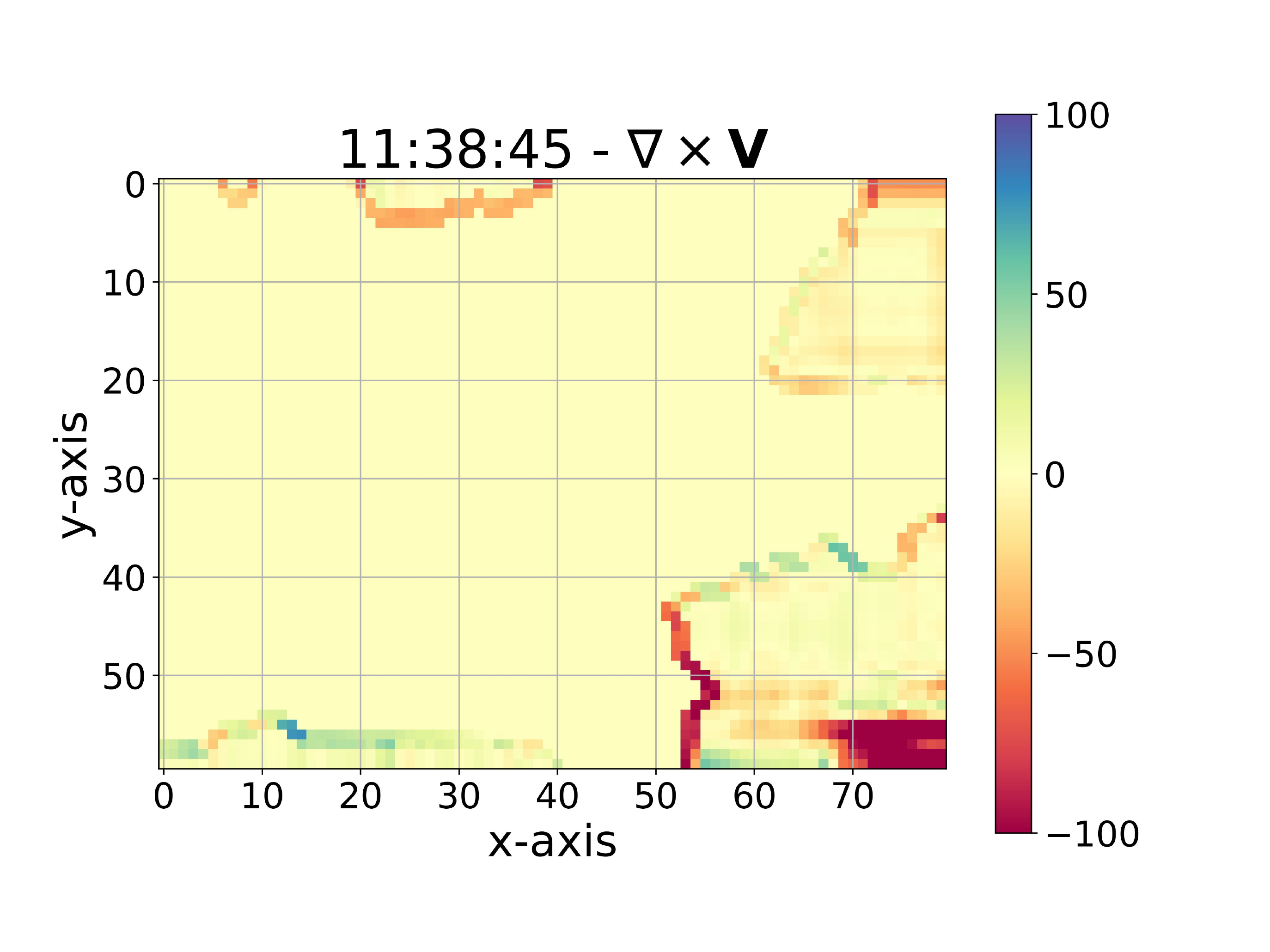}
    \end{subfigure}
    \begin{subfigure}{\linewidth}
        \centering
        \includegraphics[scale = 0.2, trim = {1cm, 1.5cm, 2cm, 1.5cm}, clip]{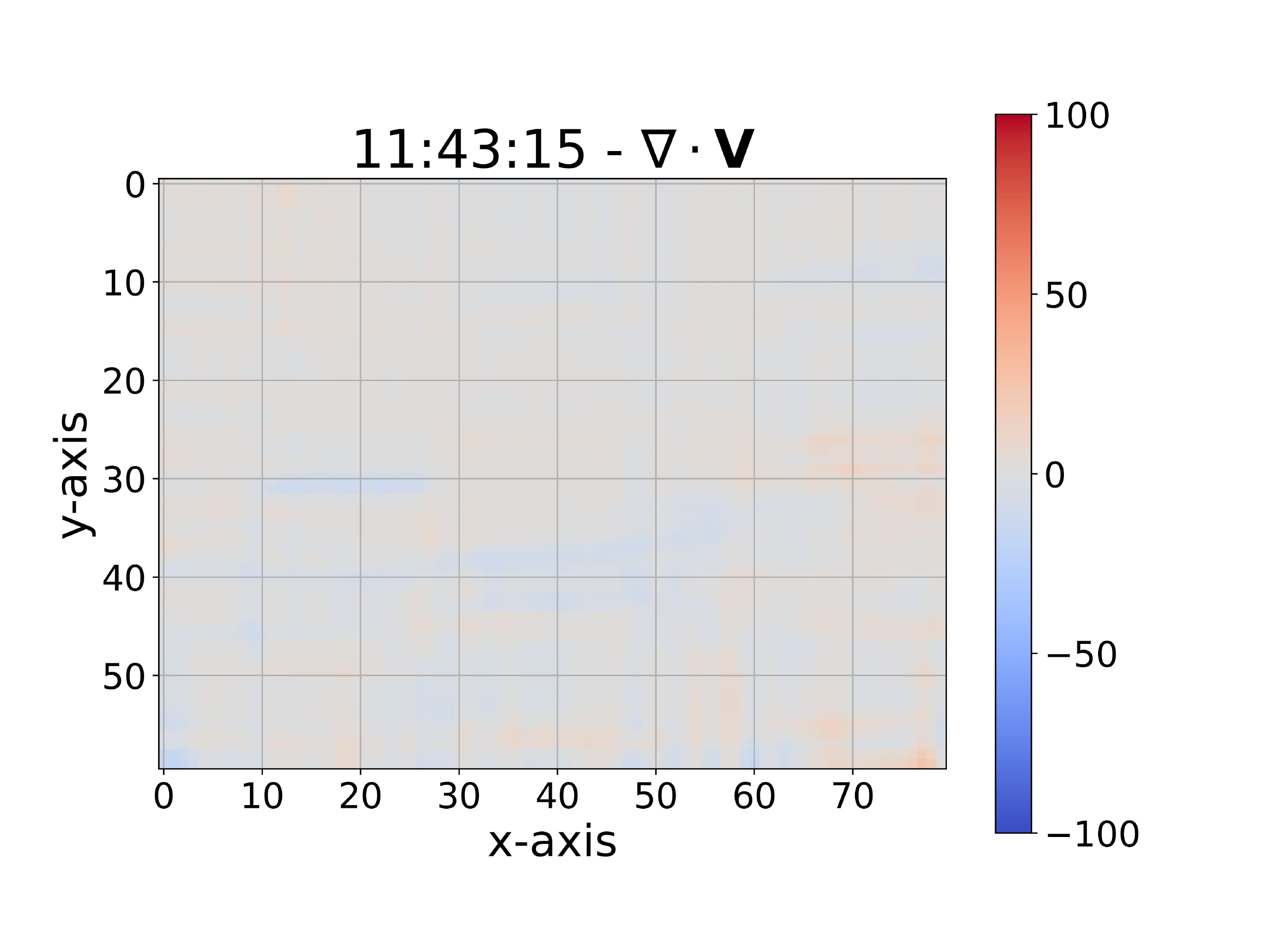}
        \includegraphics[scale = 0.2, trim = {1cm, 1.5cm, 2cm, 1.5cm}, clip]{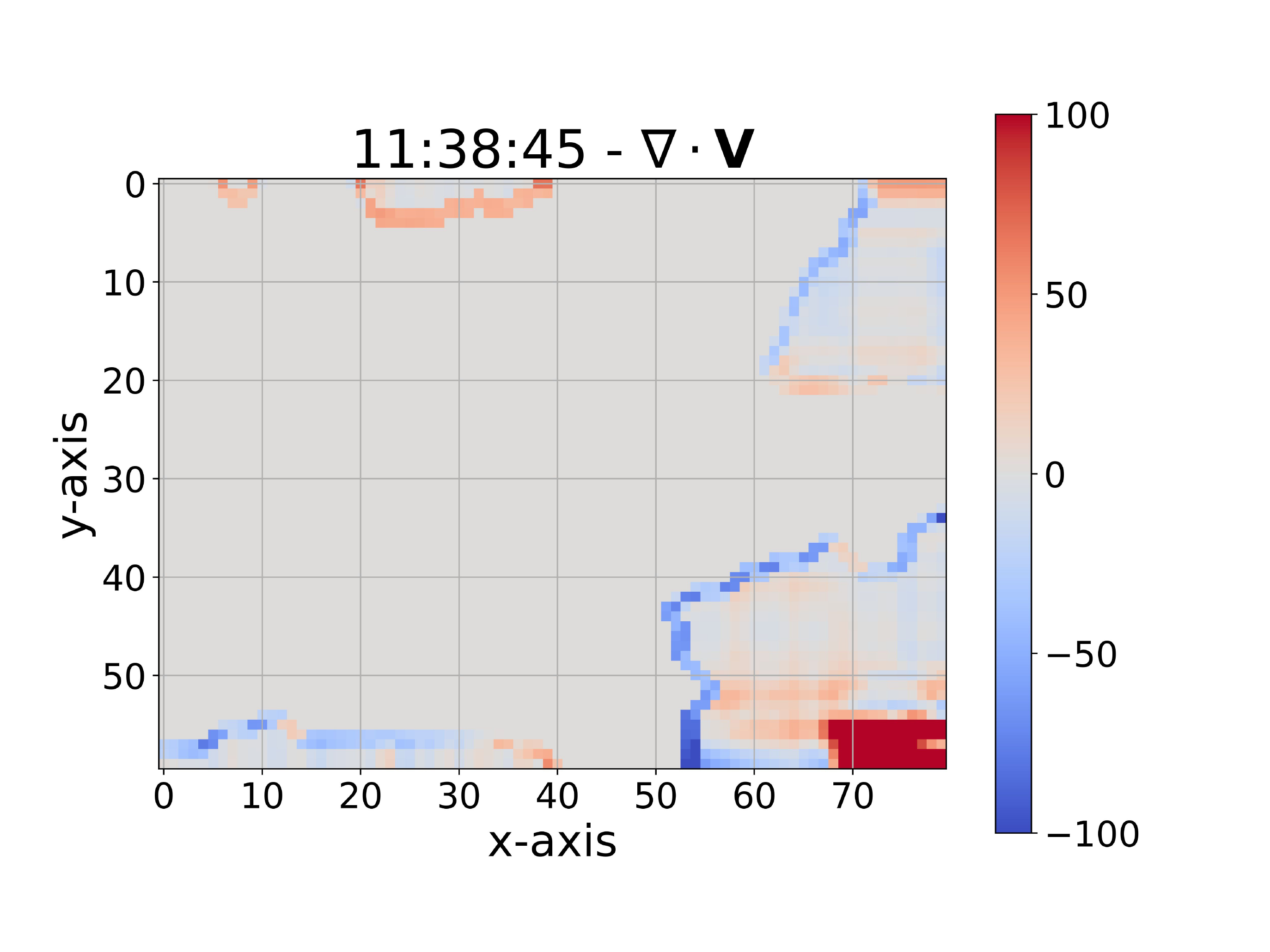}
    \end{subfigure}
    \caption{Features extracted from clouds in two different days of the same season at approximately the same hour of the day (i.e., approximately the same Sun's elevation and azimuth angle). The features in the left column were extracted from a slow evolving stratus cloud. The features in the right column were extracted from a fast evolving cumulus cloud. From top to bottom, the features are: processed temperature, height, velocity vector magnitude, vorticity and divergence.} 
    \label{fig:feature_extraction_algorithm}
\end{figure}

The IR images were processed for removing cyclostationary effects produced by radiation (see Figure~\ref{fig:image_processing}). After processing out the effects of the germanium outdoor camera window and the Sun and atmosphere effect in the IR images (see Figure~\ref{fig:radiantion_effect_models}), the heights and dynamics of clouds were computed. The heights were computed using the MALR (see Section \ref{sec:cloud_infrared_images}). The method to compute the cloud velocity vectors is a weighted implementation of the Lucas-Kanade algorithm. The images were normalized to avoid intensity fluctuations that may affect the accuracy of the velocity vectors (see \cite{TERREN2021a} for more information about the normalization and the image processing applied to the IR images). The cloud velocity vectors were used to compute the divergence and vorticity (see Section \ref{sec:cloud_dynamics}). The features extracted from IR images are shown in Figure~\ref{fig:feature_extraction_algorithm}.

\begin{figure}[!htb]
    \begin{subfigure}{\linewidth}
        \centering
        \includegraphics[scale = 0.225, trim = {1cm, 3.5cm, 2cm, 4.3cm}, clip]{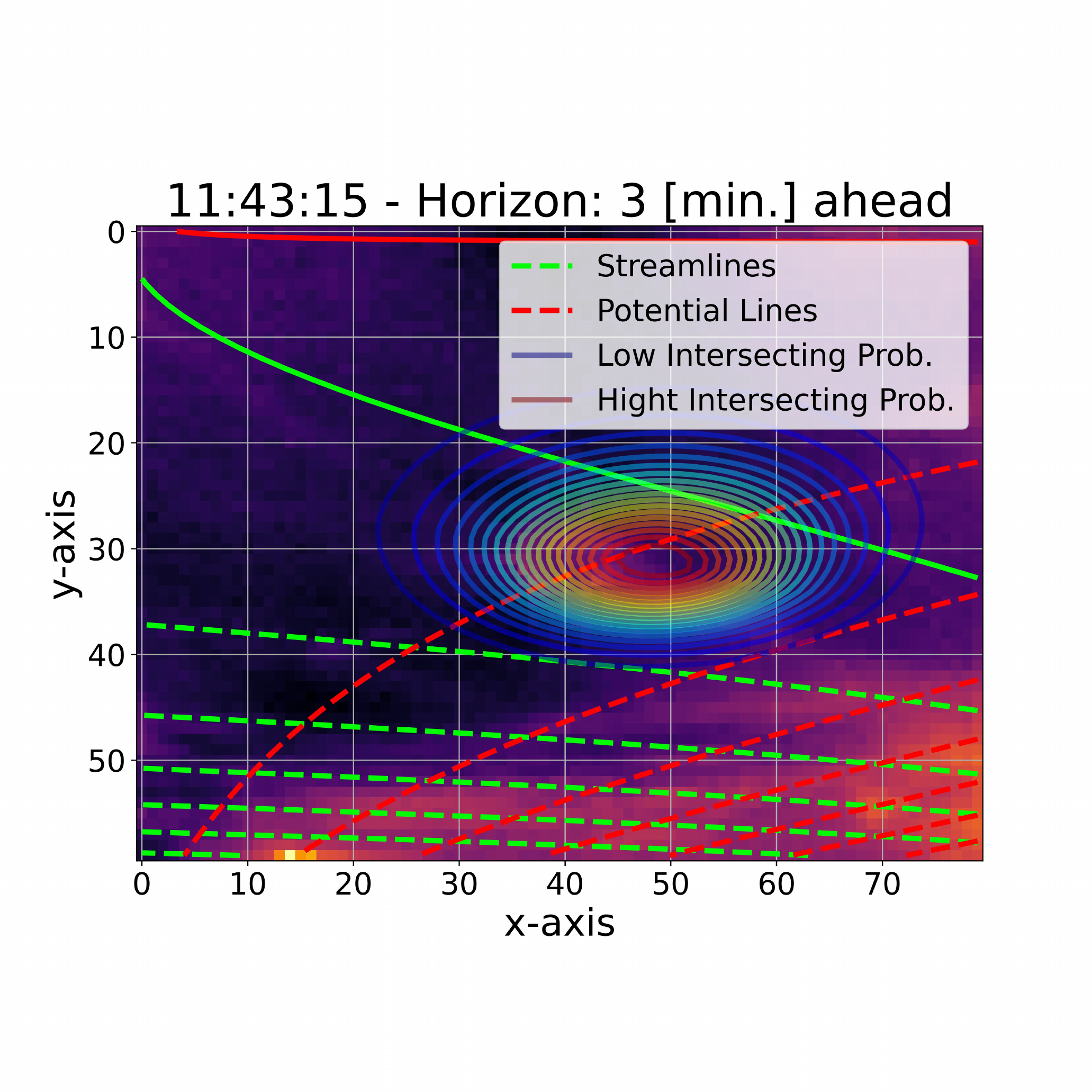}
        \includegraphics[scale = 0.225, trim = {1cm, 3.5cm, 2cm, 4.3cm}, clip]{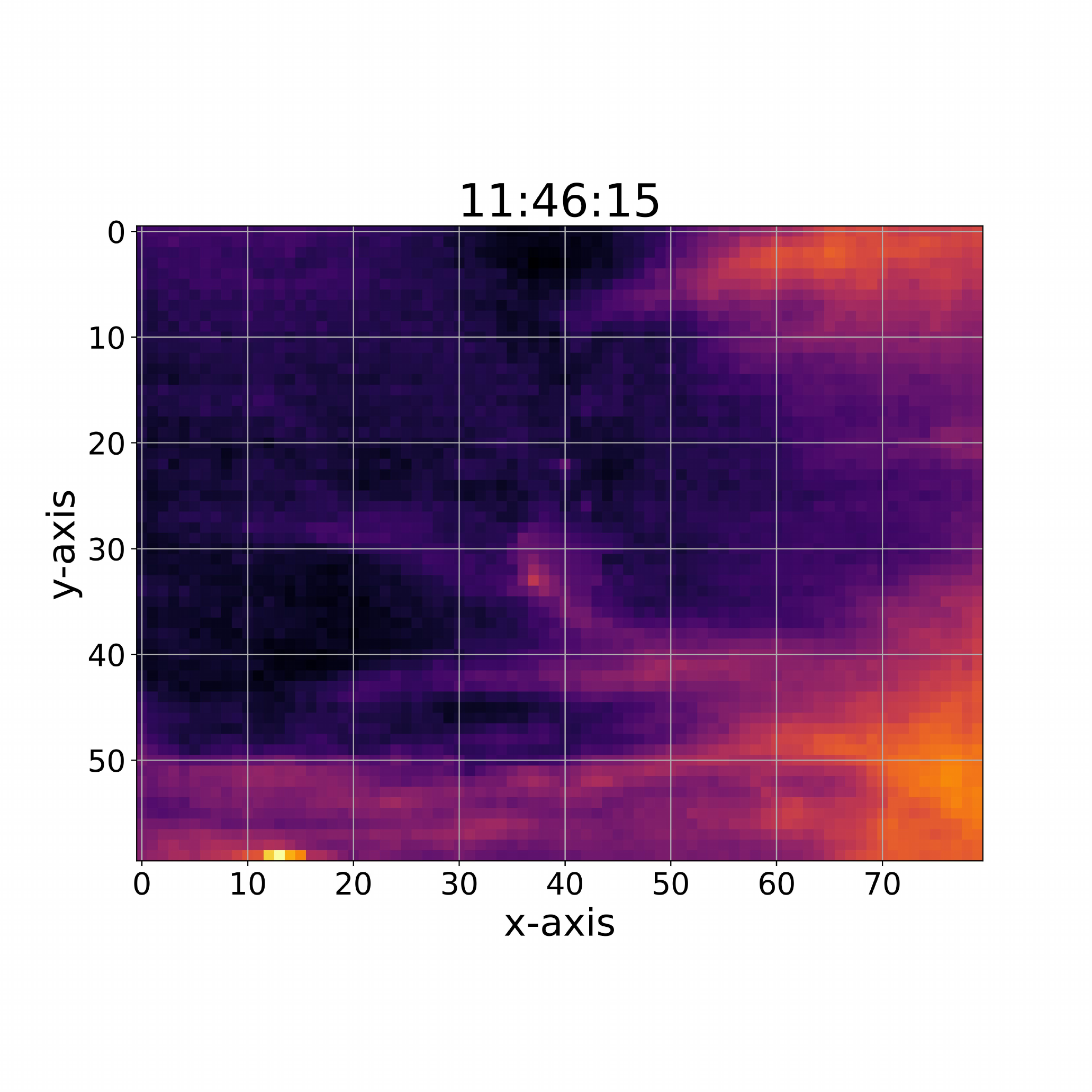}
        \includegraphics[scale = 0.225, trim = {1cm, 3.5cm, 2cm, 4.3cm}, clip]{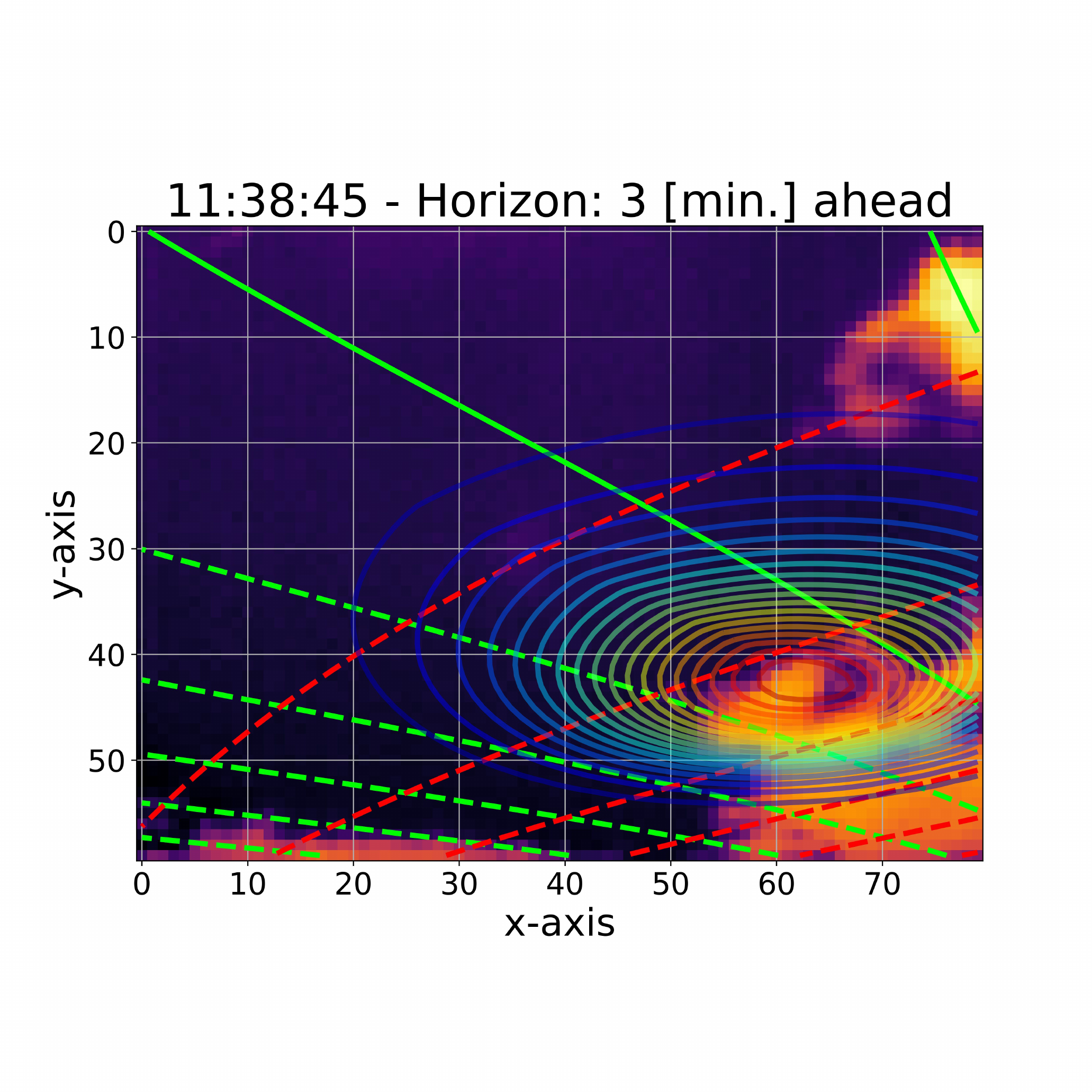}
        \includegraphics[scale = 0.225, trim = {1cm, 3.5cm, 2cm, 4.3cm}, clip]{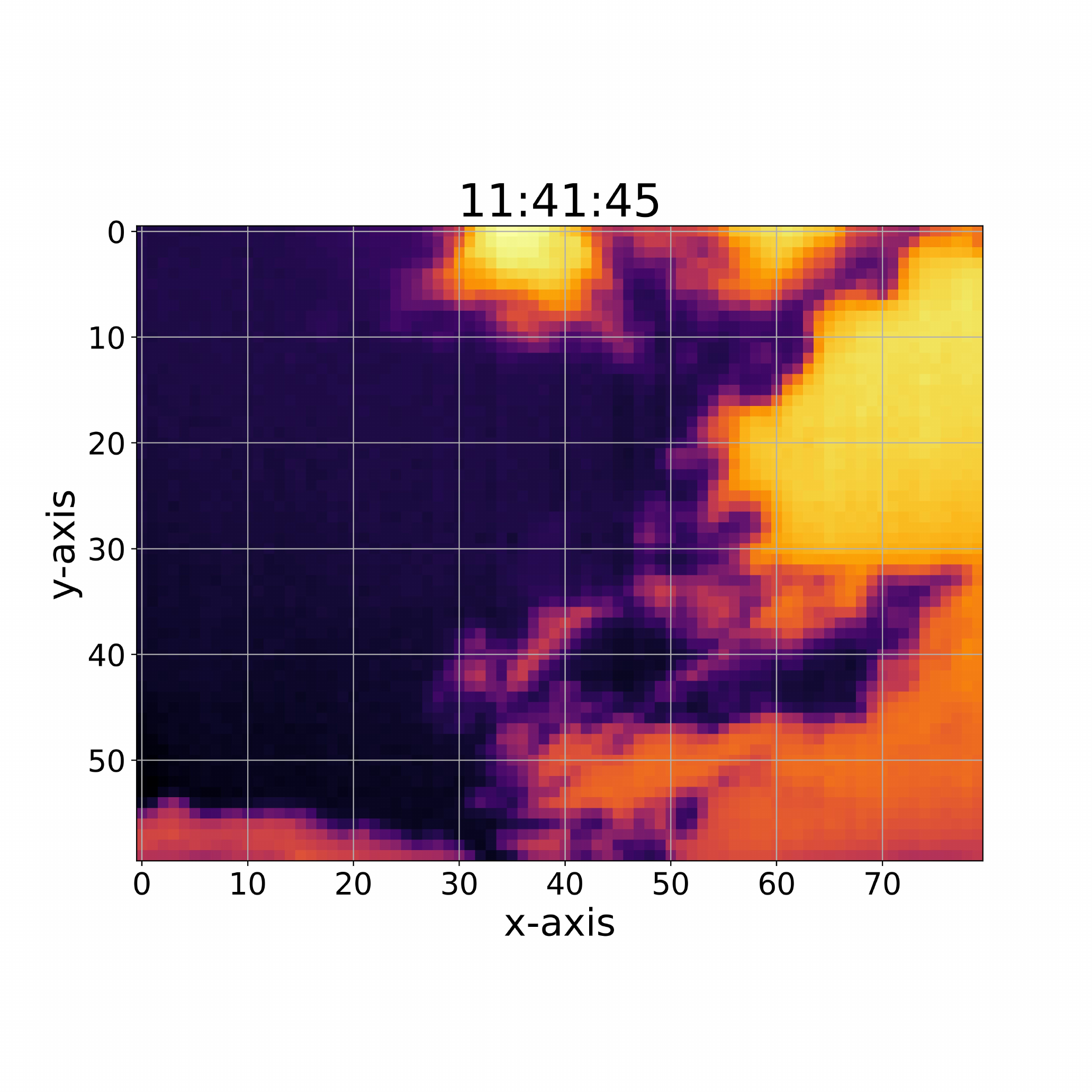}
    \end{subfigure}
    \begin{subfigure}{\linewidth}
        \centering
        \includegraphics[scale = 0.225, trim = {1cm, 3.5cm, 2cm, 4.3cm}, clip]{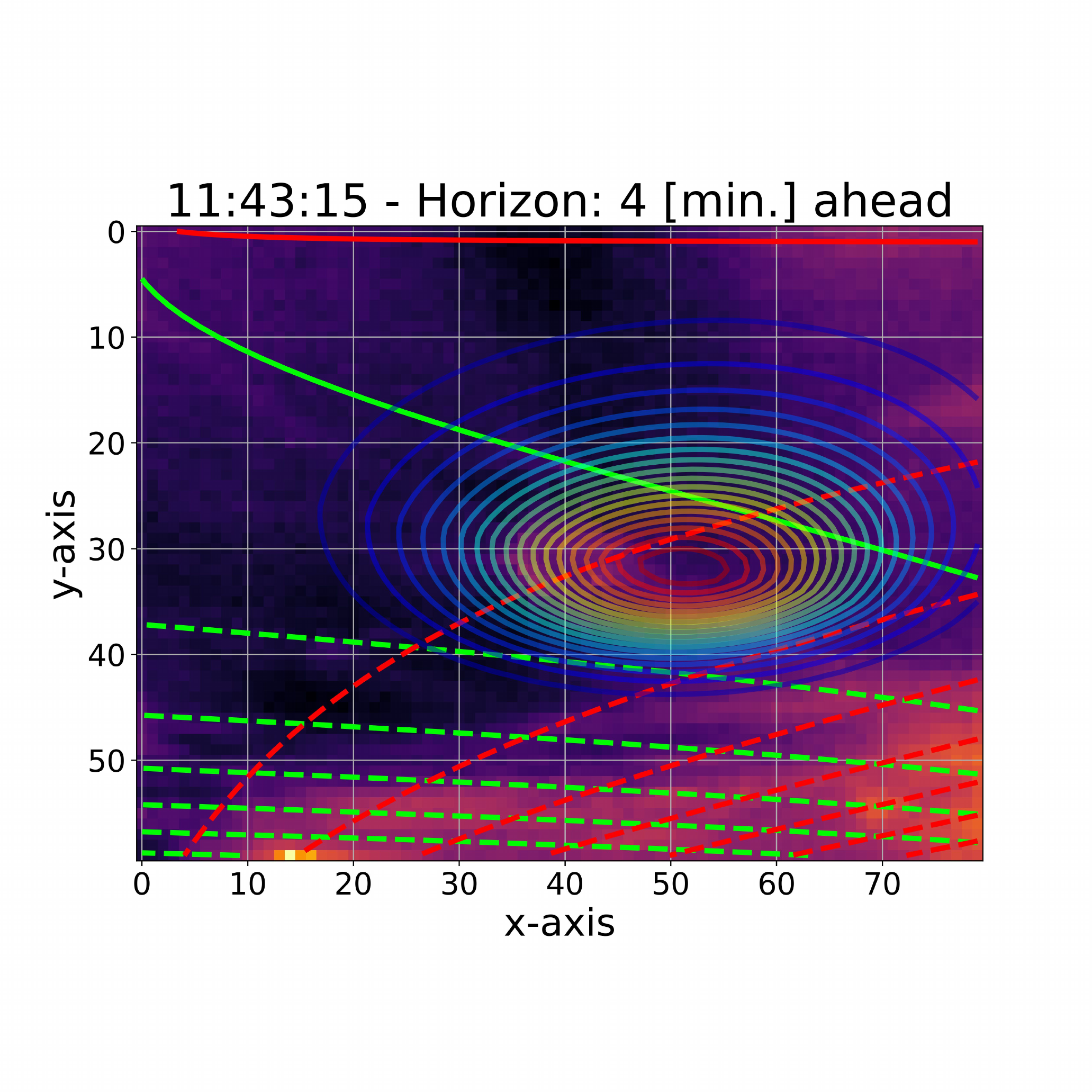}
        \includegraphics[scale = 0.225, trim = {1cm, 3.5cm, 2cm, 4.3cm}, clip]{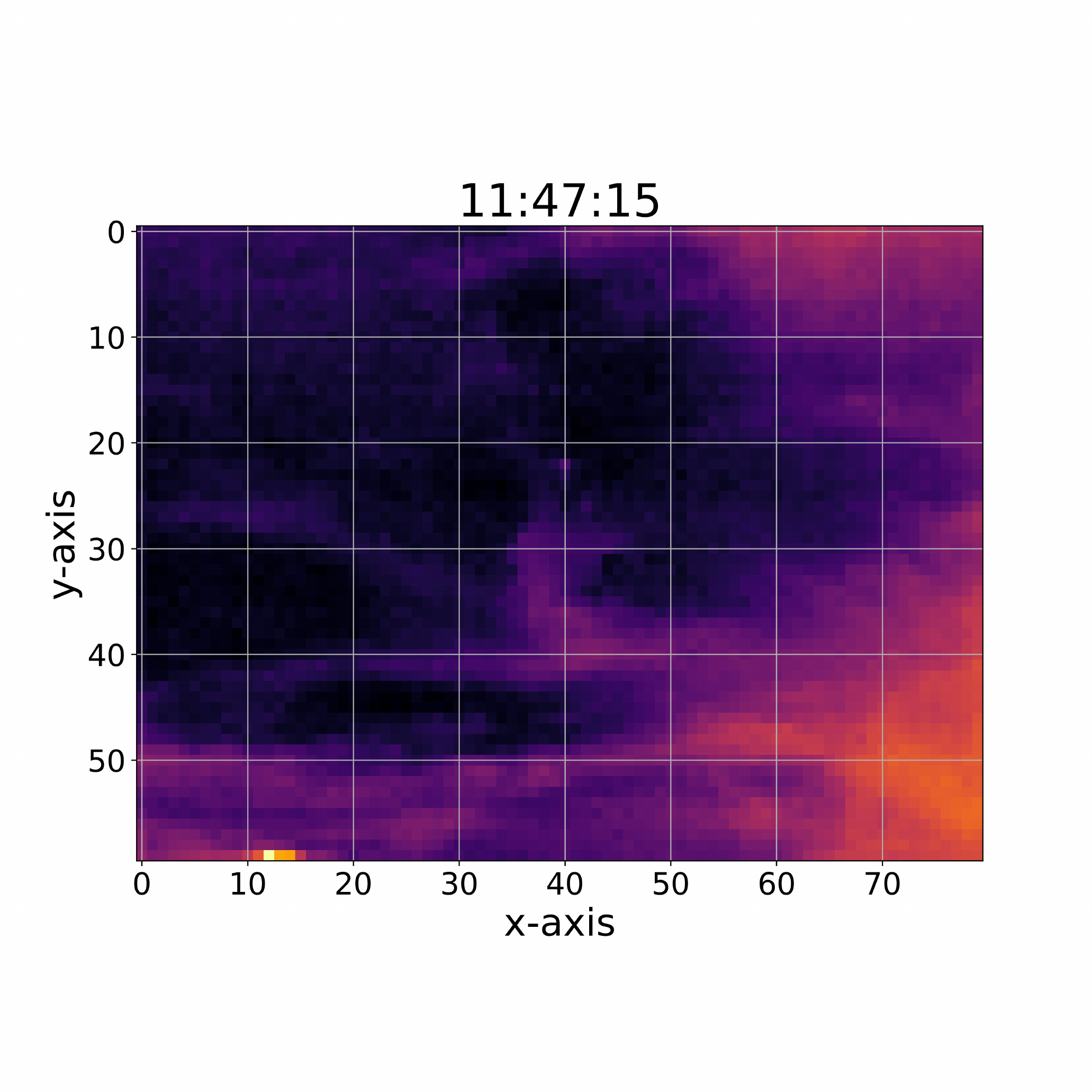}
        \includegraphics[scale = 0.225, trim = {1cm, 3.5cm, 2cm, 4.3cm}, clip]{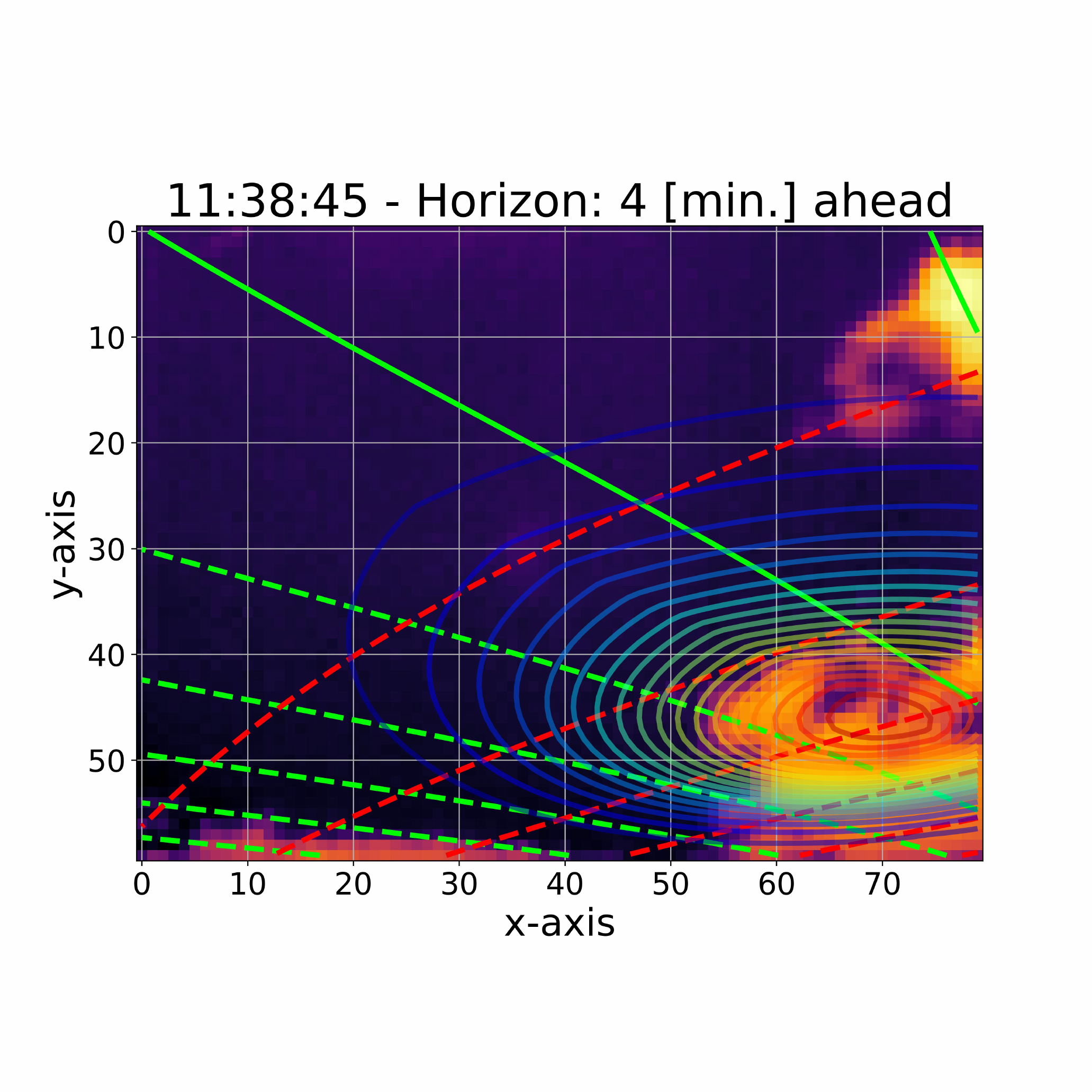}
        \includegraphics[scale = 0.225, trim = {1cm, 3.5cm, 2cm, 4.3cm}, clip]{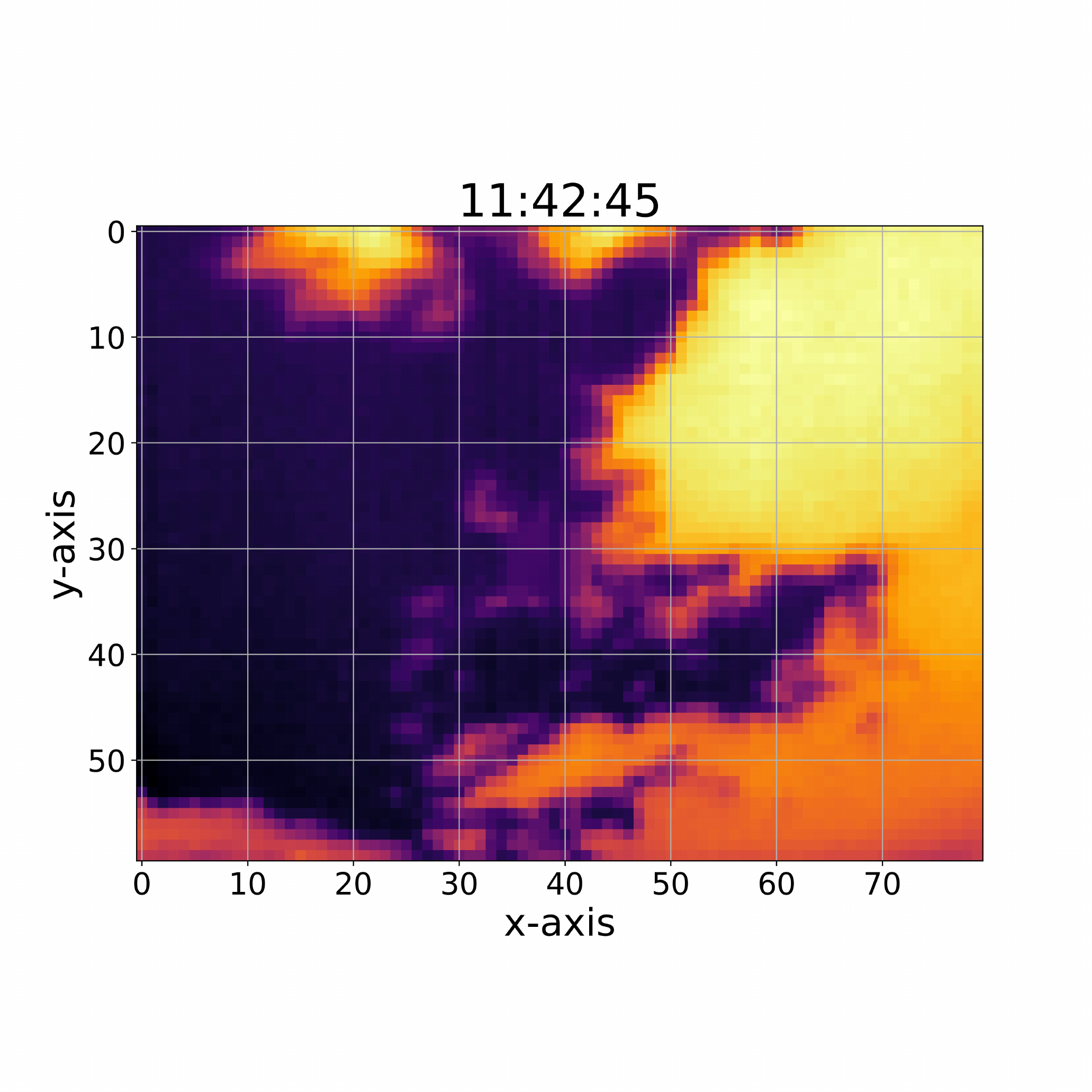}
    \end{subfigure}
    \begin{subfigure}{\linewidth}
        \centering
        \includegraphics[scale = 0.225, trim = {1cm, 3.5cm, 2cm, 4.3cm}, clip]{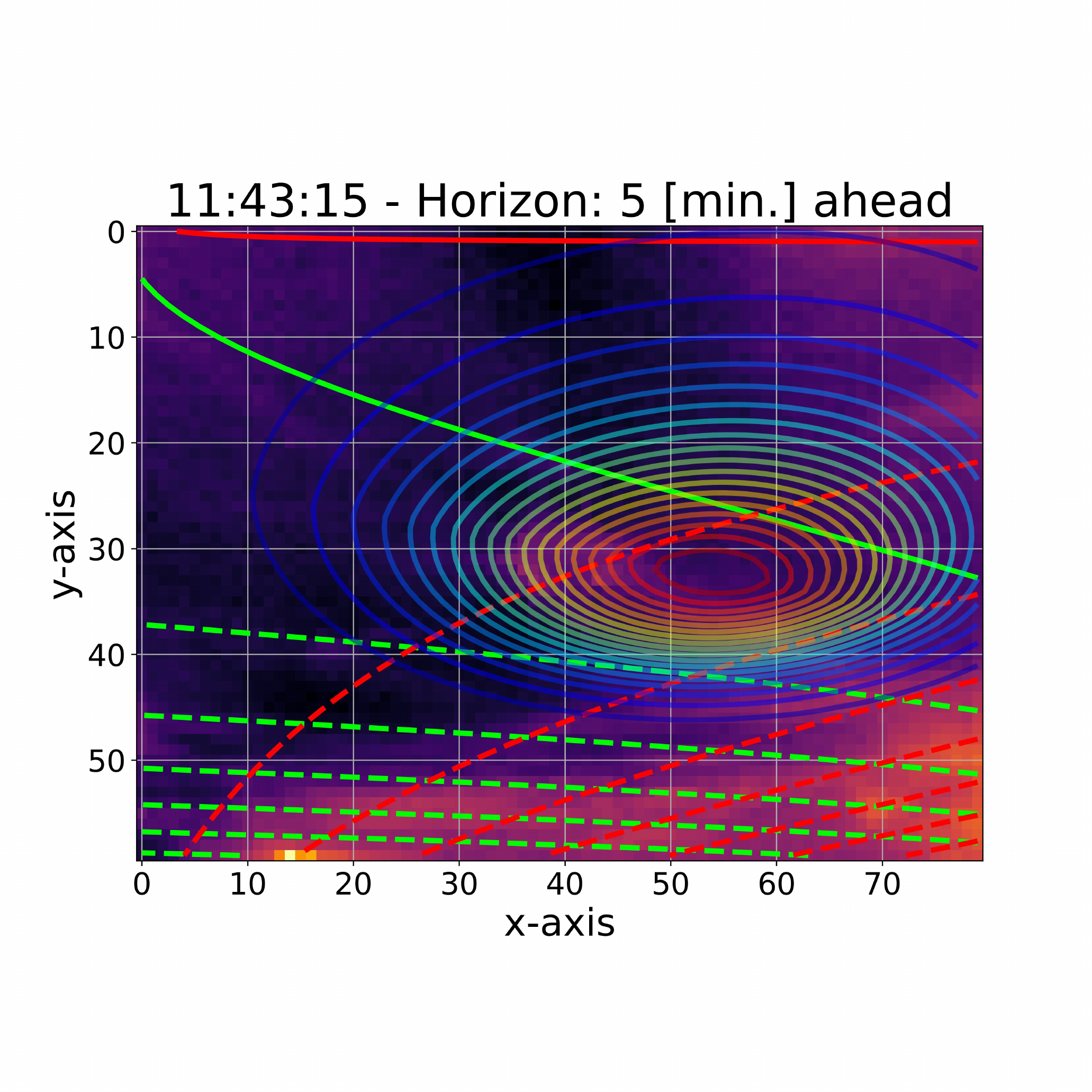}
        \includegraphics[scale = 0.225, trim = {1cm, 3.5cm, 2cm, 4.3cm}, clip]{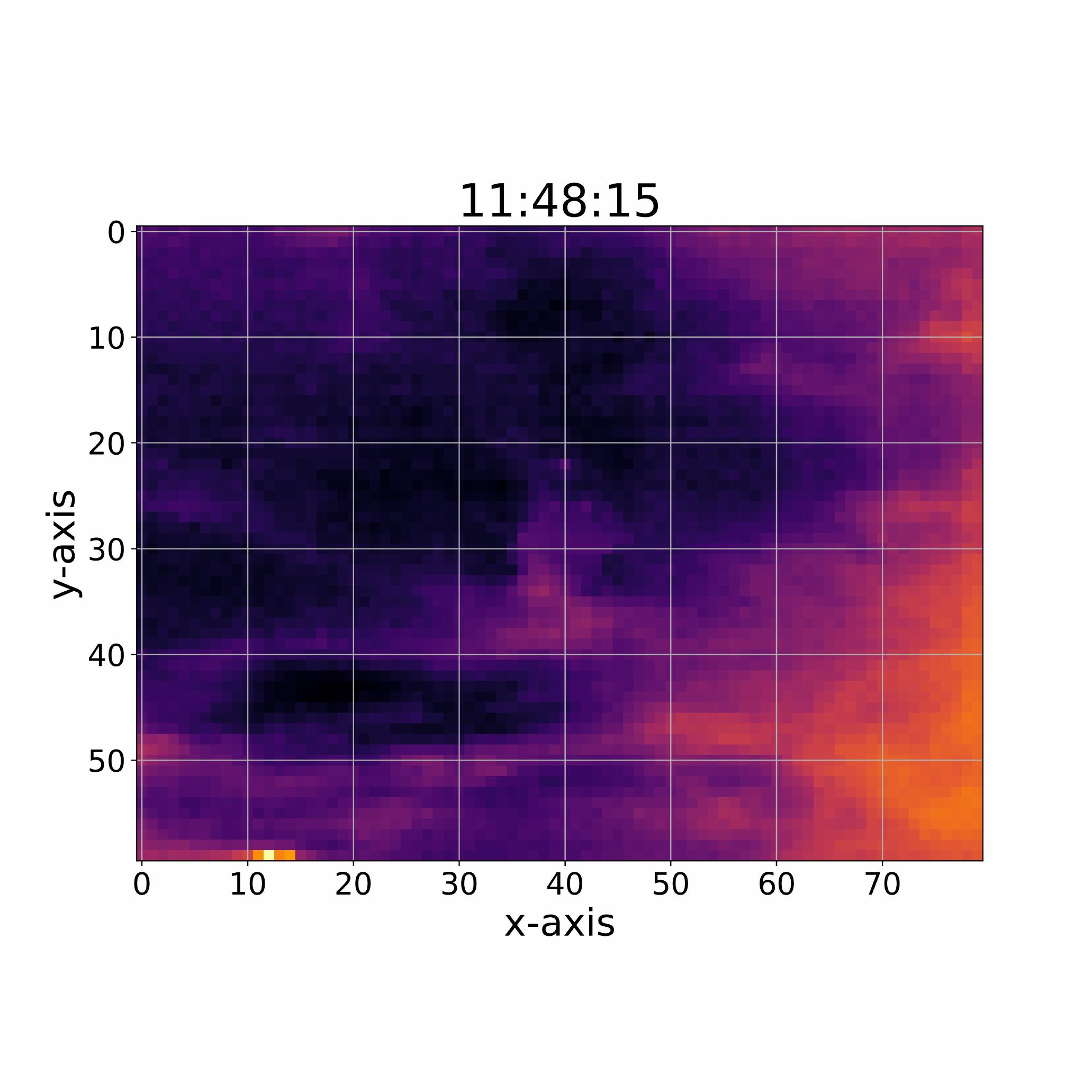}
        \includegraphics[scale = 0.225, trim = {1cm, 3.5cm, 2cm, 4.3cm}, clip]{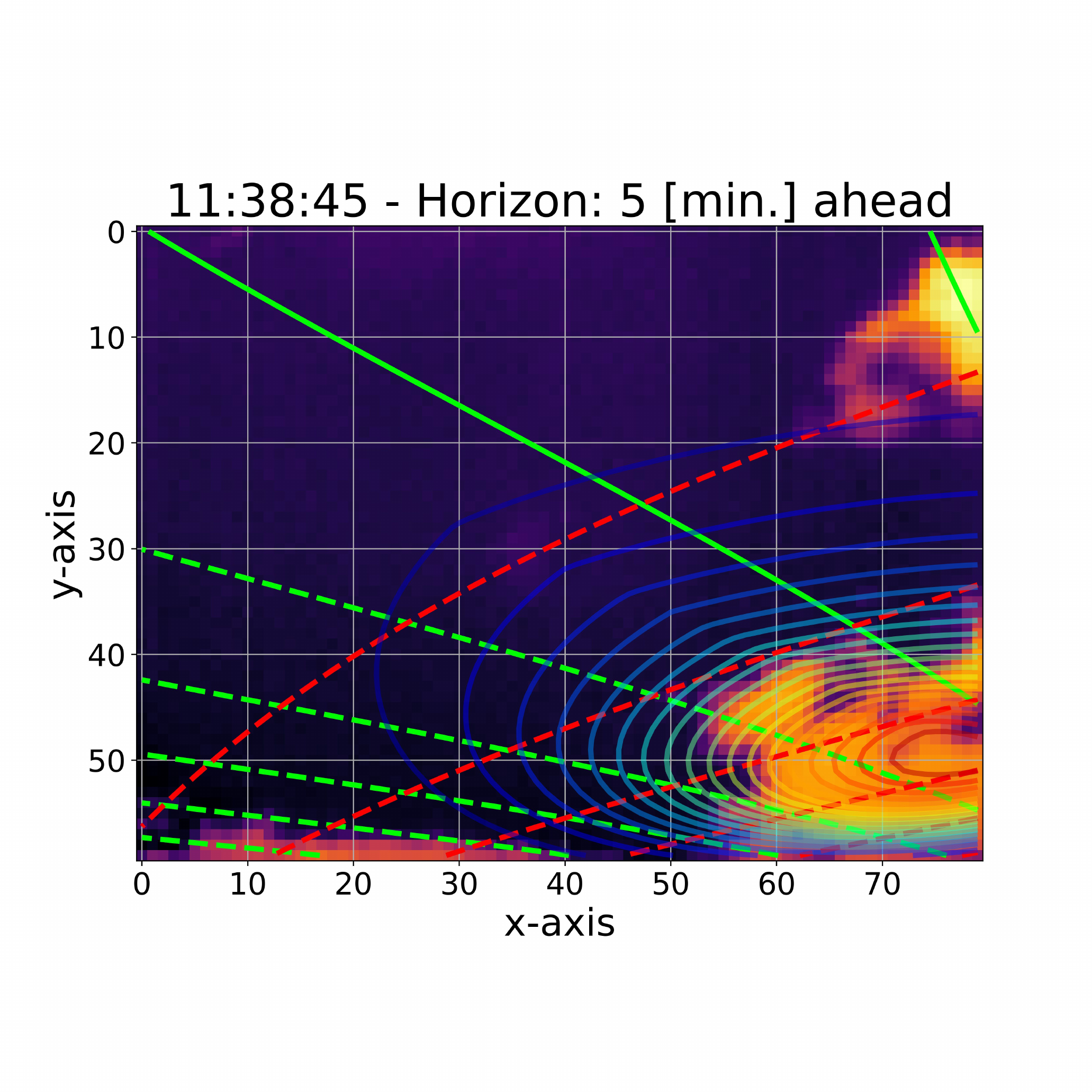}
        \includegraphics[scale = 0.225, trim = {1cm, 3.5cm, 2cm, 4.3cm}, clip]{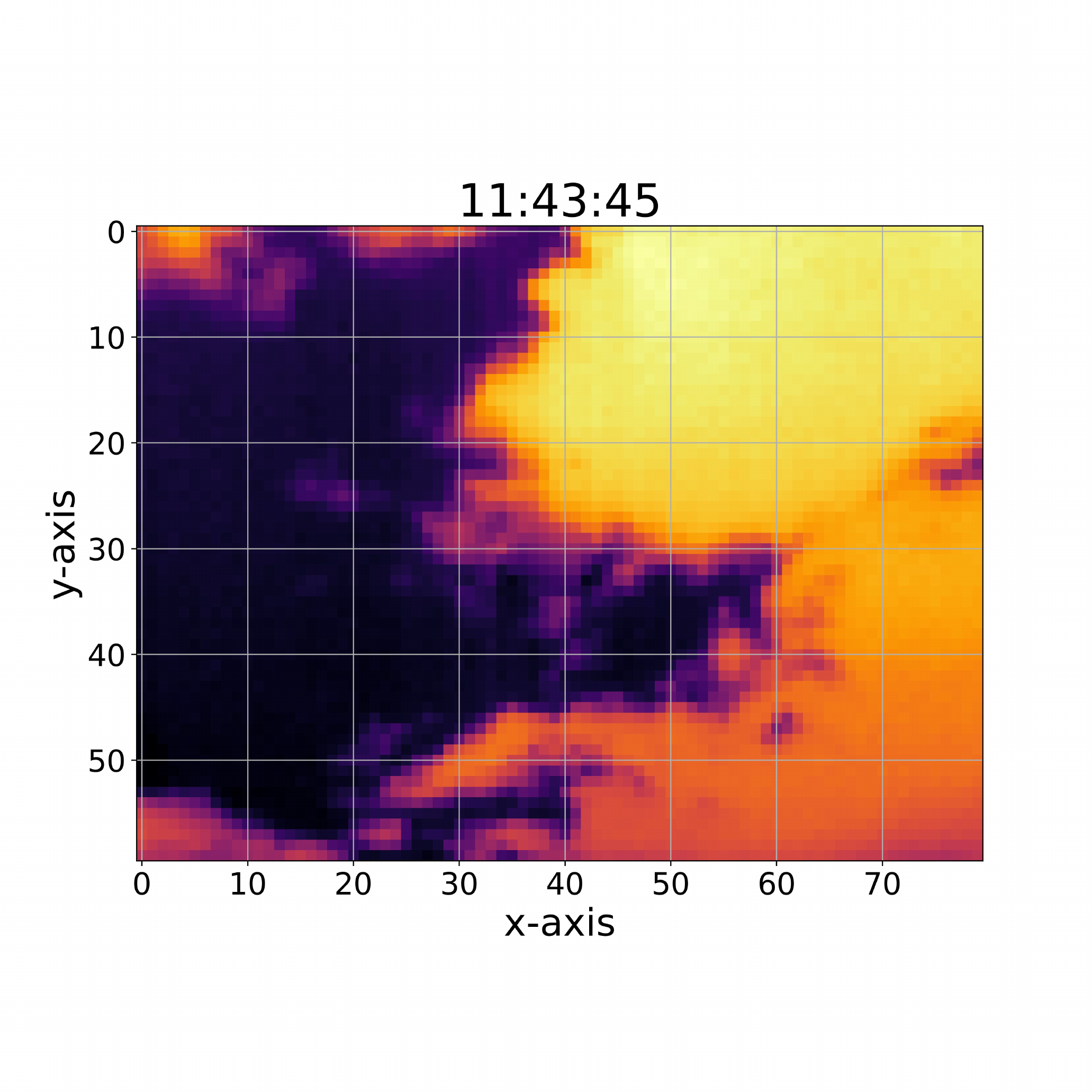}
    \end{subfigure}
    \begin{subfigure}{\linewidth}
        \centering
        \includegraphics[scale = 0.225, trim = {1cm, 3.5cm, 2cm, 4.3cm}, clip]{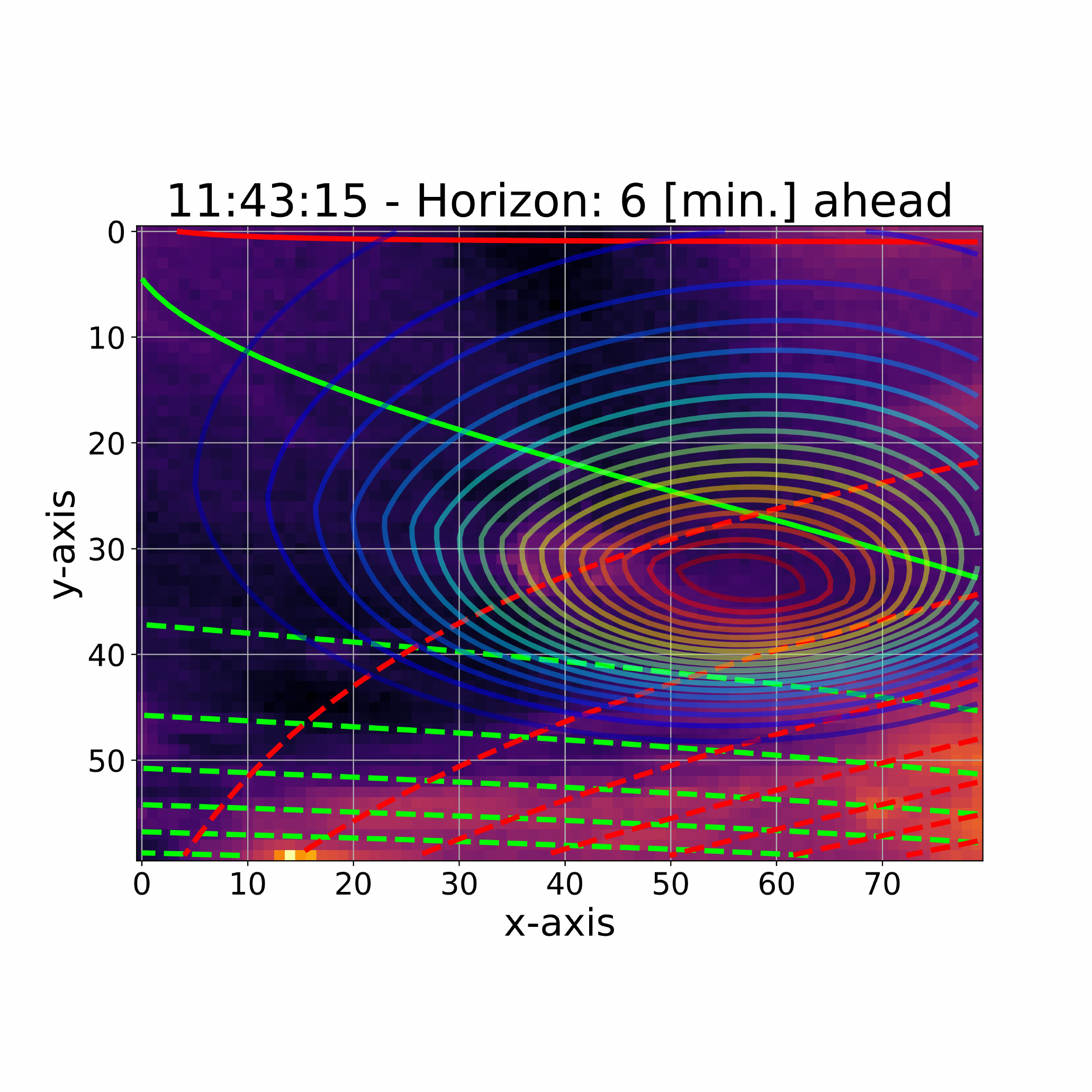}
        \includegraphics[scale = 0.225, trim = {1cm, 3.5cm, 2cm, 4.3cm}, clip]{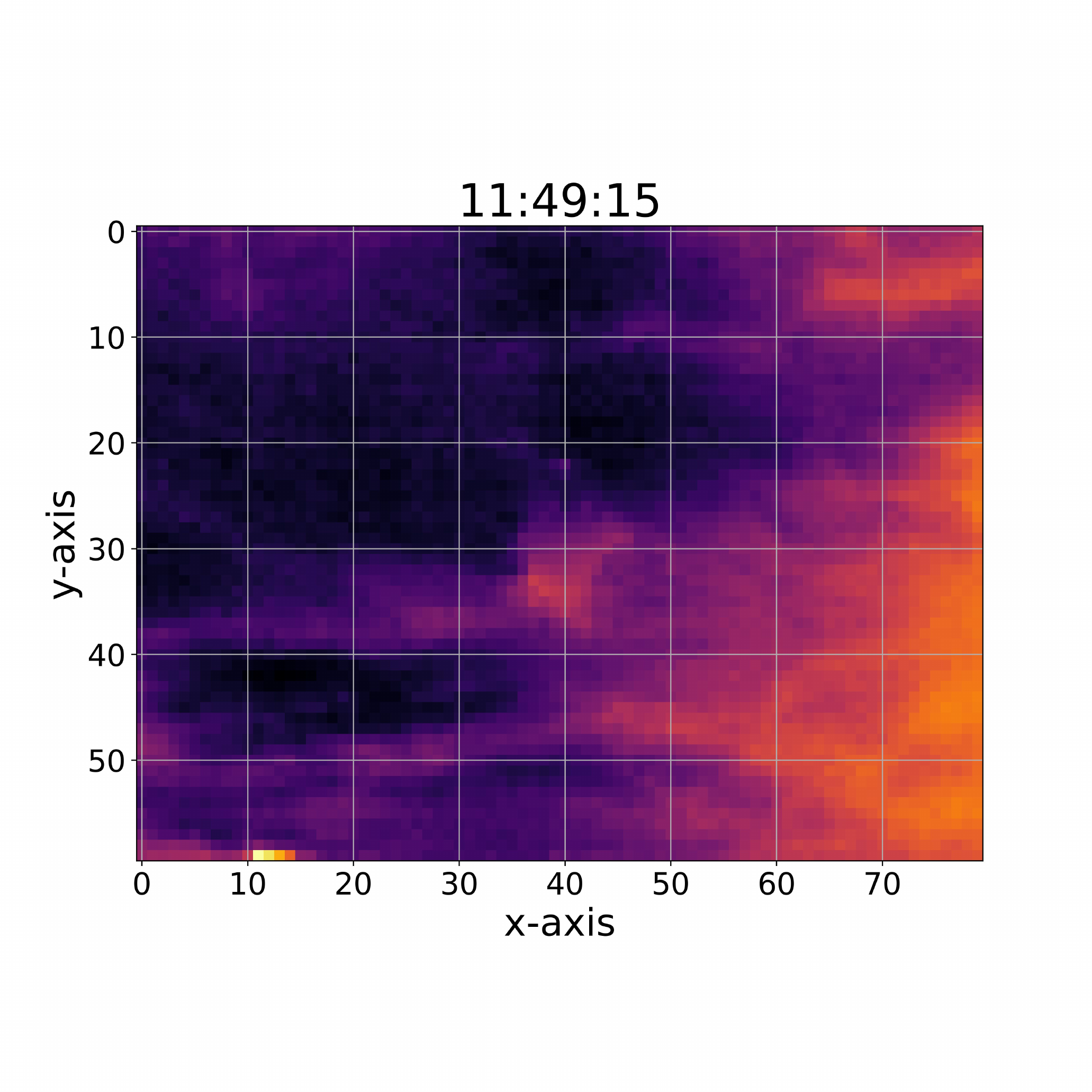}
        \includegraphics[scale = 0.225, trim = {1cm, 3.5cm, 2cm, 4.3cm}, clip]{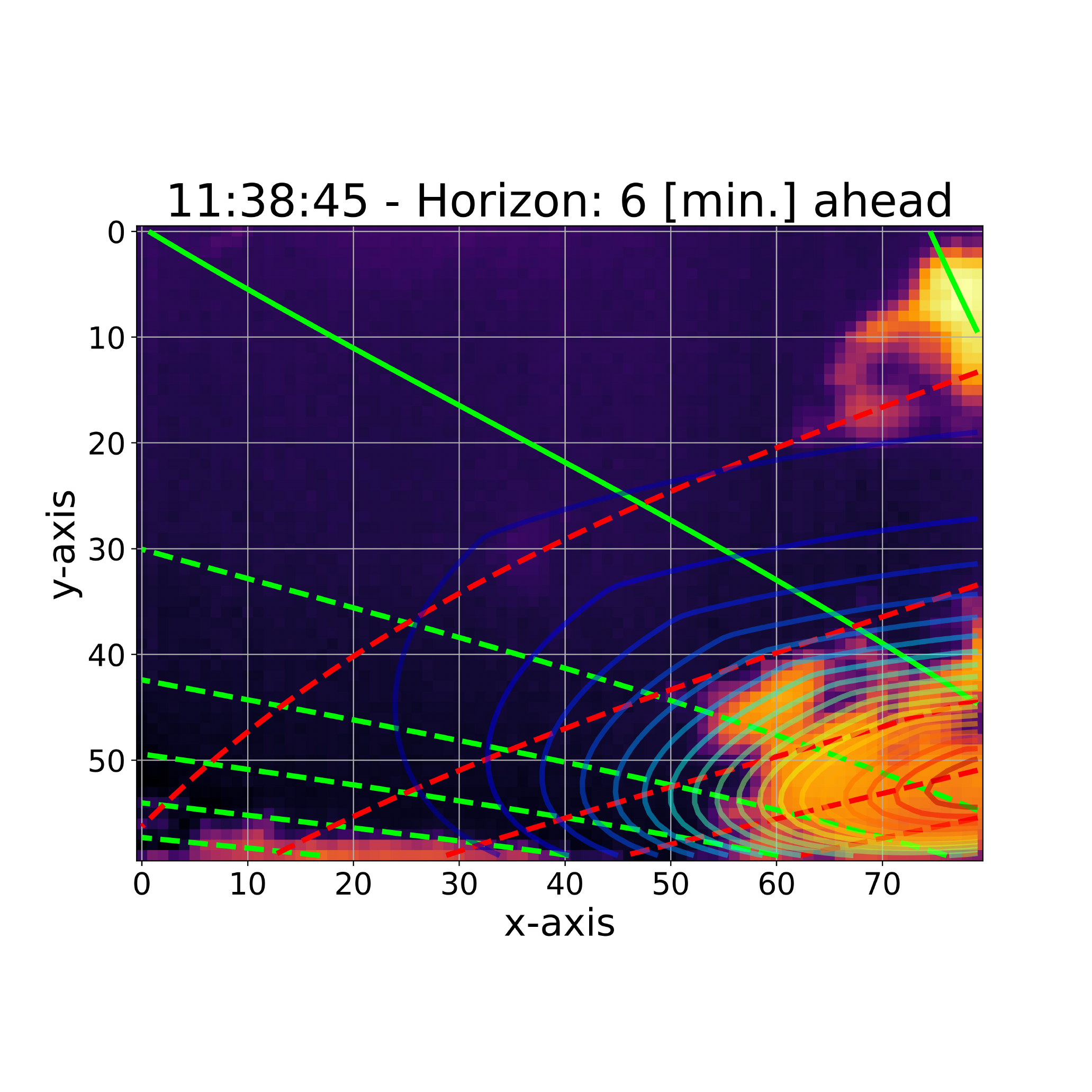}
        \includegraphics[scale = 0.225, trim = {1cm, 3.5cm, 2cm, 4.3cm}, clip]{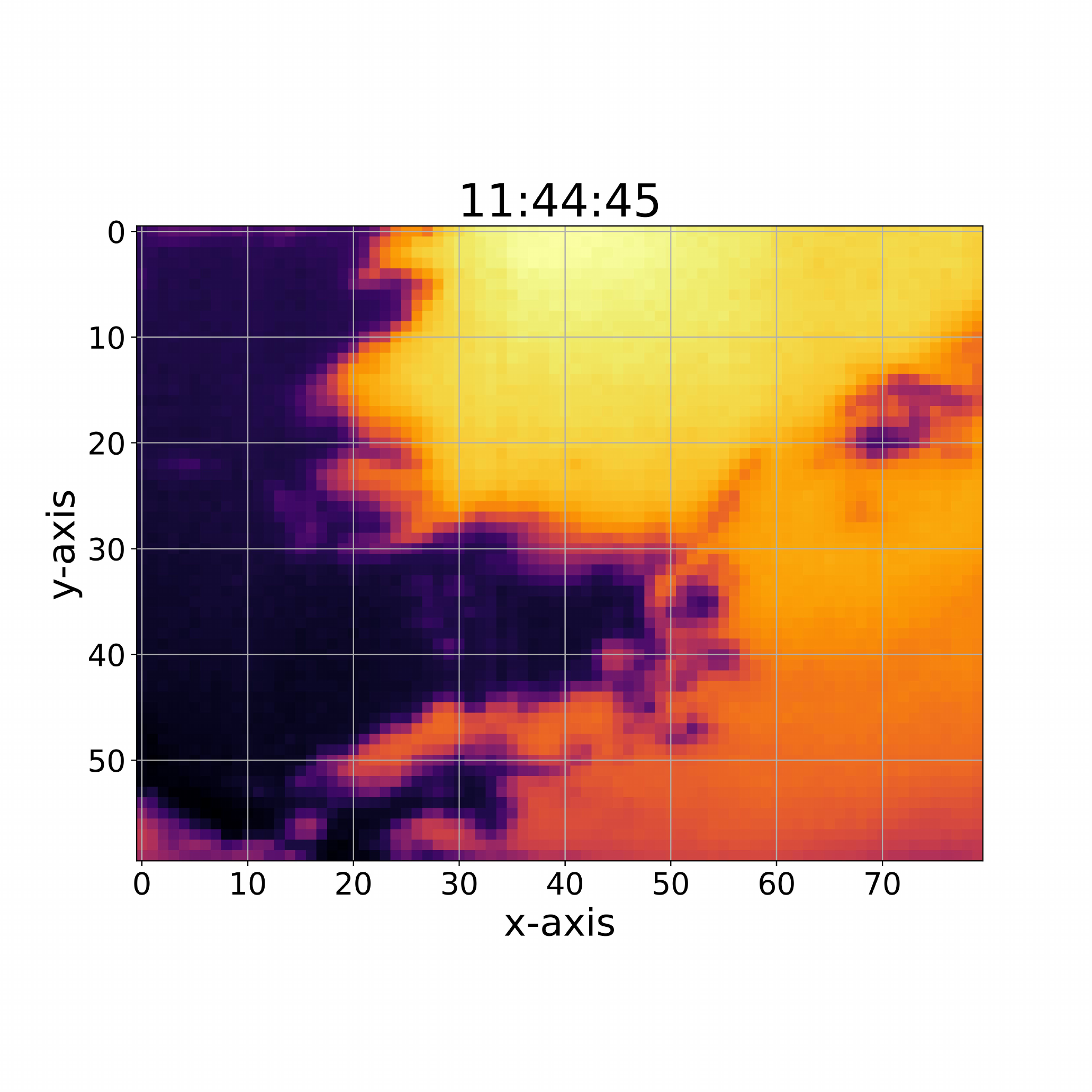}
    \end{subfigure}
    \begin{subfigure}{\linewidth}
        \centering
        \includegraphics[scale = 0.225, trim = {1cm, 3.5cm, 2cm, 4.3cm}, clip]{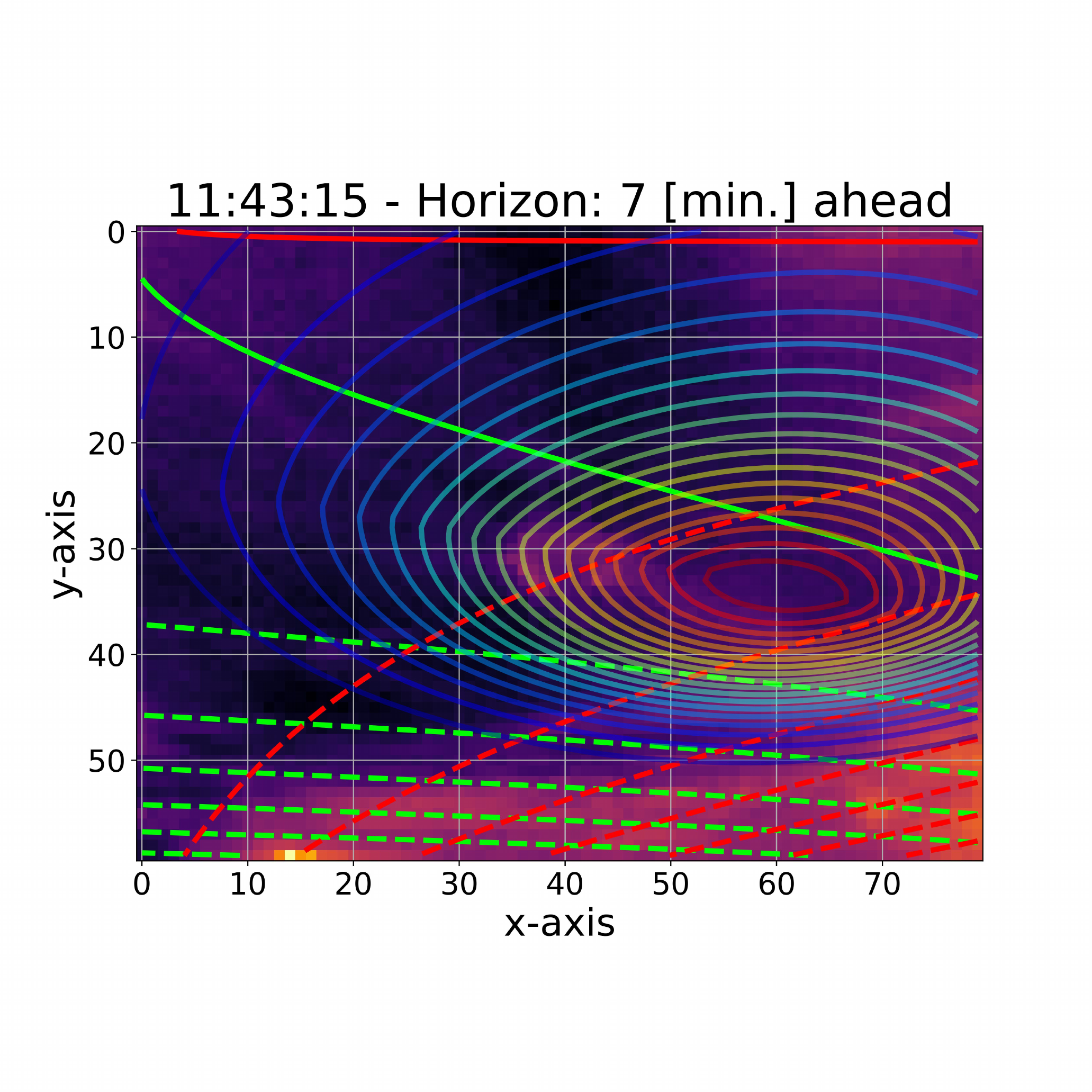}
        \includegraphics[scale = 0.225, trim = {1cm, 3.5cm, 2cm, 4.3cm}, clip]{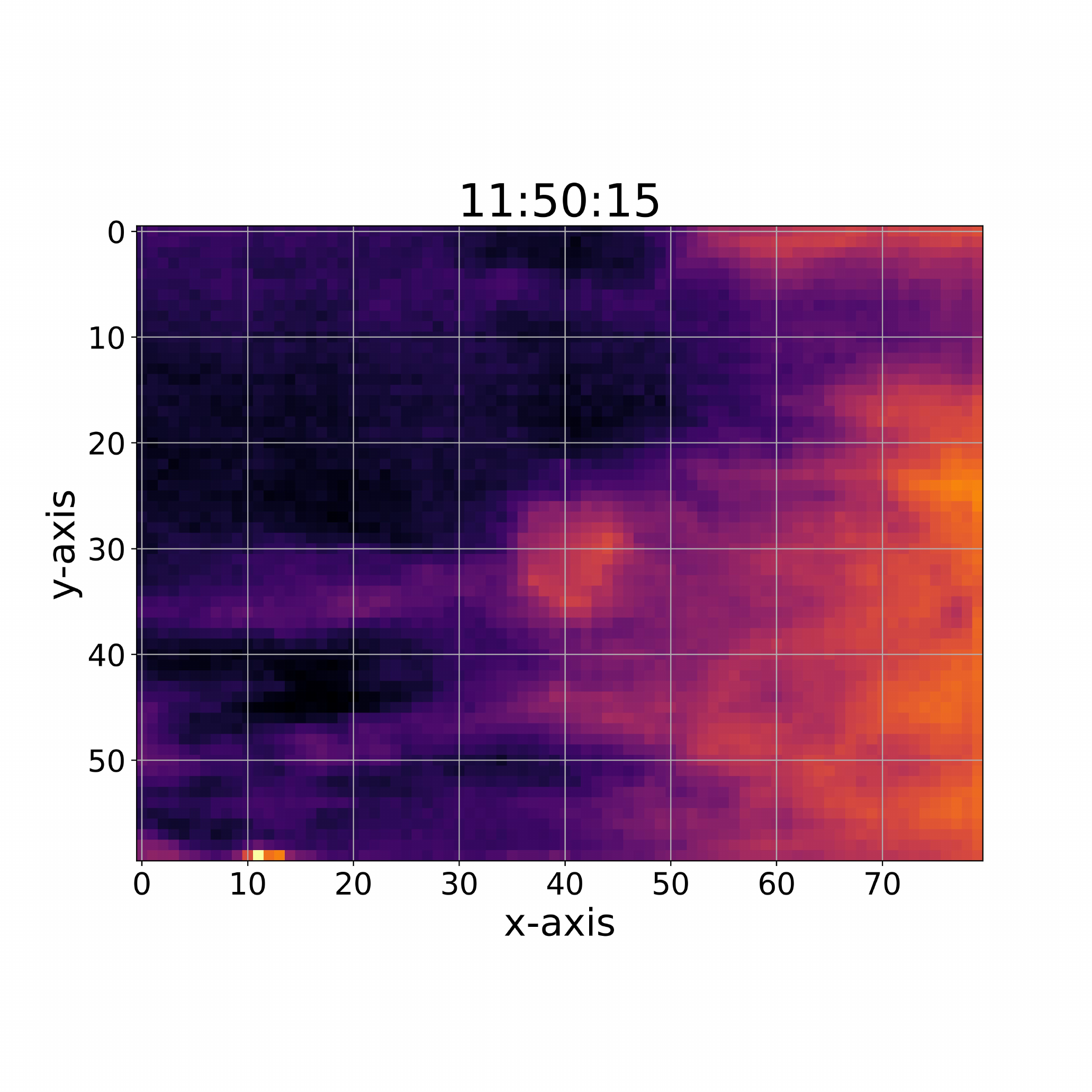}
        \includegraphics[scale = 0.225, trim = {1cm, 3.5cm, 2cm, 4.3cm}, clip]{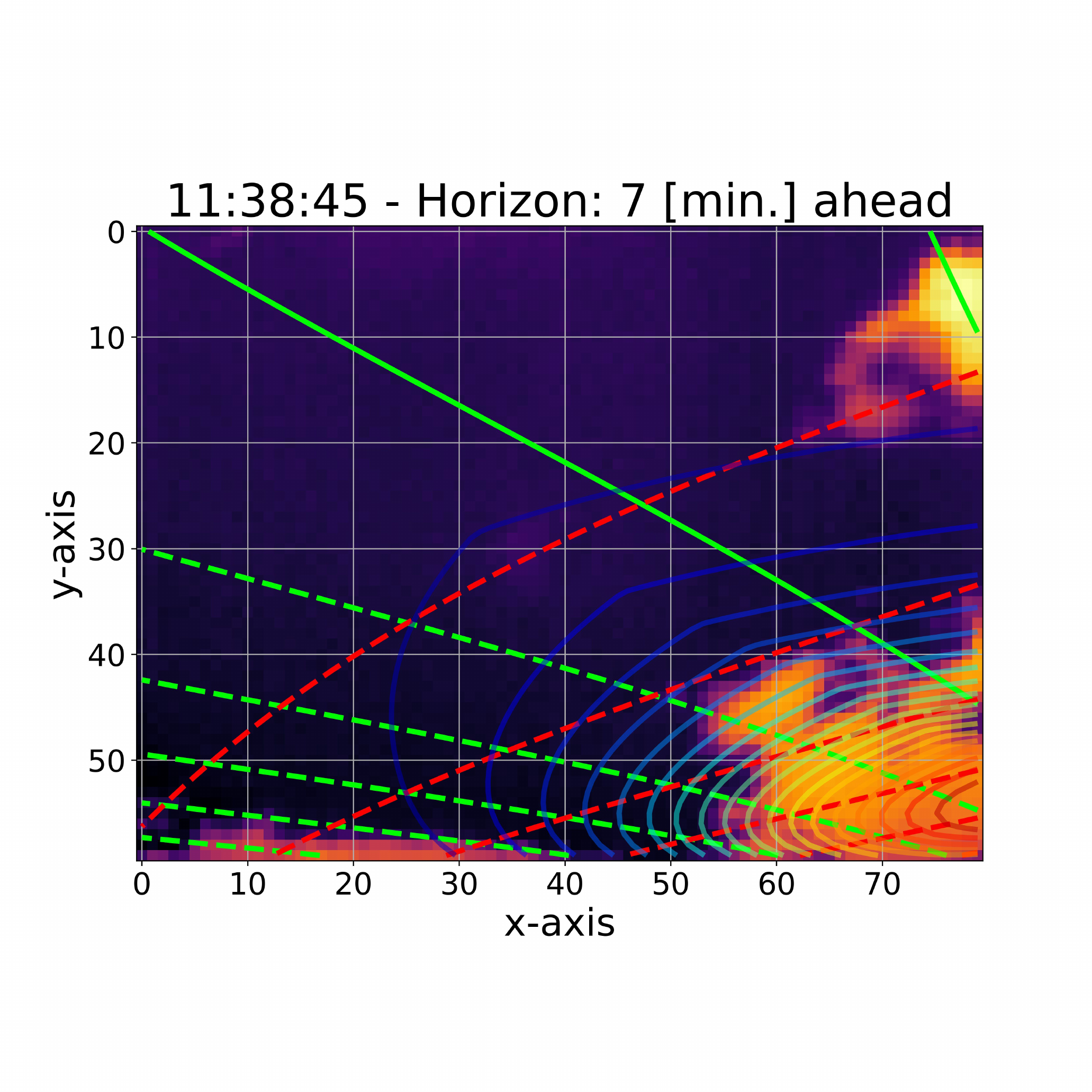}
        \includegraphics[scale = 0.225, trim = {1cm, 3.5cm, 2cm, 4.3cm}, clip]{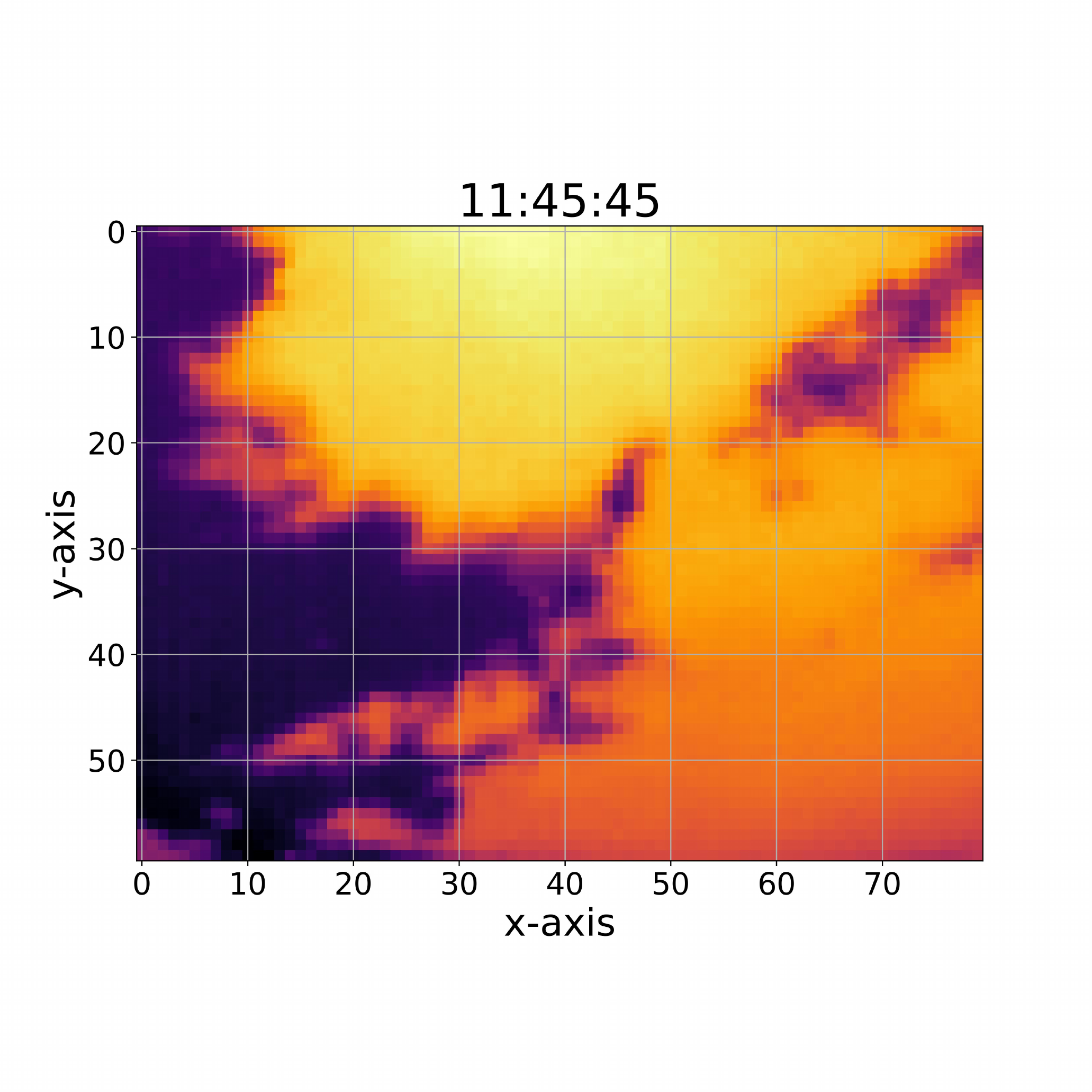}
    \end{subfigure}
    \begin{subfigure}{\linewidth}
        \centering
        \includegraphics[scale = 0.225, trim = {1cm, 3.5cm, 2cm, 4.3cm}, clip]{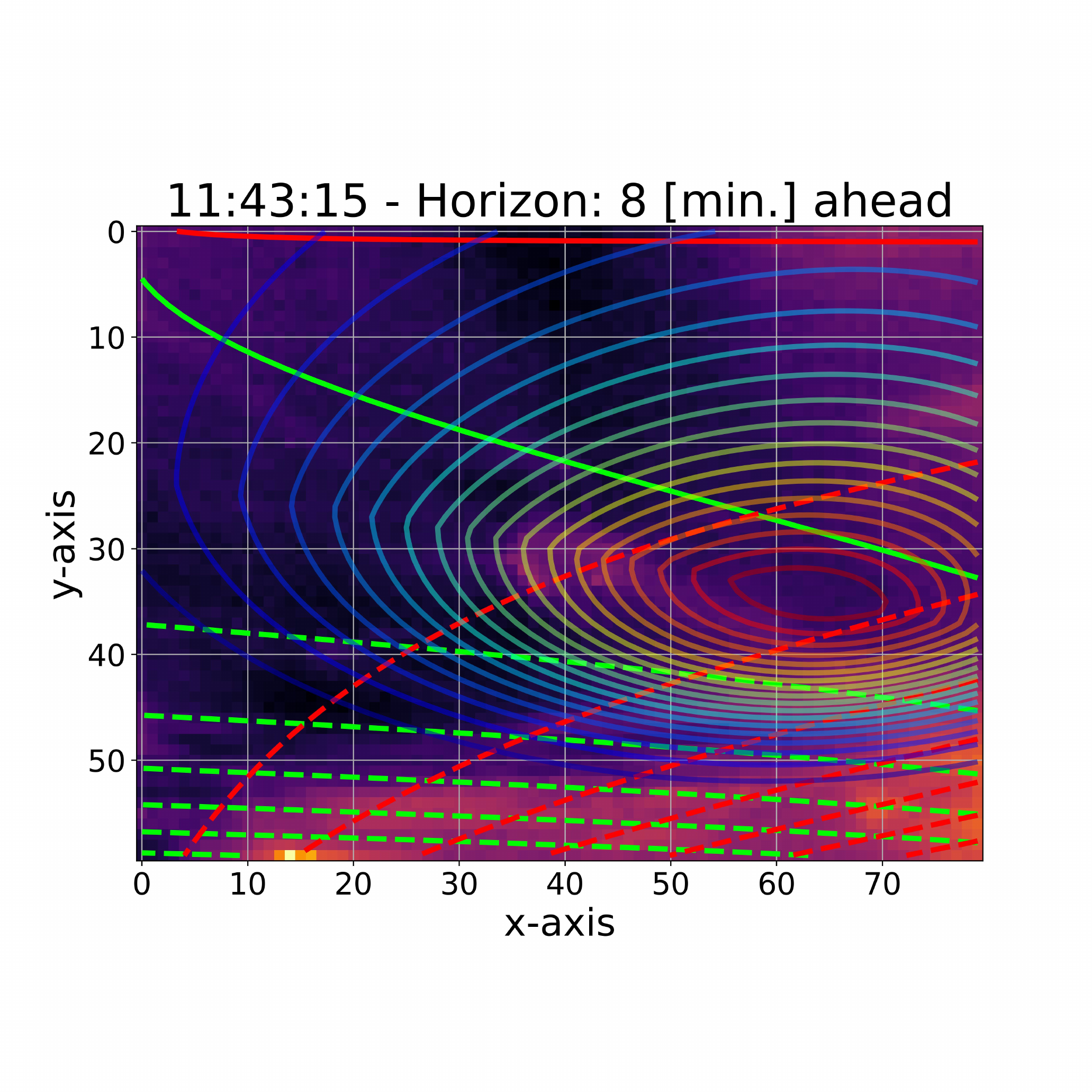}
        \includegraphics[scale = 0.225, trim = {1cm, 3.5cm, 2cm, 4.3cm}, clip]{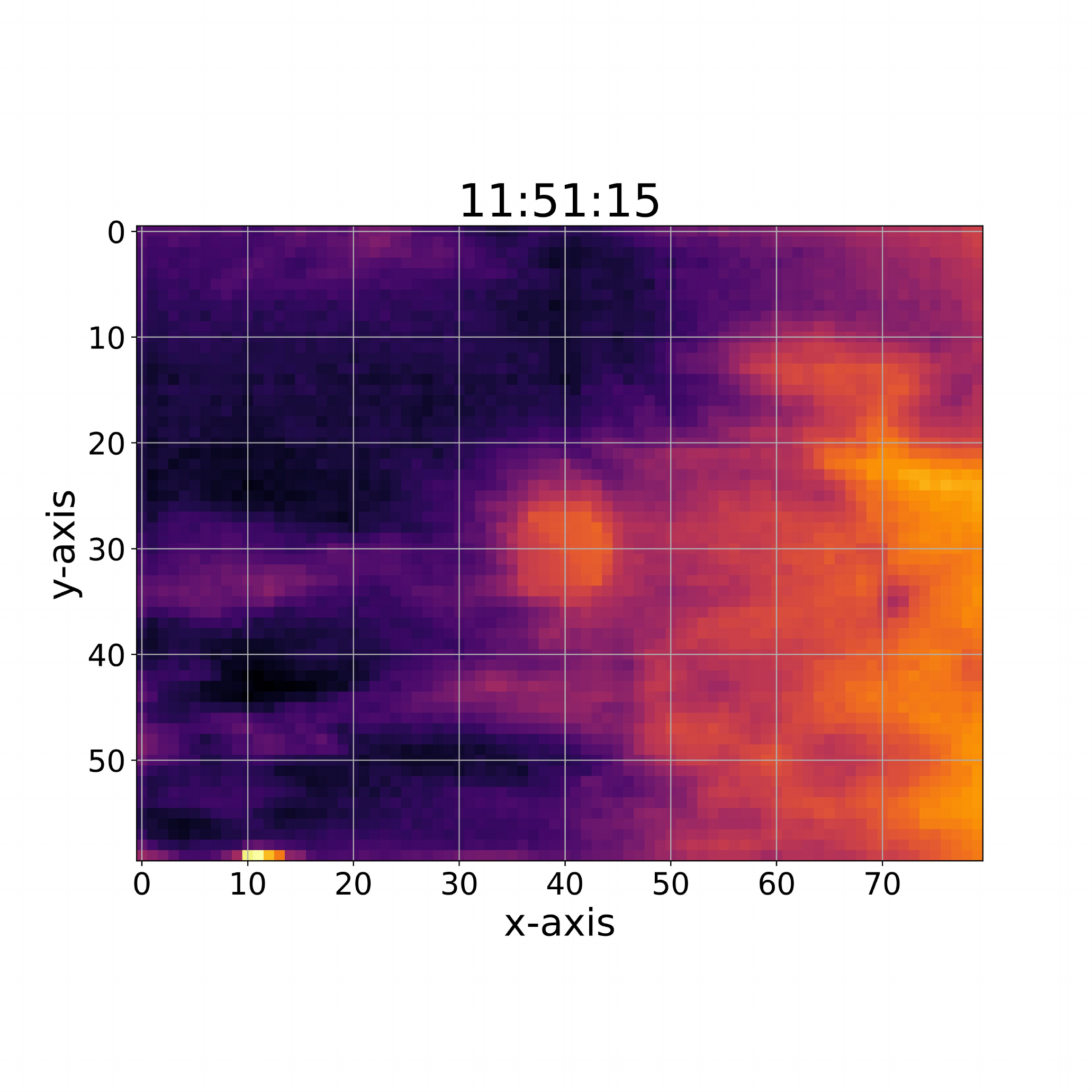}
        \includegraphics[scale = 0.225, trim = {1cm, 3.5cm, 2cm, 4.3cm}, clip]{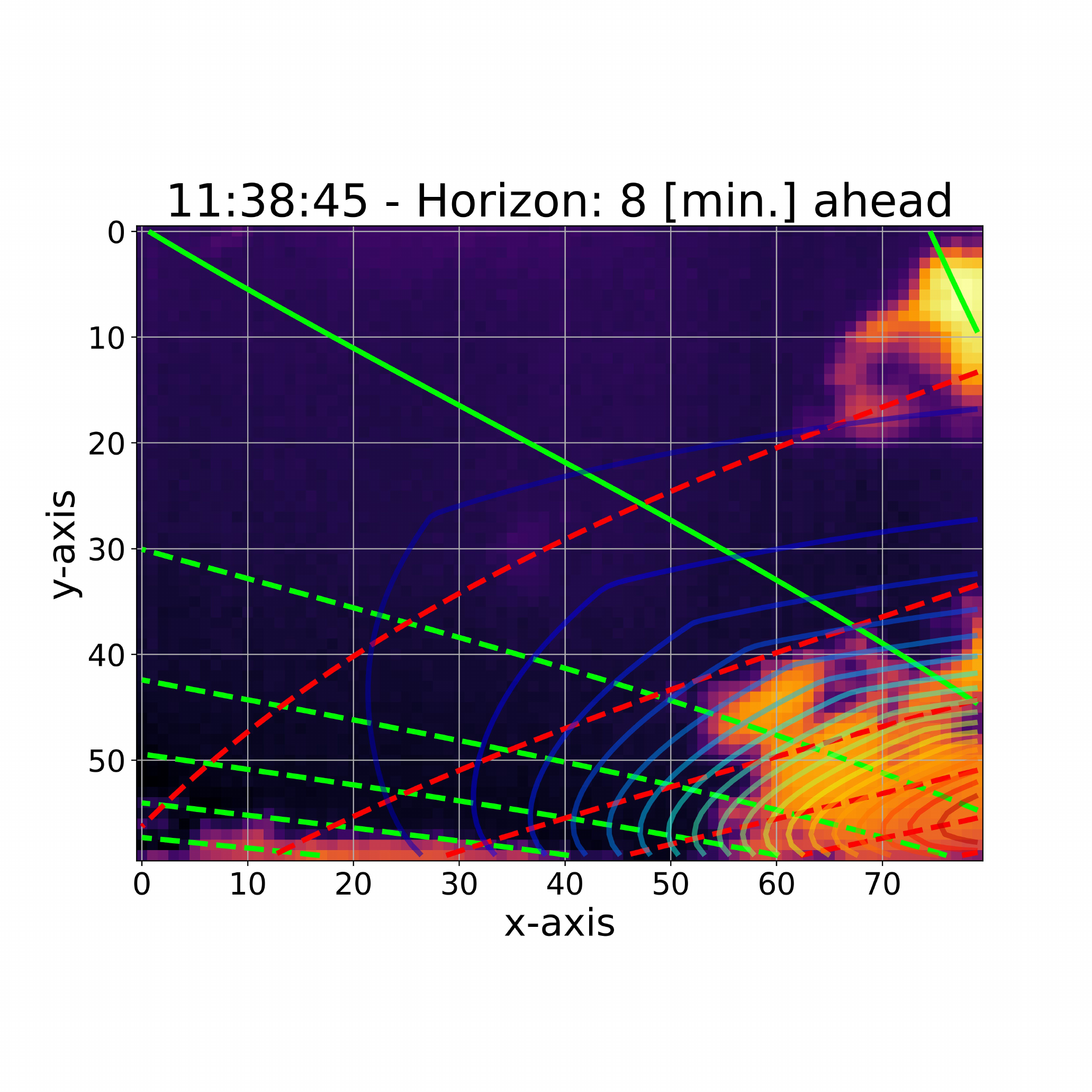}
        \includegraphics[scale = 0.225, trim = {1cm, 3.5cm, 2cm, 4.3cm}, clip]{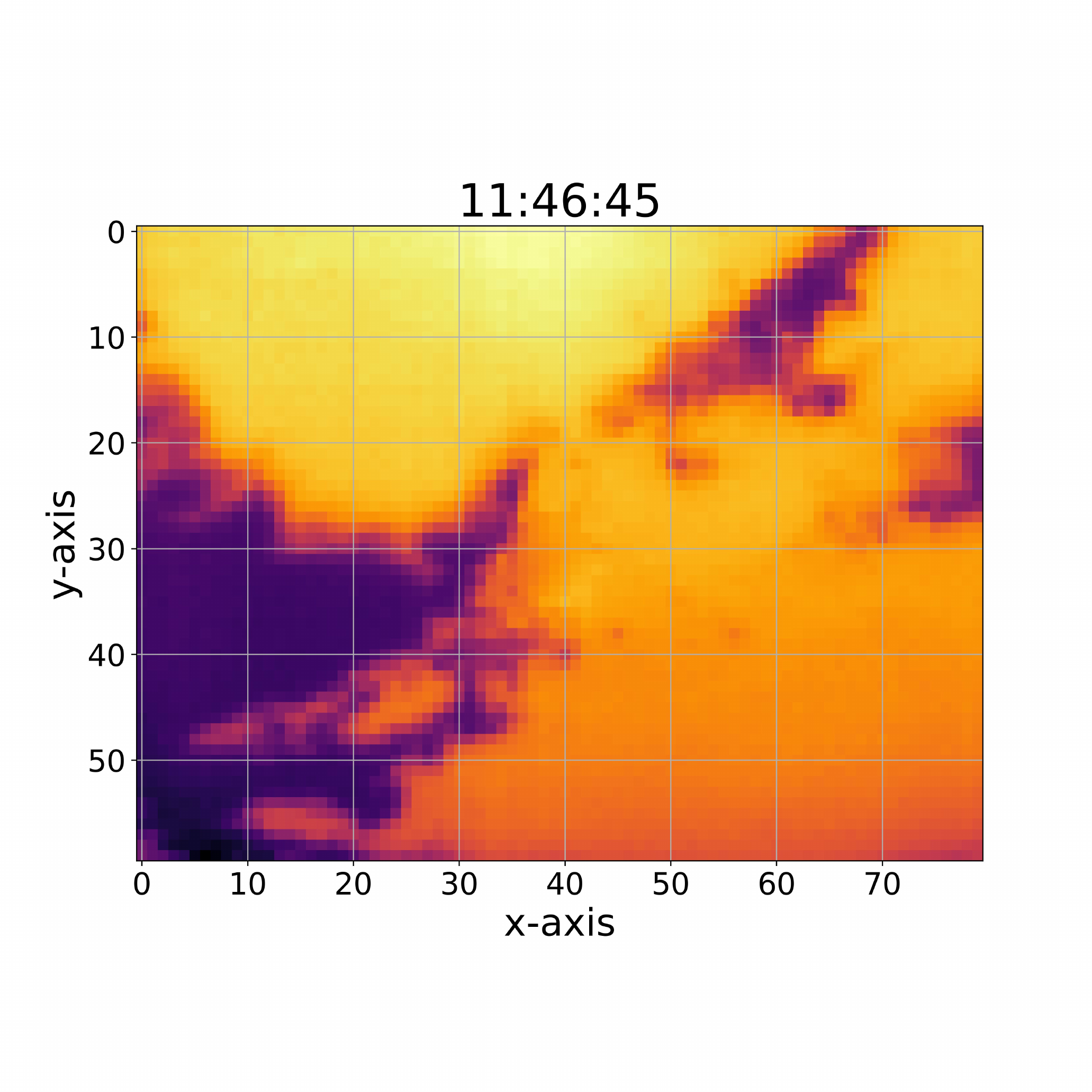}
    \end{subfigure}
    \caption{Demonstration of the feature selection algorithm in a clouds moving in low and high turbulent wind flows (see the extracted cloud features in Figure \ref{fig:feature_extraction_algorithm}). The streamlines (green) and potential lines (red) were computed for the frame shown in the first and third columns respectively. The probability of a pixel intersecting with the Sun (contours in blue to red color gradient) is computed for the proposed forecasting horizons (3, 4, 5, 6, 7 and 8 minutes ahead). The IR images in the second and fourth columns show the frames in the sequence that correspond to the forecasting horizon displayed in the image next to it in the same row.}
    \label{fig:feature_selection_algorithm}
\end{figure}

The probability of a pixel in the image intersecting the Sun given a forecasting horizon is computed for various forecasting horizons using the extracted features. Two examples used in the selection algorithm are shown in Figure~\ref{fig:feature_selection_algorithm} that corresponds to the clouds in Figure~\ref{fig:feature_extraction_algorithm}.

The feature extraction and selection algorithm was applied to consecutive sequences of images acquired on $52$ different days. The consecutive sequences show different sky conditions each day. An atmospheric condition model categorizes the IR sky images among the categories of clear sky, cumulus, stratus and nimbus clouds \cite{TERREN2021a}. The days with sequences of images only showing clear sky or nimbus clouds were avoided in this research since these are not of interest for solar energy generation. 

\subsection{Training and Testing Datasets}

Out of the entire dataset, $80$\% percent of the data was used for training ($36,932$ samples) and validation purposes while the remaining $20$\% ($9,233$ samples) of the data was used for testing. In other words, out of the 52 days in the dataset, $44$ days were used for training and $8$ days for testing. The samples in the dataset were grouped in the four categories of the atmospheric conditions model. The testing samples include $3,248$ clear sky samples, $2,143$ cumulus clouds samples, $1,596$ stratus clouds samples and $2,246$ nimbus clouds (i.e., large thick clouds) samples. The amount of data used in kernel learning methods is prohibitive due to the matrix inversion operation involved in GPRs, KRRs and RVMs. For comparison purposes the amount of data in the training was the same in each one model ($3,500$ samples). Multiple output models that use the Kronecker product require the inversion of a $NC \times NC$ dimensions matrix, where $N$ is the number of samples and $C$ is the number of  forecasting horizons. For this reason, the number of samples in the experiments carried out using the MT-KRR, MT-GPR, MT-RVM, and $\varepsilon$-MT-SVM is $2,500$. In this way, a model is trained using only samples of the same sky condition. Therefore, there are $4$ expert models per each implemented kernel learning method. For each testing samples first the class is detected (i.e., clear sky, stratus, cumulus, or nimbus clouds), and then the corresponding expert model is used for the forecasting.

\subsection{Data Preprocessing}

After dividing the training set to train the expert models in different sky condition categories, outliers in each category are removed. The method applied to detect outliers is the Local Outlier Factor (LOF) \cite{Breunig2000}, based on a k-nearest neighbour algorithm, whose optimal number of neighbors was cross-validated, and it was found to be $3$. The first $3,500$ samples with higher negative outlier factor were selected. 

The feature $\mathbf{x}_{i,j}$ and predictor $\mathbf{y}_{i,j}$ vectors were standardized as $\bar{\mathbf{x}}_{i,j} = [ \mathbf{x}_{i,j} - \mathbb{E} (\mathbf{X}) ] / \mathbb{V}^{1/2} (\mathbf{X})$ and $\bar{\mathbf{y}}_{i,j} = [ \mathbf{Y}_{i,j} - \mathbb{E} (\mathbf{Y}) ] / \mathbb{V}^{1/2} (\mathbf{Y})$. However, the feature vectors $\mathbf{x}_{i,j}$ used in a polynomial kernel did not require standardization, and the predictors $\mathbf{y}_{i,j}$ used in the GPRs and RVMs (i.e., Bayesian models) neither required standardization.

\subsection{Hyperparameters Cross-Validation}

\begin{figure}[!htb]
    \begin{subfigure}{\linewidth}
        \centering
        \includegraphics[scale = 0.25, trim = {1cm, 1cm, 0cm, 1.5cm}, clip]{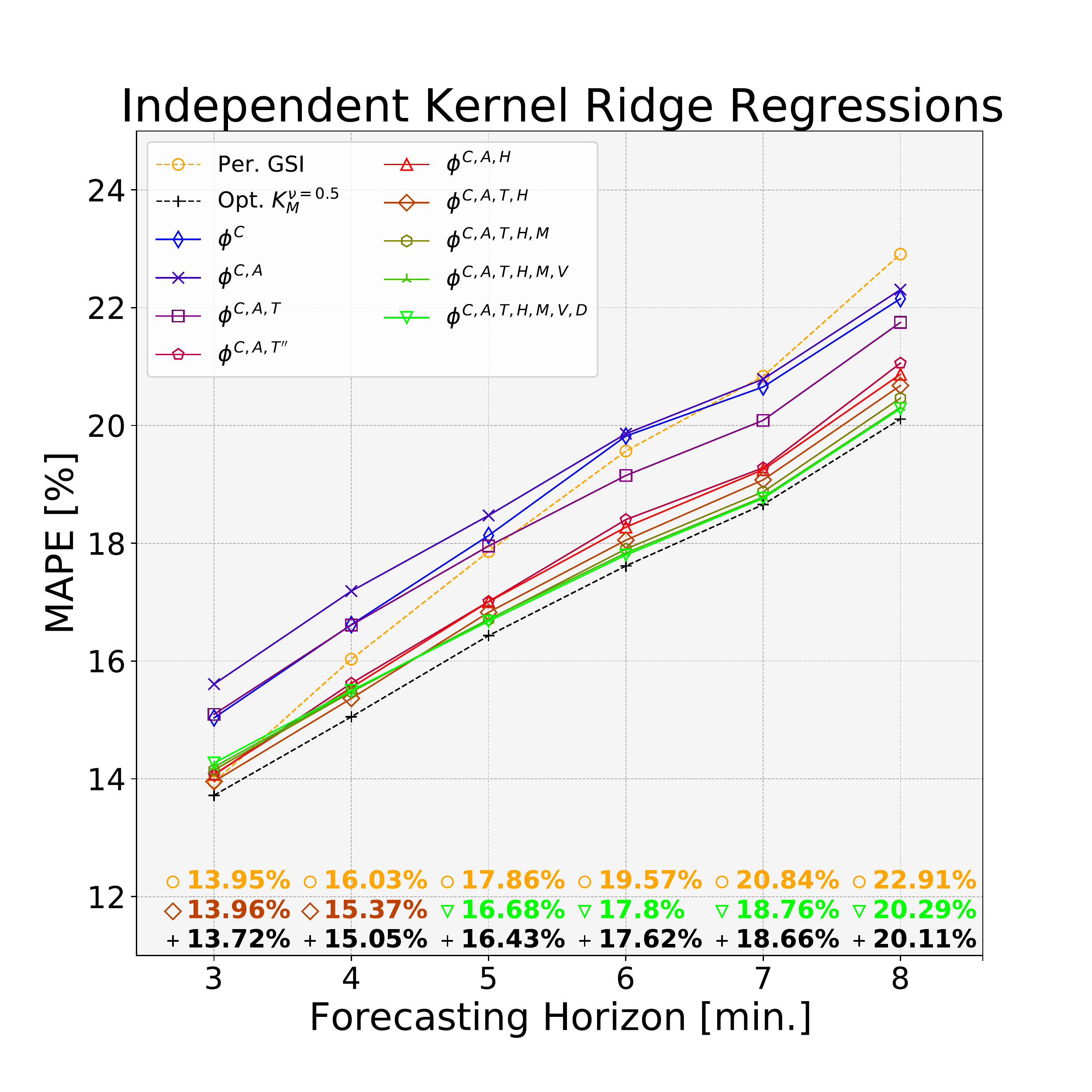}
        \includegraphics[scale = 0.25, trim = {1cm, 1cm, 0cm, 1.5cm}, clip]{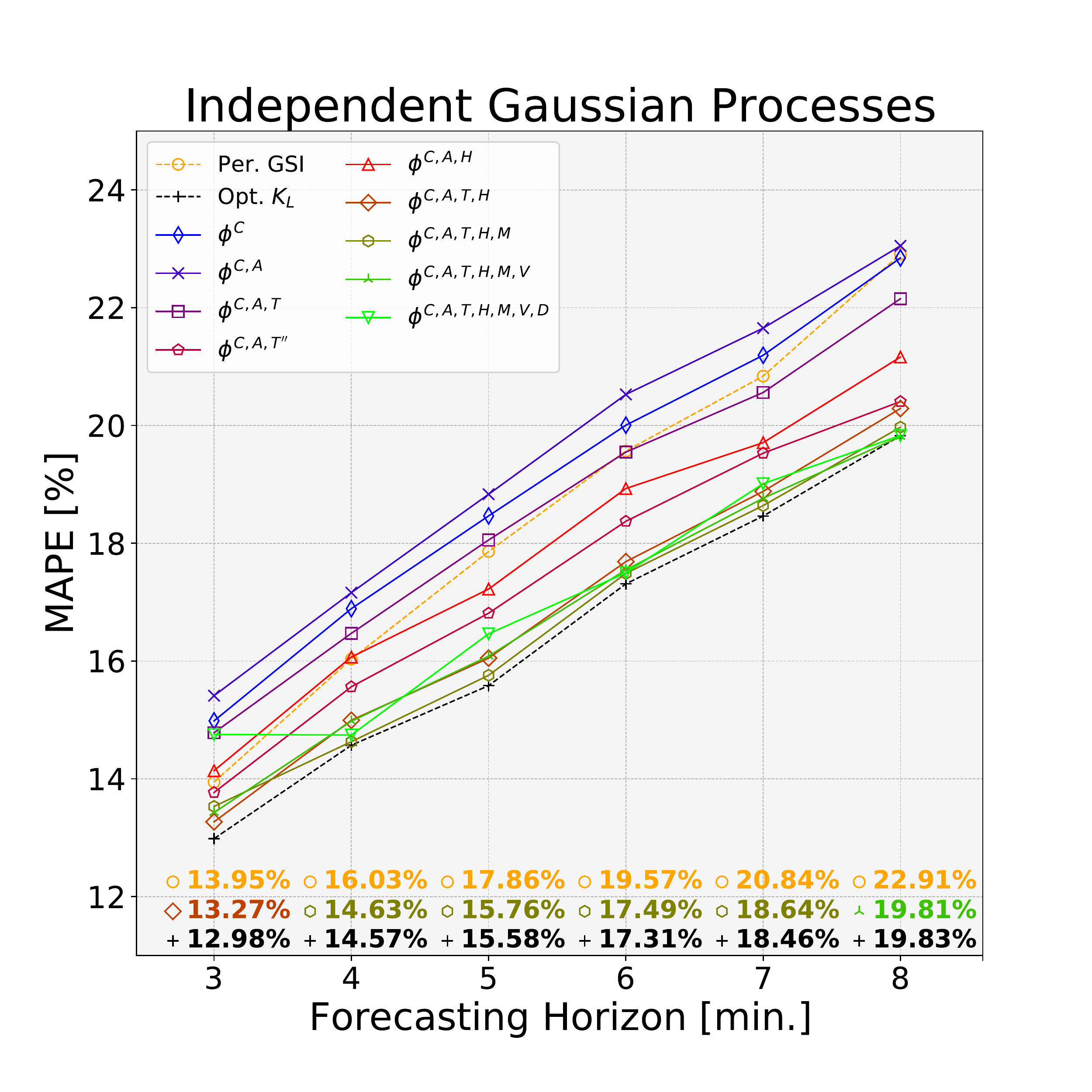}
    \end{subfigure}
    \begin{subfigure}{\linewidth}
        \centering
        \includegraphics[scale = 0.25, trim = {1cm, 1cm, 0cm, 1.5cm}, clip]{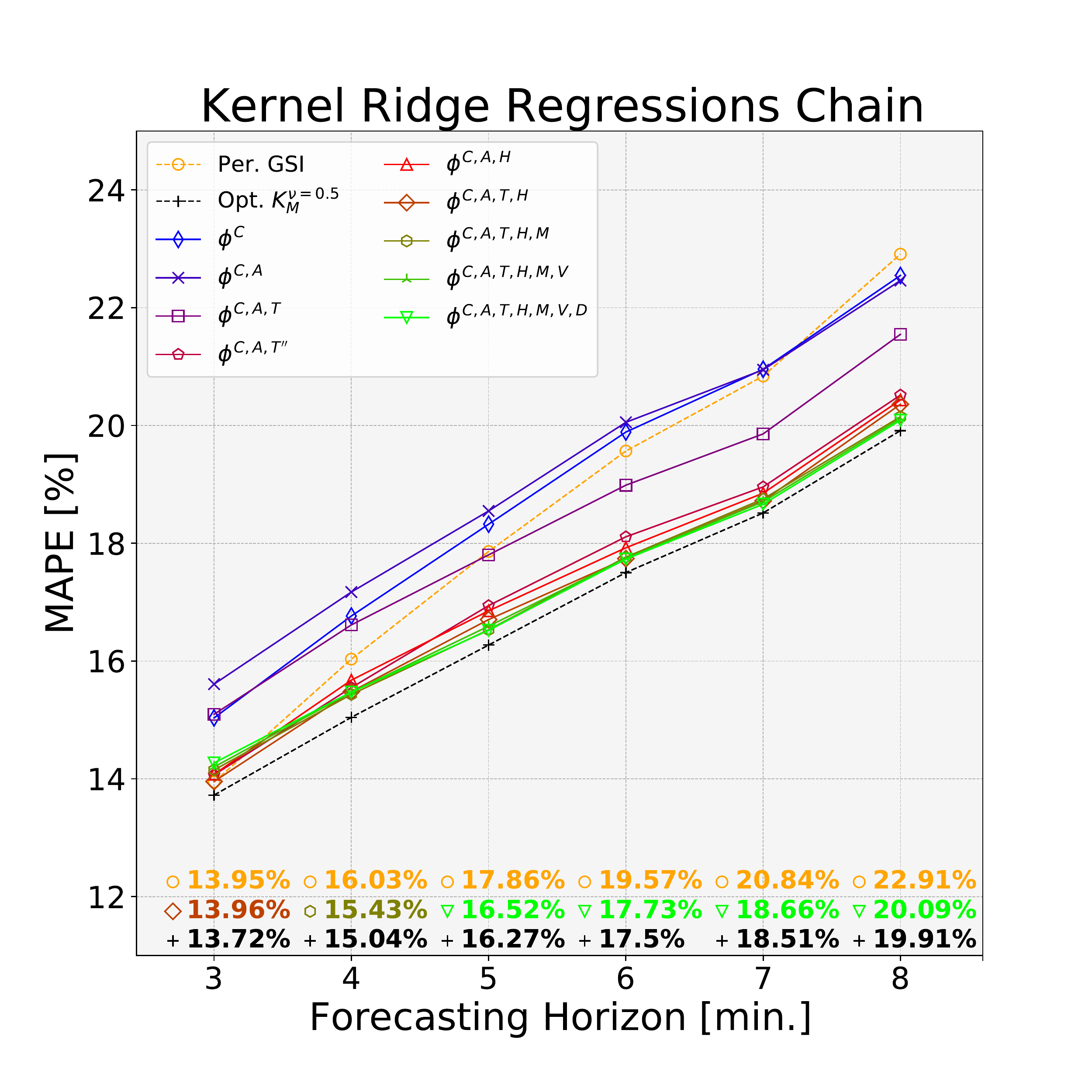}
        \includegraphics[scale = 0.25, trim = {1cm, 1cm, 0cm, 1.5cm}, clip]{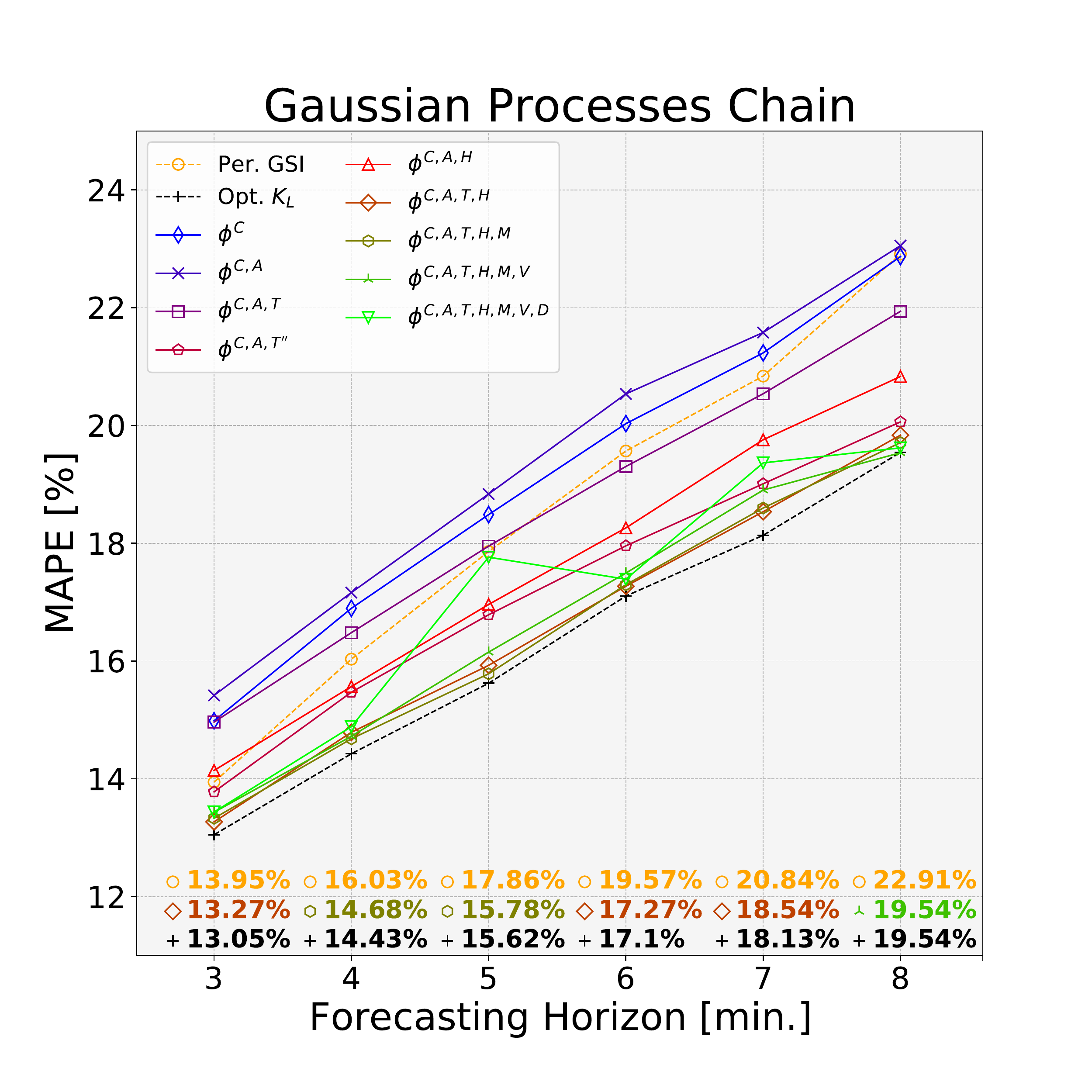}
    \end{subfigure}
    \begin{subfigure}{\linewidth}
        \centering
        \includegraphics[scale = 0.25, trim = {1cm, 1cm, 0cm, 1.5cm}, clip]{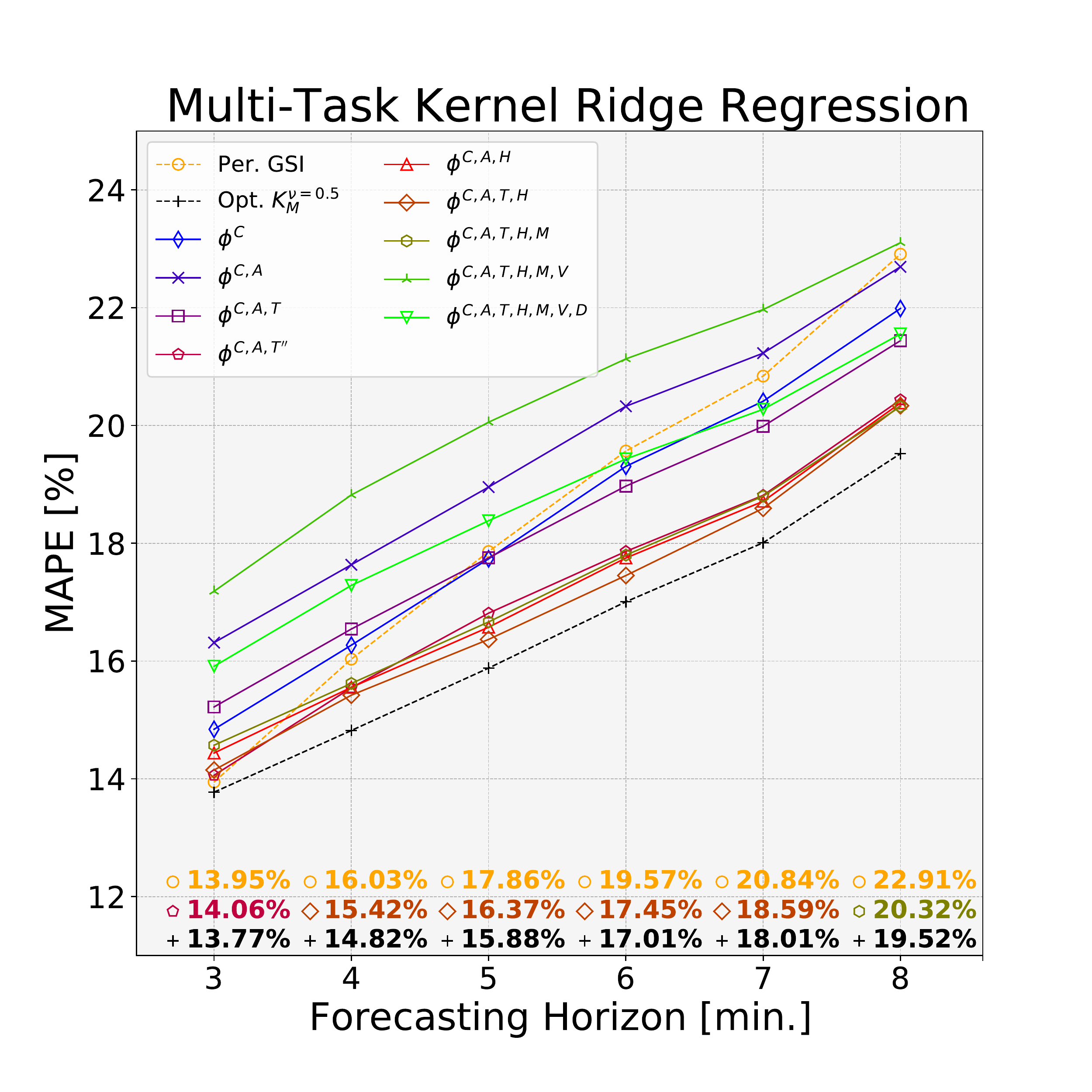}
        \includegraphics[scale = 0.25, trim = {1cm, 1cm, 0cm, 1.5cm}, clip]{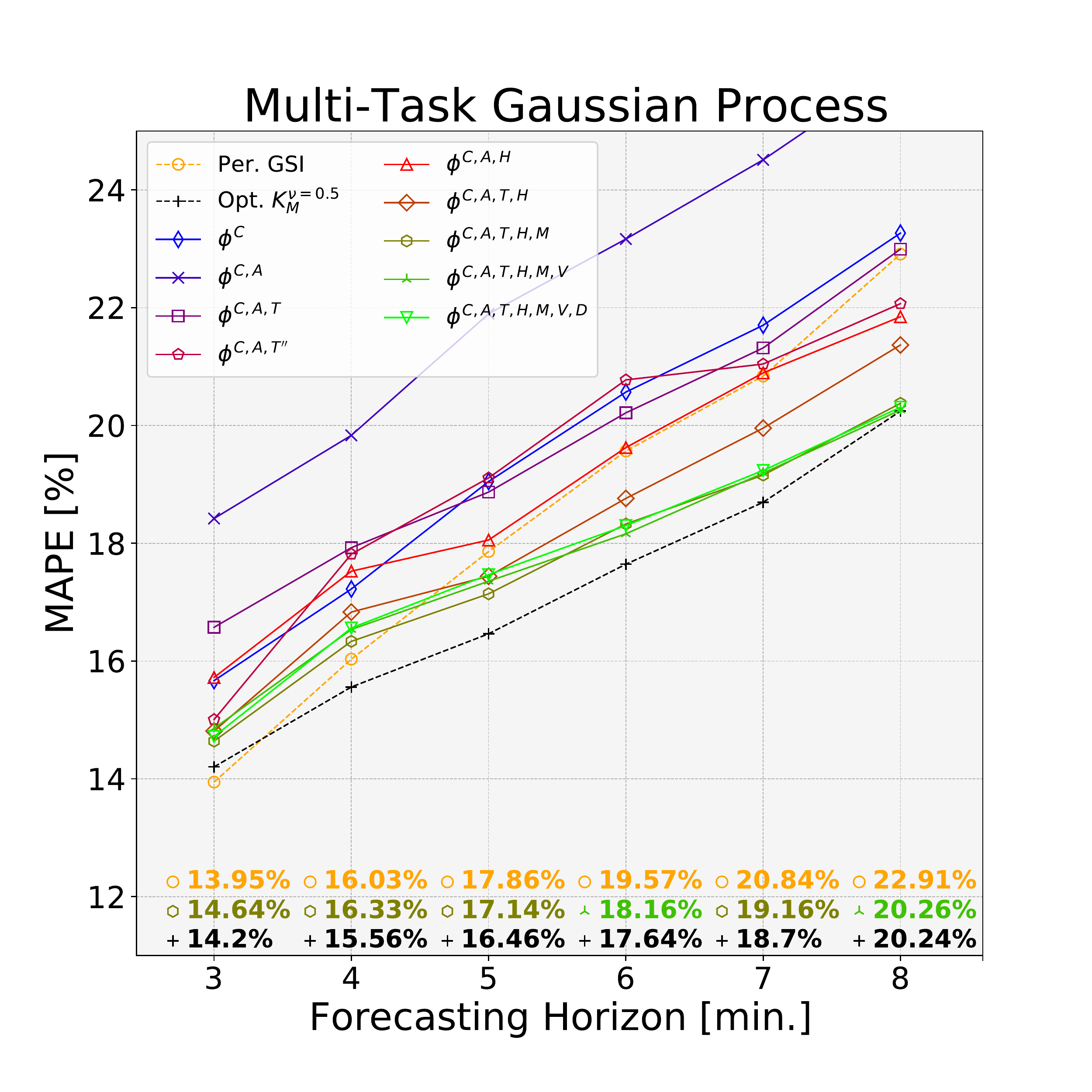}
    \end{subfigure}
    \caption{Forecasting accuracy achieved by the different multi-task dense kernel learning methods. The graphs in the right column shows the results obtained using KRRs (deterministic). The graphs in the left column shows the results obtained using GPRs (Bayesian). From top to bottom, the methods implemented to perform a multi-task forecast are multiple independent models, chain of models, and a multi-task models. The persistence model (i.e., baseline) is displayed in orange, the colors show the models using different feature vectors (combination of the same model for each sky condition), and optimal method is black (combination of the best model for each sky condition). The kernel used by the optimal model is denoted in the legend. In the bottom of the graphs, the MAPE of the baseline (circle in orange), best model (denoted by its corresponding color and marker shape), and optimal model (plus in black) are outline for each forecasting horizon.}
    \label{fig:dense_kernel_learning}
\end{figure}

The cross-validation routine implemented was a $3$-fold validation method. The implementation of another more intensive validation method was not possible due to computational time constraints. A grid search was performed to validate all possible sets of parameters. The size of the grid was 4. The parameters of the model and hyperparameters of the kernel are cross-validated for the KRRs, RVMs and $\varepsilon$-SVMs. In the case of GPRs, the noise variance parameter $\sigma^2_{c,n}$ and $\Sigma_n$ in Eq.~\eqref{eq:gaussian_process_prediction} and Eq.~\eqref{eq:multitask_gaussian_process_prediction} respectively, and the hyperparameters of the kernel (see Section \ref{sec:kernels}) are optimized via gradient maximization of the MLL in Eq.~\eqref{eq:gaussian_process_mll} and Eq.~\eqref{eq:multitask_gaussian_process_mll}. The optimal correlation matrix coefficients in Eq.~\eqref{eq:multitask_gaussian_process_prediction} of the MT-GPR model are also found maximizing the MLL. The forecasting test results in Mean Absolute Percentage Error (MAPE) performed by independent GPRs, chain of GPR and MT-GPR are shown in Figure~\ref{fig:dense_kernel_learning}. The experiments have been conducted for different combinations of features. Each feature vector used in the figure is represented by a symbol $\phi$ with superindexes denoting the different features as $C$: Clear Sky Index, $A$: elevation and azimuth angles, $T$: raw temperature, $T^{\prime\prime}$: processed temperature, $H$: processed height $M$: magnitude  $V$: vorticity, and $D$: diverence of the velocity vectors.

The regularization parameter $\gamma_c\mathbf{I}_{N \times N}$ in the KRR in Eq.~\eqref{eq:kernel_ridge_regression} and the kernel hyperparameters (see Subsection \ref{sec:kernels}) require cross-validation for each forecasting horizon $c$. However, the regularization parameter $\gamma_c \mathbf{I}_{CN \times CN}$ in the MT-KRR in Eq.~\eqref{eq:multiouput_kernel_ridge_regression} is simplified as $\gamma_1 = \dots = \gamma_C = \gamma$ in order to reduce the computational cost of the cross-validation procedure. Similarly, the parameters in the correlation matrix $\boldsymbol{\Gamma}_{C \times C}$ in Eq.~\eqref{eq:multiouput_kernel_ridge_regression_prediction} are also simplified as explained in Subsection \ref{sec:multi-task Kernel Simplification}. Figure~\ref{fig:dense_kernel_learning} shows the testing MAPE obtained by independent KRRs, a chain of KRRs and MT-KRR.

\begin{figure}[!htb]
    \begin{subfigure}{\linewidth}
        \centering
        \includegraphics[scale = 0.25, trim = {1cm, 1cm, 0cm, 1.5cm}, clip]{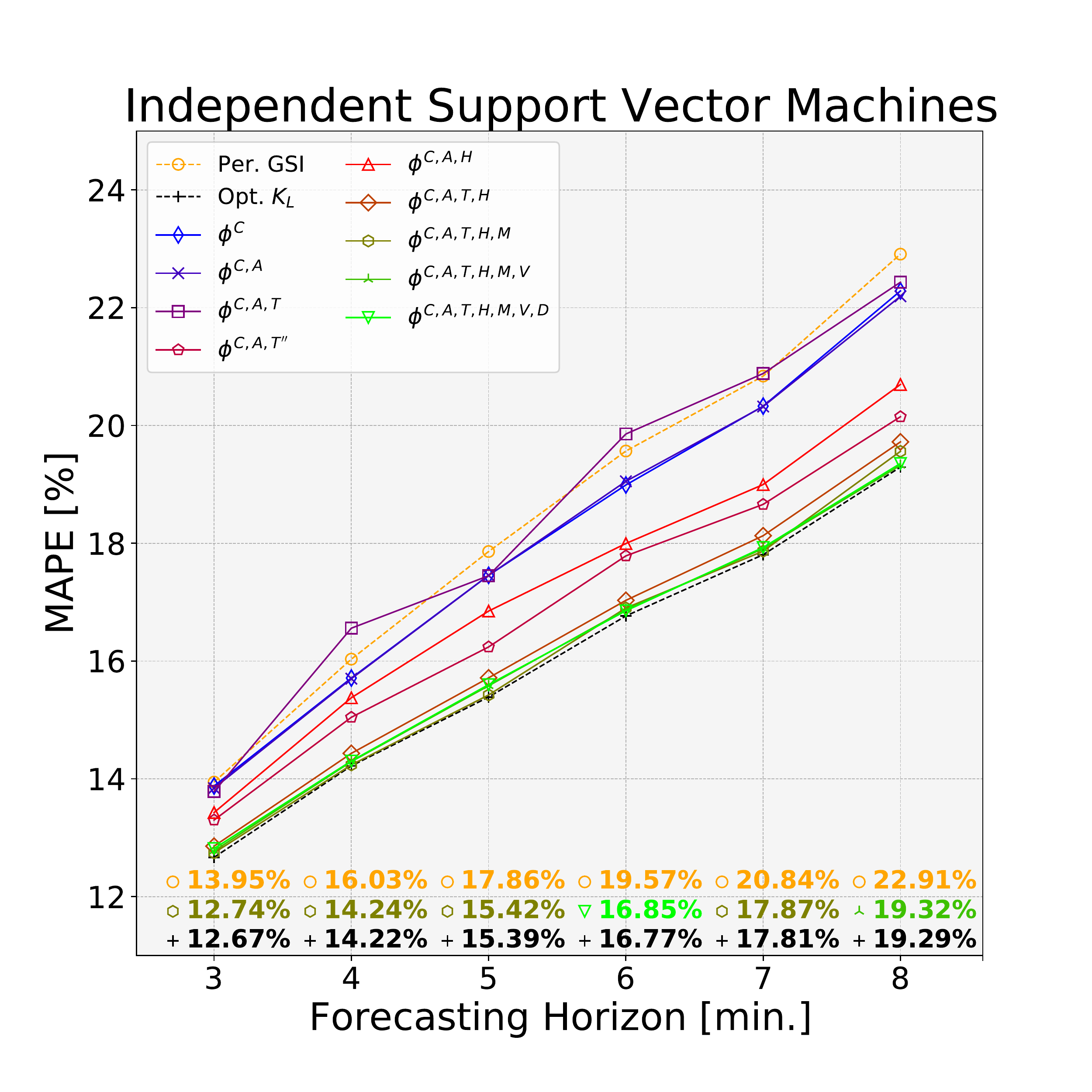}
        \includegraphics[scale = 0.25, trim = {1cm, 1cm, 0cm, 1.5cm}, clip]{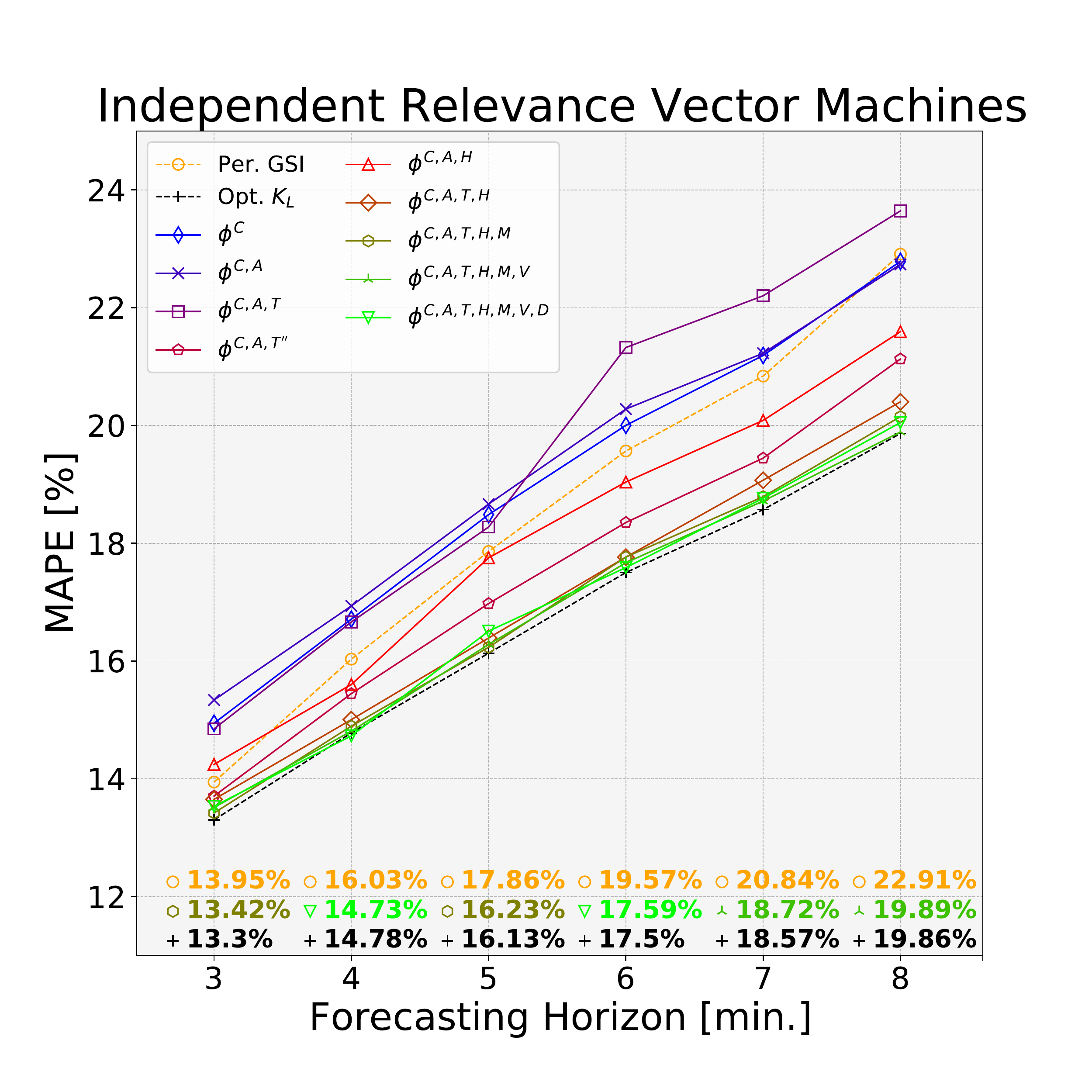}
    \end{subfigure}
    \begin{subfigure}{\linewidth}
        \centering
        \includegraphics[scale = 0.25, trim = {1cm, 1cm, 0cm, 1.5cm}, clip]{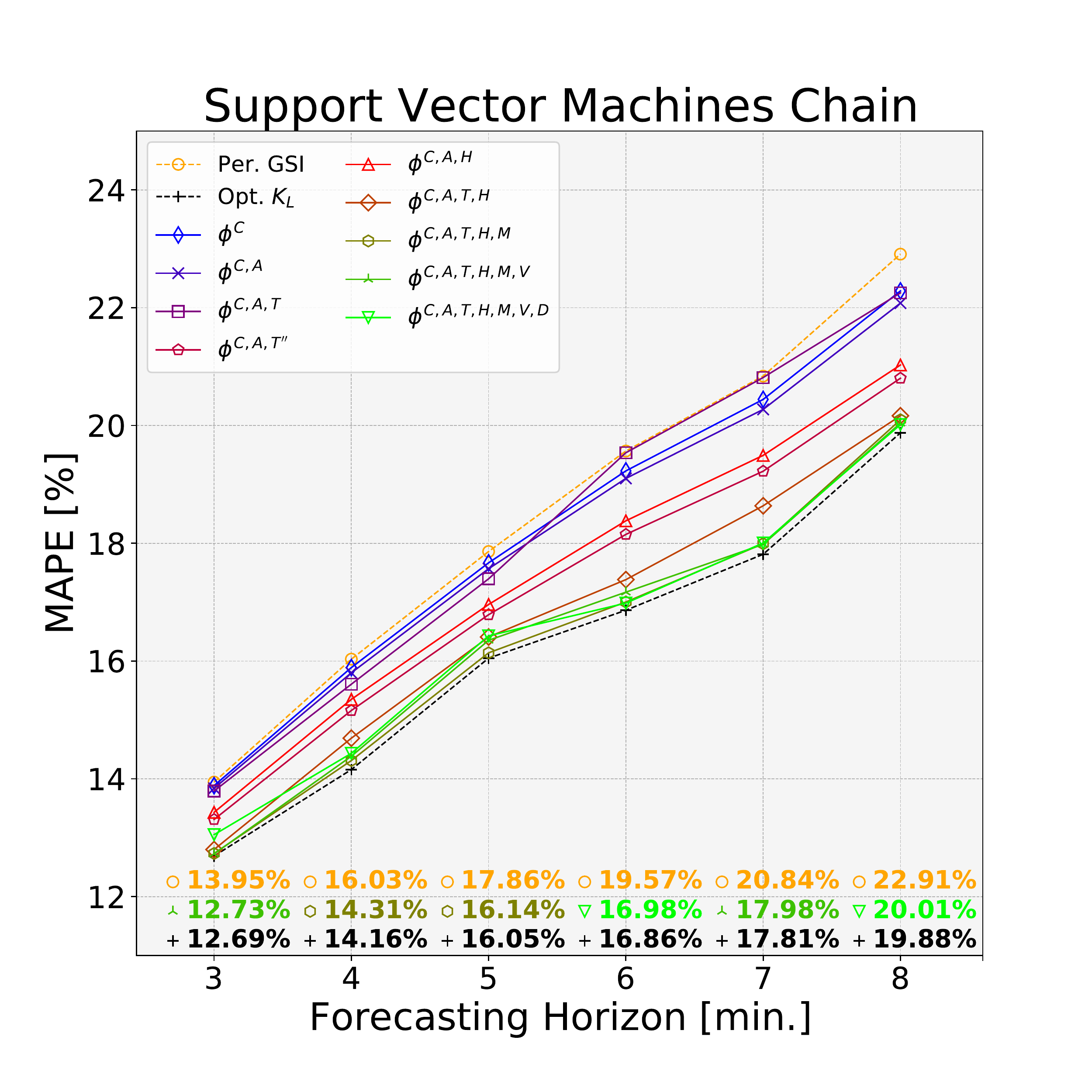}
        \includegraphics[scale = 0.25, trim = {1cm, 1cm, 0cm, 1.5cm}, clip]{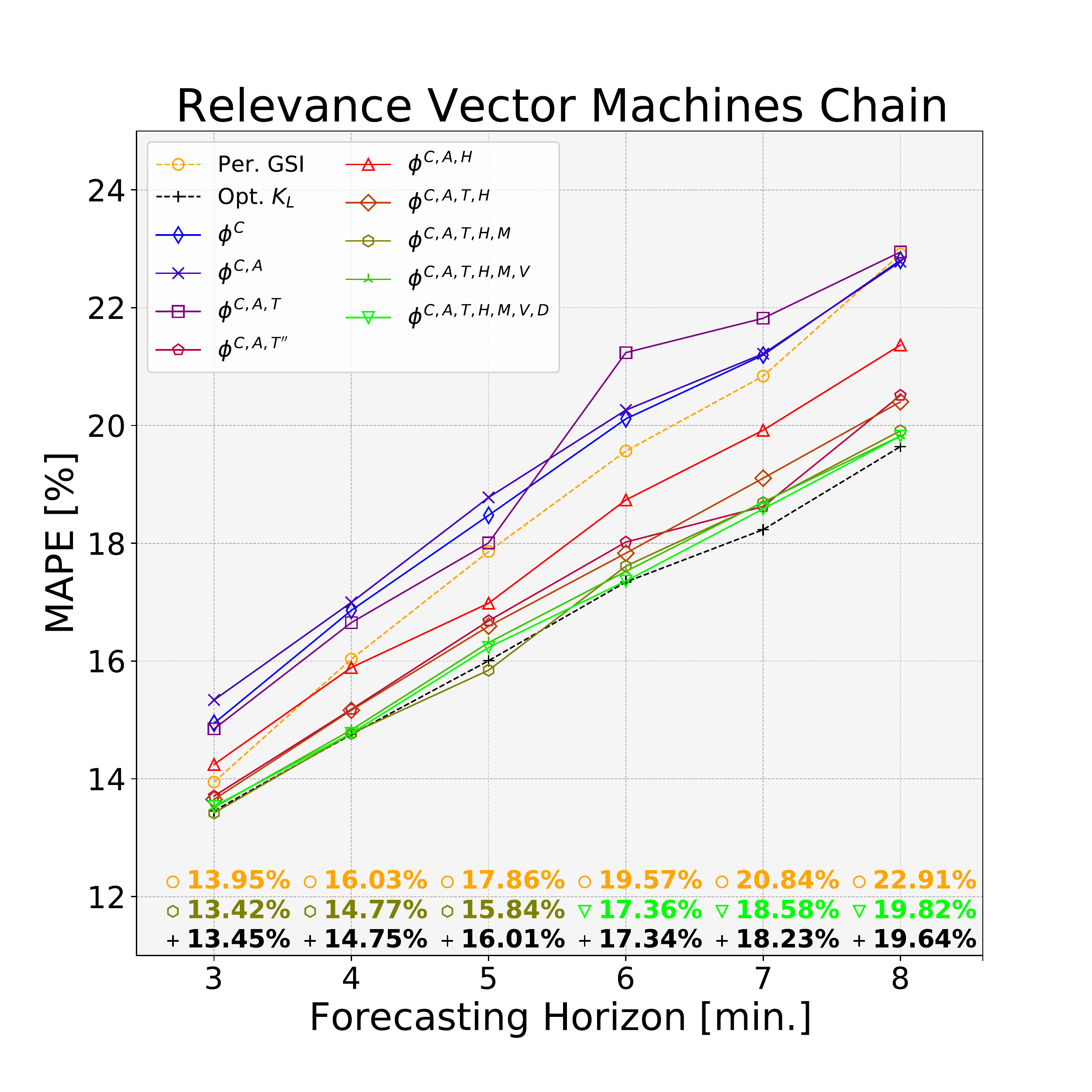}
    \end{subfigure}
    \begin{subfigure}{\linewidth}
        \centering
        \includegraphics[scale = 0.25, trim = {1cm, 1cm, 0cm, 1.5cm}, clip]{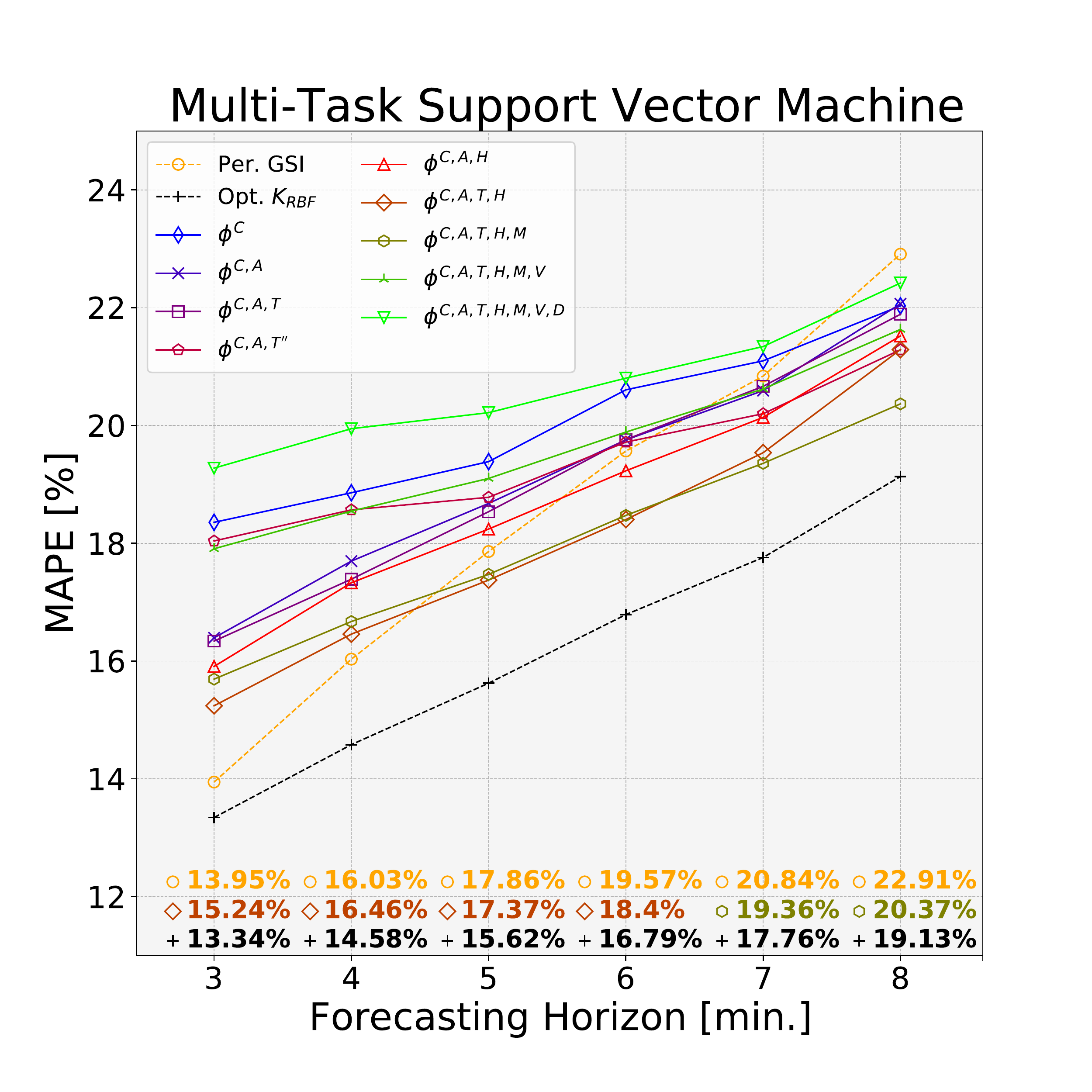}
        \includegraphics[scale = 0.25, trim = {1cm, 1cm, 1cm, 1cm}, clip]{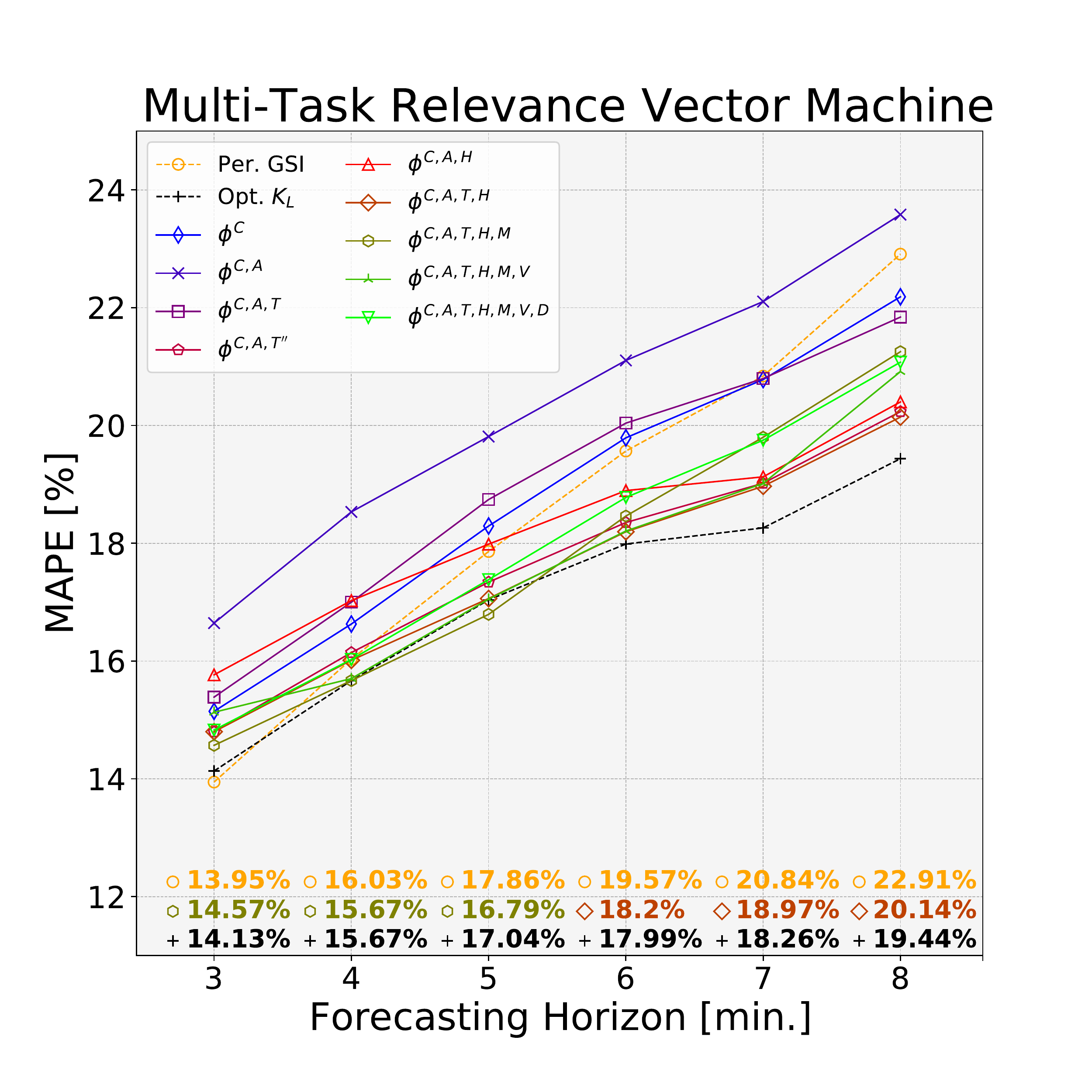}
    \end{subfigure}
    \caption{Forecasting accuracy achieved using sparse kernel learning methods. The graph on the left shows the performances of the SVMs (deterministic). The graph on the left shows the performances of the RVMs (Bayesian). From top to bottom, the graphs kernel learning methods are organized as independent models, chain of models, and multi-task models. The persistence model is shown in orange (i.e. baseline), in the different colors are shown the performance for each feature vector when the same model is used in the four different sky conditions. The optimal model composed by the best model for each sky condition is shown in black. The kernel used by the optimal model is denoted in the legend. In the bottom of the graphs, the MAPE of the baseline (circle in orange), best model (denoted by the corresponding color and marker shape), and optimal model (plus in black) are outline for each forecasting horizon.}
    \label{fig:sparse_kernel_learning}
\end{figure}

The complexity $\mathcal{C}$ and $\varepsilon$ parameter in Eq.~\eqref{eq:support_vector_machine_dual}, plus the hyperparameters of the kernel in the $\varepsilon$-SVMs require cross-validation for each forecasting horizon. However, the parameter $\mathcal{C}$ and $\varepsilon$ in Eq.~\ref{eq:multitask_support_vector_machine_dual}, and the hyperparameters of kernel in the $\varepsilon$-MT-SVM are the same for all forecasting horizons (see Subsection \ref{sec:support_vector_machine}), but notice that the $\varepsilon$-MT-SVM have different bias $b_c$ (see Eq.~\eqref{eq:multitask_support_vector_machine_prediction}) for each forecasting horizon as the independent $\varepsilon$-SVMs and the chain of $\varepsilon$-SVMs. The correlation matrix $\boldsymbol{\Gamma}_{C \times C}$ in Eq.~\eqref{eq:multitask_support_vector_machine_prediction}, is simplified in the same way proposed for the MT-KRR. The testing results achieved by independent $\varepsilon$-SVMs, chain of $\varepsilon$-SVMs and $\varepsilon$-MT-SVM are in Figure~\ref{fig:sparse_kernel_learning}.

RVMs do not require the cross-validation of any model parameters (see Subsection \ref{sec:relevance_vector_machine}). Nevertheless, a convergence criterion has to be established to determine when the optimal relevance vectors were found. The converge criterion implemented in this investigation consists of stopping the optimization when a new minimum in the sum of squared residuals is not achieved after $25$ iterations. The kernel hyperparameters do require cross-validation (see Subsection \ref{sec:kernels}) for each forecasting horizon. The MT-RVM parameters of the correlation matrix $\boldsymbol{\Gamma}_{C \times C}$ in Eq.~\eqref{eq:multitask_relevance_vector_machine_prediction} requires cross-validation but the matrix is simplified similarly to MT-KRR and MT-SVM. Figure~\ref{fig:sparse_kernel_learning} shows the testing results obtained by independent RVMs, chain of RVMs and MT-RVM. 

% Local Outlier parameters cross-validation

The optimal number of \emph{sources} (i.e., sectors or density functions) included in the feature vector of each forecasting was cross-validated using the $3$-Fold validation method. The cross-validation was performed using the $3,500$ samples training dataset. The independent GPRs method was chosen as a forecasting model in order to avoid the burden of the model parameters and kernel hyperparameters cross-validation. The optimal indexes of \emph{sources} $c^\prime$ included in the feature vector $\mathbf{x}^{c}_{c^\prime,k}$ of each forecasting horizon $c$ are: $\mathbf{x}^1_{c^\prime,k}, \ c^\prime \in \{1, 2, 3, 4\}$; $\mathbf{x}^2_{c^\prime,k}, \ c^\prime \in \{1, 2, 3\}$; $\mathbf{x}^3_{c^\prime,k}, \ c^\prime \in \{2, 3, 4\}$, $\mathbf{x}^4_{c^\prime,k}, \ c^\prime \in \{3, 4, 5\}$; $\mathbf{x}^5_{c^\prime,k}, \ c^\prime \in \{3, 4, 5, 6\}$; and $\mathbf{x}^6_{c^\prime,k}, \ c^\prime \in \{5, 6\}$. They were selected for performing the highest MAPE after averaging the results for across the expert models. The feature vectors $\mathbf{x}^{c}_{c^\prime,k}$ of the multi-tasks models include the indexes of all \emph{sources} $c^\prime \in \{1, 2, 3, 4, 5, 6\}$. The feature vector included features from the CSI, angles, raw temperature and processed height. The procedure to determine the optimal number of neighbors in the LOF algorithm was the same, but the feature vectors include all forecasting horizons.

\begin{figure}[!htb]
    \begin{subfigure}{\linewidth}
        \centering
        \includegraphics[scale = 0.3, trim = {0cm, 0cm, 0cm, 0cm}, clip]{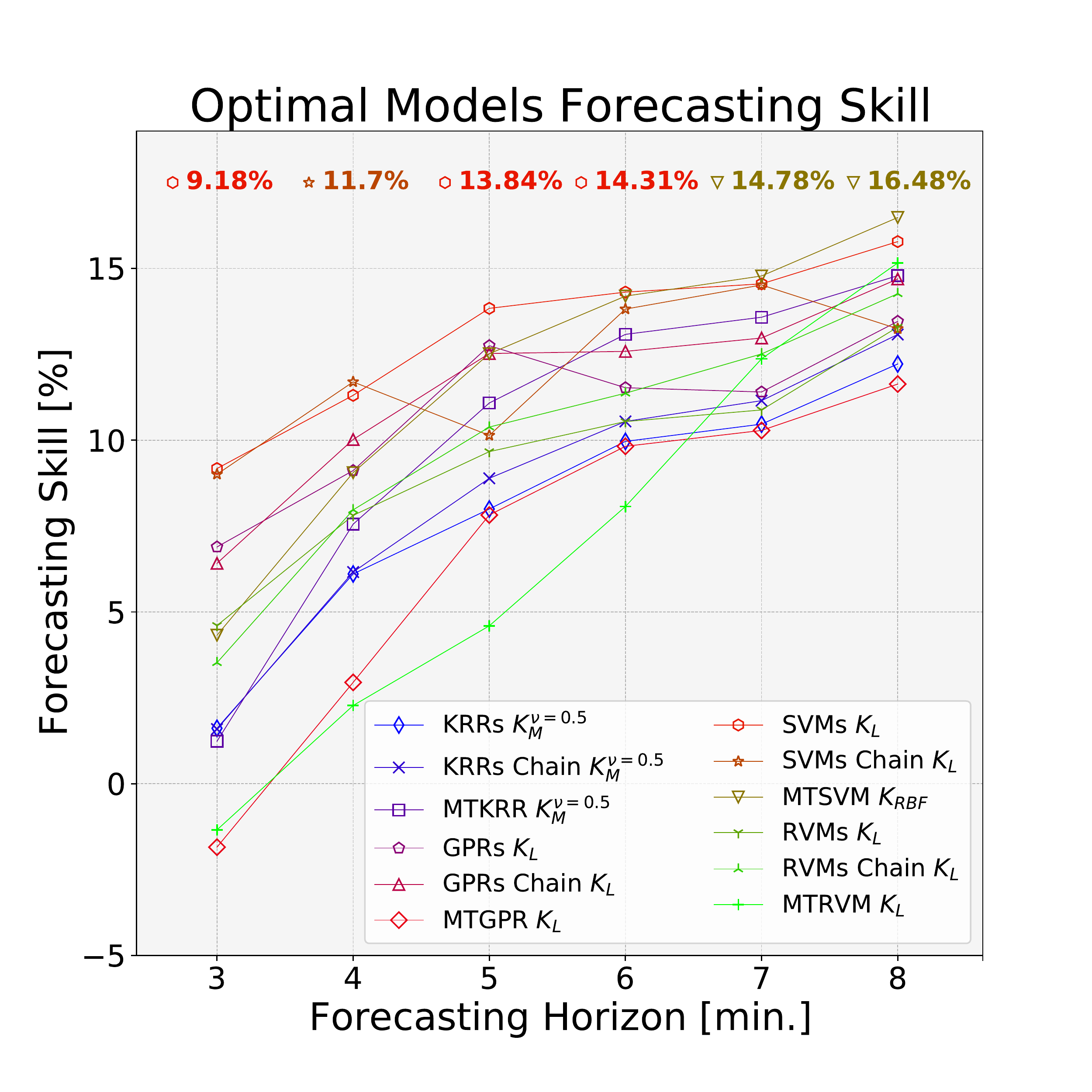}
        \includegraphics[scale = 0.3, trim = {0cm, 0cm, 0cm, 0cm}, clip]{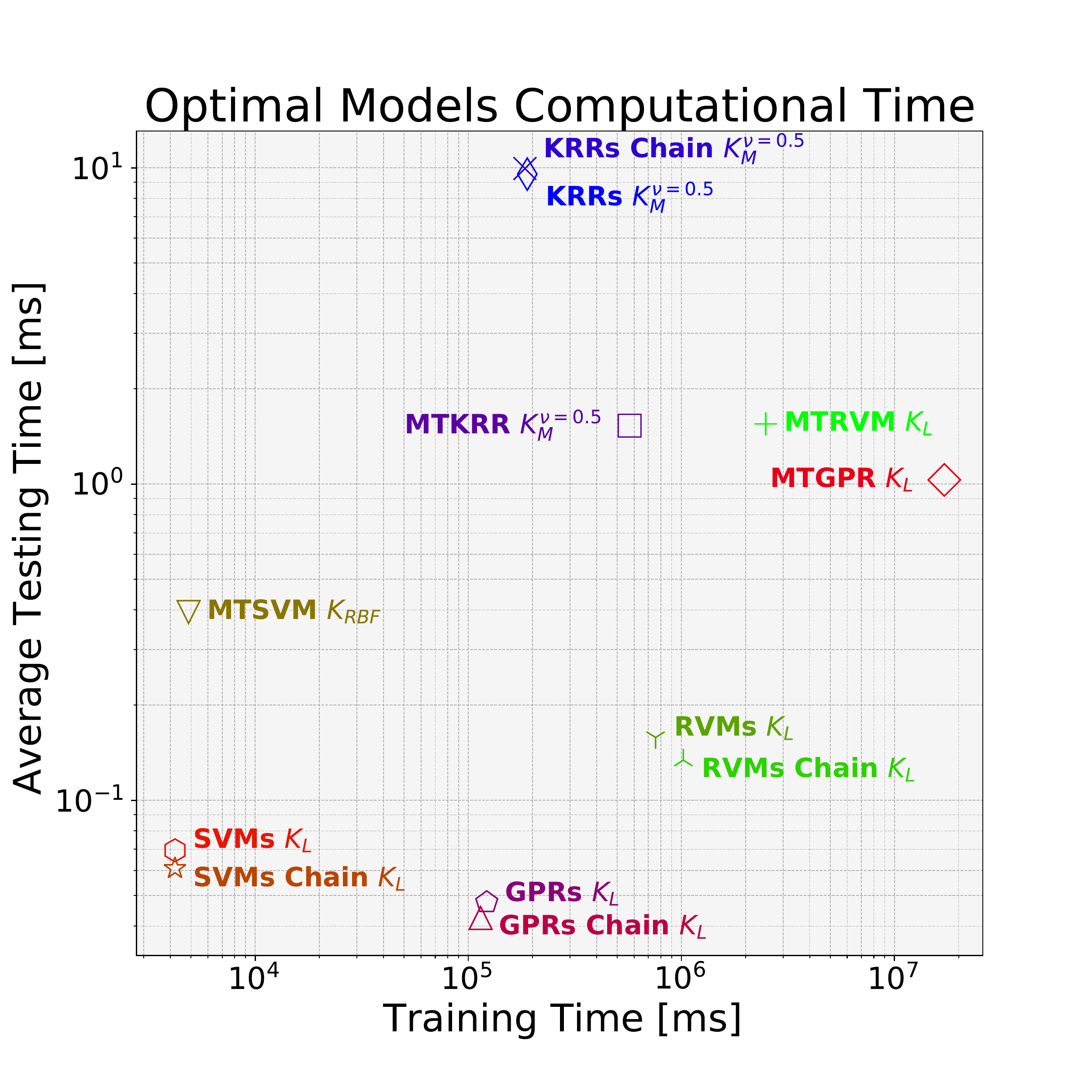}
    \end{subfigure}
    \caption{Summary of the optimal models performances. The graph in the left shows the FS achieved by the optimal models in each forecasting horizon. The graph in the right shows the training and testing time of the optimal models. In the case of independent regressions and chain of regressions, the training time is the sum of each conditions models training time. The testing time is the average time to perform a forecasting (i.e., $6$ horizons) of the four different models (i.e., sky conditions). The kernel used by the optimal models is denoted next to the models' name. In the top of the left graph, the higher FS is outline using maker shape and color of the model that achieved it.}
    \label{fig:forecasting_skill_summary}
\end{figure}

The experiments were carried out in the Wheeler high performance computer of the UNM Center for Advanced Research Computing (CARC), which uses a SGI AltixXE Xeon X5550 at 2.67GHz with 6GB of RAM memory per core, 8 cores per node, 304 nodes total, and runs at 25 theoretical peak FLOPS. Linux CentOS 7 is installed.

\begin{figure}[!htb]
    \begin{subfigure}{\linewidth}
        \centering
        \includegraphics[scale = 0.275, trim = {1.5cm, 0cm, 2.5cm, 0cm}, clip]{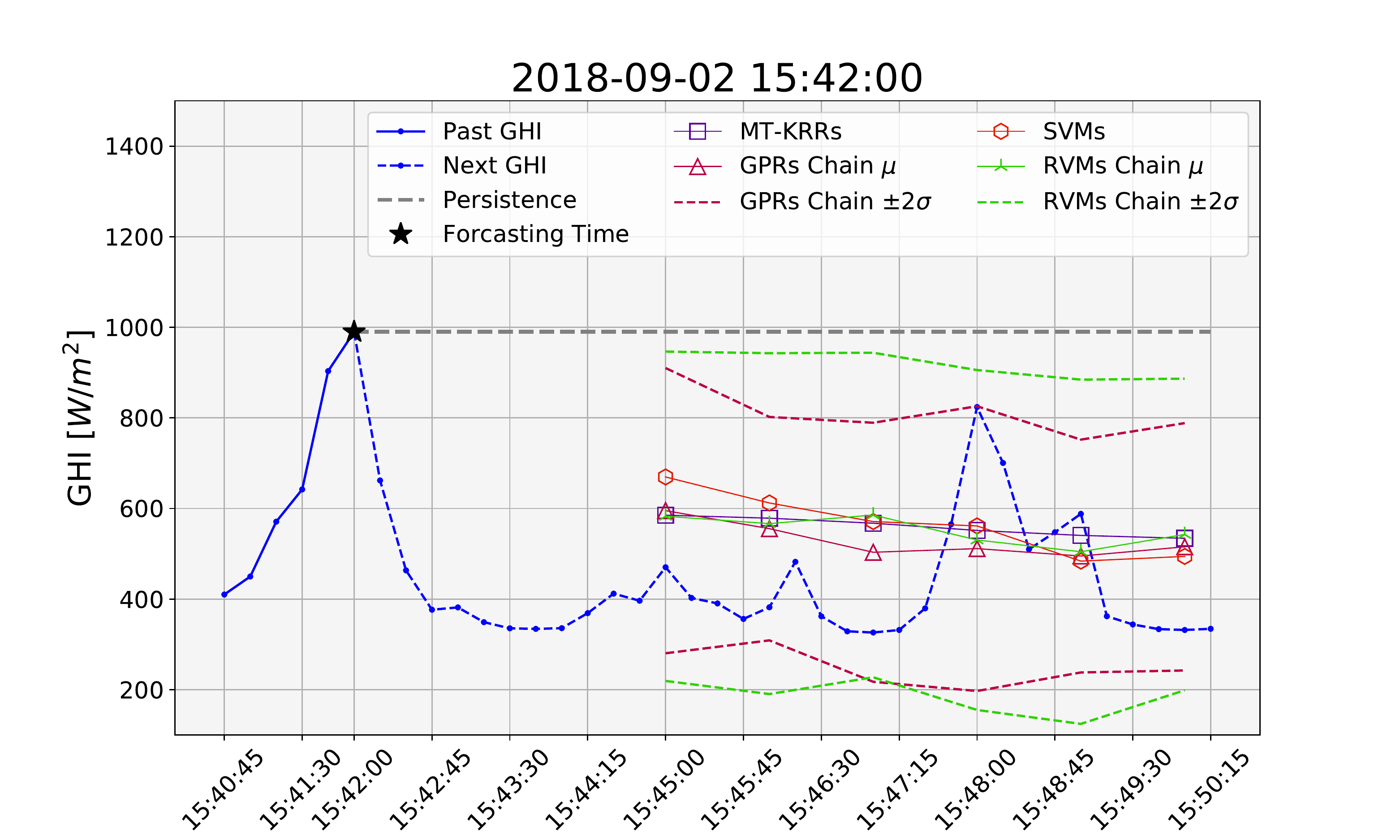}
        \includegraphics[scale = 0.275, trim = {1.5cm, 0cm, 2.5cm, 0cm}, clip]{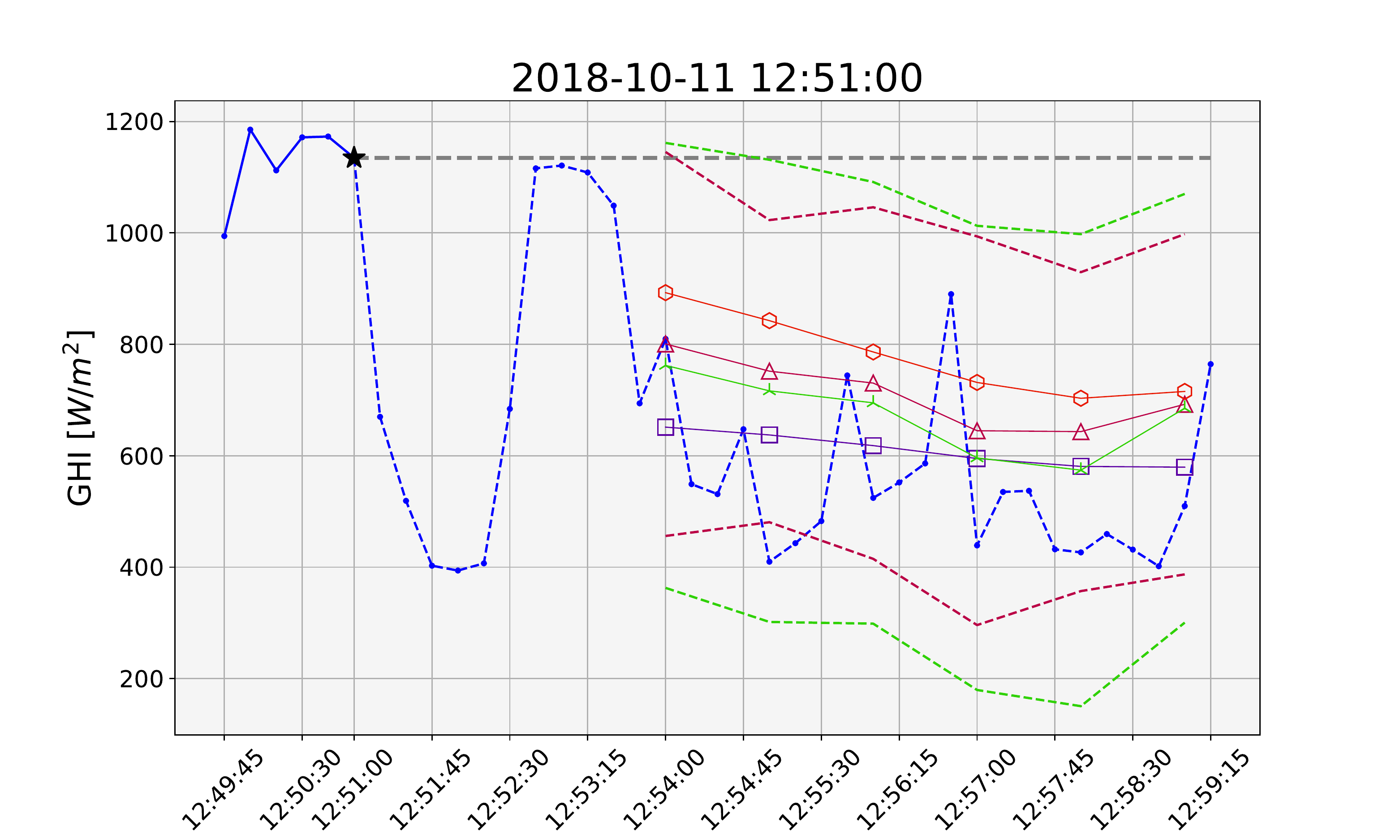}
    \end{subfigure}
    \begin{subfigure}{\linewidth}
        \centering
        \includegraphics[scale = 0.275, trim = {1.5cm, 0cm, 2.5cm, 0cm}, clip]{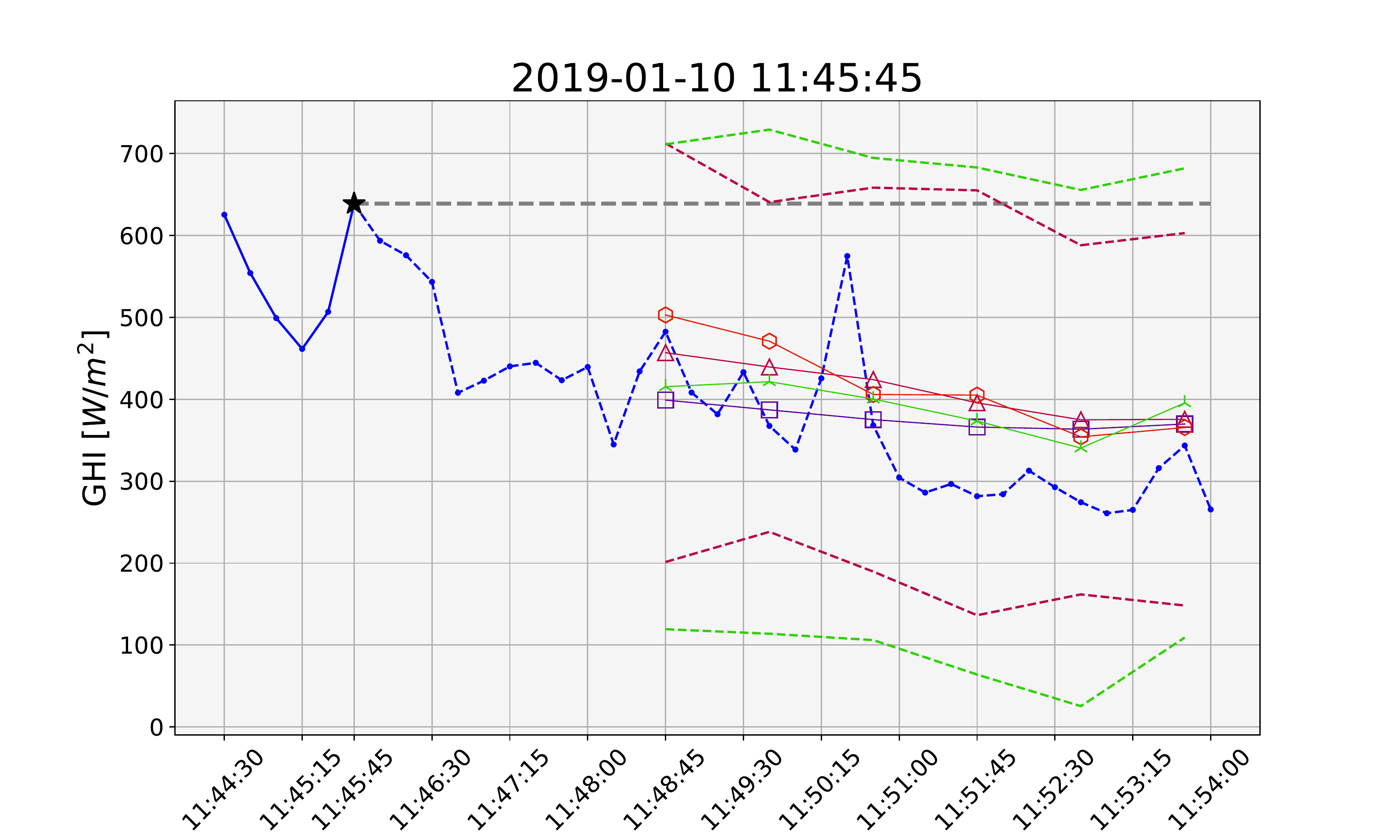}
        \includegraphics[scale = 0.275, trim = {1.5cm, 0cm, 2.5cm, 0cm}, clip]{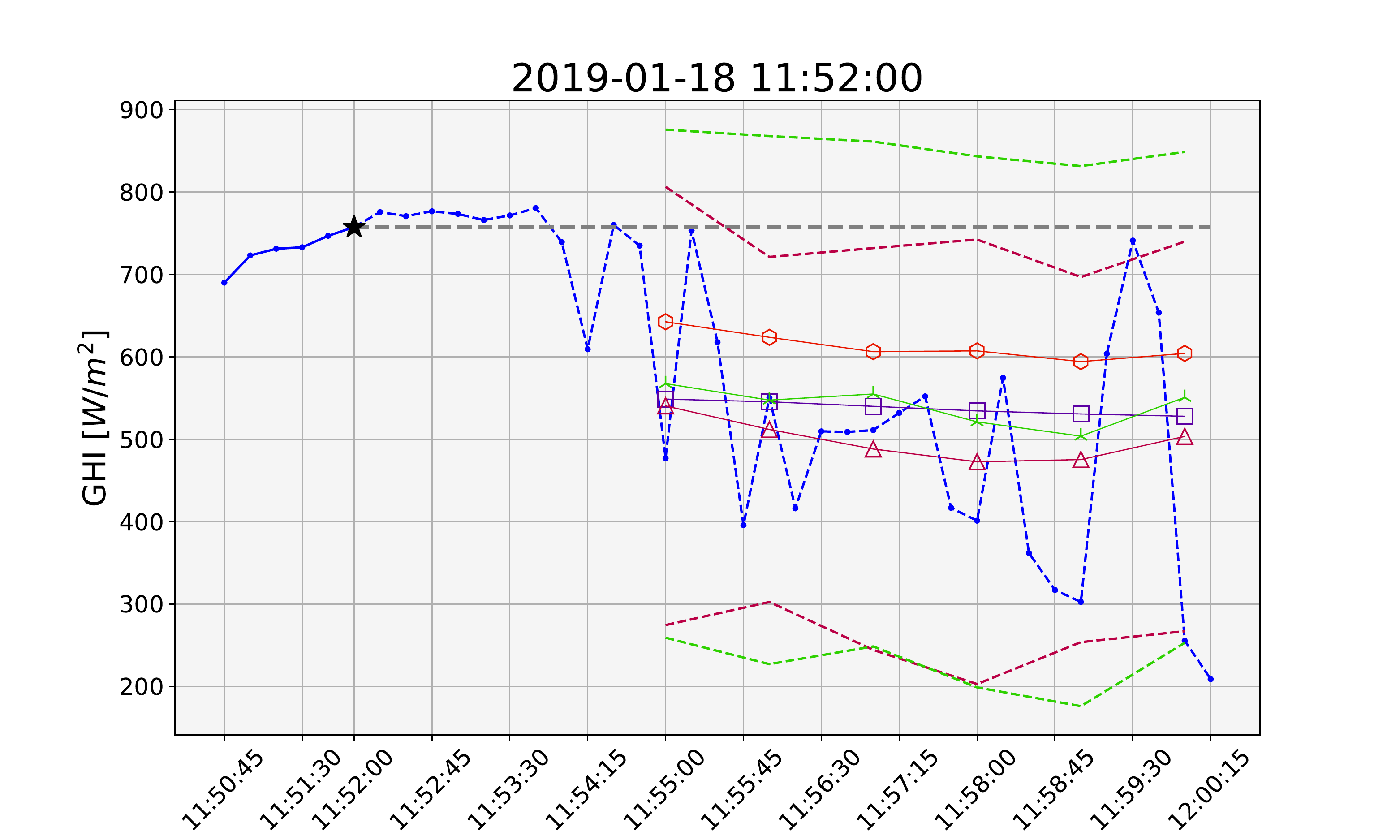}
    \end{subfigure}
    \caption{Illustration of the forecasting performed by the best deterministic models (i.e., independent SVMs and MT-KRR) in purple and red, and Bayesian models (i.e., chain of GPRs and RVMs) in violet and green respectively. The graphs shows four different ramp-down events. The forecasting is performed at the time marked by a bright green star. The predictions and confidence intervals corresponds to the forecast performed by the different multi-tasks forecasting models in that ramp-down event. The presented multi-tasks forecasting models outperforms the persistence (i.e., baseline) anticipating the arrival of clouds. The persistence is shown in gray dashed lines. The chain of GPRs and RVMs perform a probabilistic forecasting, the 95\% lower and upper confidence intervals are shown in dashed lines.} 
    \label{fig:forecasting_example}
\end{figure}

\section{Discussion}

When a cloud is quickly evolving, it has high vorticity and divergence. This is visible in the cloud dynamics displayed by the cloud in Figure~\ref{fig:feature_extraction_algorithm} (right column). These images belong to the sequence shown in Figure~\ref{fig:feature_selection_algorithm}. As seen this figure, the trajectory of the cloud may be predicted but the shape of the cloud is  unpredictable. The forecasting results shown in Figure~\ref{fig:dense_kernel_learning} and Figure~\ref{fig:sparse_kernel_learning} are consistent with this observation. The longer the forecasting horizon is, the more necessary cloud dynamics features are to increase the accuracy of the solar forecasting.

% * forecasting method

A single multi-task model is advantageous when using KRR or RVM (see Figure~\ref{fig:dense_kernel_learning}-\ref{fig:sparse_kernel_learning}). When using SVMs, the best practice is to use independent models (see Figure~\ref{fig:sparse_kernel_learning}), and the chain of models is the best multi-task approach when using GPRs (see Figure~\ref{fig:dense_kernel_learning}). In particular, independent SVMs perform better in shorter forecasting horizons than the chain of GPRs. GPRs is a Bayesian method, thus it is capable of predicting the uncertainty in the forecast. The $\pm 2\sigma$ confidence interval of the Bayesian methods shown in Figure~\ref{fig:forecasting_example} was verified in the entire testing dataset and includes $\approx95\%$ of the forecasting samples. The examples shown in Figure~\ref{fig:forecasting_example} are particularly challenging (i.e., ramp-down events), and consequently the confidence intervals are wider than in safer, or more gradual events.

The models with higher Forecasting Skill (FS) are the independent SVMs and the chain of GPRs when all forecasting horizons are averaged, see left graph in Figure~\ref{fig:forecasting_skill_summary}. The advantage of the chain of GPRs is that the forecast has a confidence interval (Figure~\ref{fig:forecasting_example}). The SVMs require less training time than the rest of models. However, once the independent and chain of GPRs are trained, they require the same time to perform the forecast, (right graph Figure~\ref{fig:forecasting_skill_summary}). The forecasting time required by the multiple output models that use the Kronecker product is higher, because the kernel matrix has more samples. A priori the spare models (i.e., SVMs and RVMs) should have faster testing time than GPRs (see right graph in Figure~\ref{fig:forecasting_skill_summary}), however the optimal GPRs use feature vectors which have less dimensions see Figure~\ref{fig:dense_kernel_learning}-\ref{fig:sparse_kernel_learning}.

The FS of the proposed algorithms are 13.85\% (independent SVMs) and 16.48\% (MT-SVM), 5 and 8 minutes ahead respectively. A previous forecasting method found in the literature proposed 5 and 8 minutes ahead with resolution of 1 minutes and FS of 5.6\% and 15.4\% respectively \cite{Wen2021}. Another method proposed 5 and 10 minutes ahead with resolution of 1 minutes and FS of 14.4\% and 12.1\% respectively \cite{ZANG2020}. Similar work, but not using the exact same forecasting horizons in intra-hour solar forecasting using sky images, achieved a FS of 15.7\% and 10.91\%, 15 and 10 minutes ahead with resolution of 1 minute and 10 minutes respectively \cite{SUN2019, FENG2020}.

The kernel learning method proposed in this investigation improves the intra-hour solar FS with respect to other methods in the literature. We investigate several forecasting horizons between 3 to 8 minutes. The resolution of the forecast is 1 minute, but it is possible to increases the forecasting resolution to 15 seconds as well as the forecasting interval above 1 minute and below 10 minutes. The fact of forecasting the CSI instead of GSI or PV power yields better performances. Equivalently, adding features extracted from sky images also improves the performances in an intra-hour solar forecasting algorithm. In addition, the application of image processing methods to remove cyclostationary effects from the IR camera increase the performance of solar forecasting.

When approaching problems that involve handling large amounts of data, kernel learning methods are constrained by the matrix operations. The ability of a model to generalize fully depends on the minimization of the structural risk and the completeness of the dataset. Therefore, the formation of a training dataset which contains the samples that best represent the underlying function is critical. For this reason, online learning methods that will find the most adequate samples to re-train the models would be the most suitable approach to this problem. Another limitation of the method proposed in this investigation is the hardware. The FOV of the IR camera is narrow so the maximum feasible forecasting horizon is limited to $10$ minutes ahead. In addition, the resolution of the camera is low, and this may affect the accuracy in the approximation of the clouds dynamics \cite{FERMIGIER2017}.
    
\section{Conclusion}

This investigation introduces a comparison of probabilistic and deterministic multi-task intra-hour solar forecasting algorithm based on kernel learning. The forecasting horizon ranges from $3$ to $8$ minutes ahead with resolution of $1$ minute, however the forecasting resolution is adaptable up to intervals of $15$ seconds. The solar forecasting algorithm uses novel feature vectors that include previous clear sky index measurements (i.e., $6$ lags), the position of the Sun (i.e., elevation and azimuth angles), and statistics of features extracted from the cloud dynamics (i.e., temperature, height, and velocity vectors magnitude, divergence and vorticity). An innovative probability of a cloud intersecting the Sun is computed using the approximated potential and streamlines of the wind velocity field. The potential and streamlines are computed using a flow visualization algorithm that approximates the wind velocity field using sequences of consecutive infrared sky images with clouds.

The application of image processing methods to the raw infrared images to remove the cyclostationary effects produced by the atmosphere and the camera outdoor germanium window, increases the forecasting skill of an intra-hour solar forecasting algorithm. In addition, when the clear sky index is used as a predictor instead of the power output or the global solar irradiance the accuracy of a solar forecasting also increases. At the same time, the fact of forecasting clear sky index allows sharing information and the extrapolation of the forecasting between nearby weather stations (i.e., sky imagers).

Convolutional neural networks are the most commonly used machine learning method but they require the optimization of thousands of free parameters. In contrast, the method proposed in this investigation uses the approximation of the wind velocity field streamlines (i.e., pathlines) to anticipate when a cloud will intercept with the Sun. This method is based on fluid mechanics and it is feasible in real time. Beside that, it does not require the inference of the convolutional filters parameters and is adaptive to different weather conditions. 

The accuracy of an intra-hour solar forecasting may be improved by implementing a more efficient outliers removal or sample selection on an online learning algorithm. Other machine learning methods such as ensemble learning or deep learning (i.e., recurrent neural networks) may be investigated to develop solar forecasting based on the wind velocity field. In addition, the comparison of the performances between a convolutional neural network and the proposed method to anticipate the trajectory of clouds using the same dataset would be of great interest to determine which method is more promising. The most straightforward development would be to implement a convolutional neural network that uses the sky images and the global solar irradiance after applying the same processing that is used in this investigation. Future work may also pursue the extension of the proposed method to sky imagers with larger field of view to address applications that require intra-hour solar forecasting from $10$ minutes to $1$ hourss ahead.

\section{Data Availability}

The procedure to acquire and preprocessing the dataset composed of GSI measurements and IR images, plus the hardware are fully described in \cite{TERREN2020d}. The data used in this work is publicly available in a DRYAD repository (\url{https://doi.org/10.5061/dryad.zcrjdfn9m}. Real time and historic data acquired by the weather station is publicly accessible (\url{https://www.wunderground.com/dashboard/pws/KNMALBUQ473}). The software for the feature extraction and selection is available in a GitHub repository (\url{https://github.com/gterren/feature_extraction_and_selection}). The kernel learning methods used in the solar forecasting were implemented using PyTorch\footnote{\url{https://pytorch.org}} and GPyTorch\footnote{\url{https://gpytorch.ai}}. The developed software is also available in a GitHub repository (\url{https://github.com/gterren/kernel_intra-hour_solar_forecasting}).

\section*{Acknowledgments}

This work has been supported by  National Science Foundation (NSF) EPSCoR grant number OIA-1757207 and the King Felipe VI endowed Chair of the UNM. Authors would like to thank the UNM CARC, supported in part by NFS, for providing the high performance computing and large-scale storage resources used in this work. We would also like to thank Marie R. Fernandez for proof reading the manuscript, and Assist. Prof. Matthew Fricke for helping us to efficiently run the experiments the high performances computer.

\appendix

\section{Wavefunction Probabilities}\label{sec:wavefunction_probability}

The wind velocity vector field is defined as the Riemann-Cauchy equations for a complex function $\boldsymbol{Z} = \boldsymbol{\Phi} + \sqrt{-1} \cdot \boldsymbol{\Psi}$ in each pixel $\mathbf{x}_{i,j}$. However, the maxima of the velocity vector field has to be in the Sun intersecting streamline for our purposes,
\begin{equation}
    \begin{split}
        \boldsymbol{\Phi}^\prime &= | \boldsymbol{\Phi} - \Phi_{i_0, j_0} | \\ 
        \boldsymbol{\Phi}^{\prime\prime} &= \max \left[ \boldsymbol{\Phi}^\prime \right] - \boldsymbol{\Phi}^\prime.
    \end{split}
\end{equation}
The corresponding velocity vector field in complex function is $\boldsymbol{Z} = \boldsymbol{\Phi}^{\prime\prime} + \sqrt{-1} \cdot \boldsymbol{\Psi}$, and its probability is modelled as a wave,
\begin{equation}
    p \left(\mathbf{x}_{i,j} \middle| \phi_{i,j}, \psi_{i,j} \right) = \frac{ | \phi_{i,j}^{\prime\prime} + \sqrt{-1} \cdot \psi_{i,j} |^2}{\| \boldsymbol{\Phi}^{\prime\prime} + \sqrt{-1} \cdot \boldsymbol{\Psi} \|^2_\mathcal{F}}
\end{equation}
where $\| \cdot \|_\mathcal{F}$ is the Frobenius norm.

\bibliographystyle{unsrt}  
\bibliography{mybibfile_arxiv}

\end{document}